\pdfoutput=1
\RequirePackage{fix-cm}
\documentclass[twocolumn]{svjour3}          %
\smartqed  %
\usepackage[pagebackref=true,breaklinks=true,letterpaper=true,colorlinks,citecolor=blue,bookmarks=false]{hyperref}
\usepackage{graphicx}
\usepackage{times}
\usepackage{epsfig}
\usepackage{graphicx}
\usepackage{amsmath}
\usepackage{amssymb}
\usepackage{tikz}
\usepackage{pgfplots}
\usepackage{stmaryrd}
\usepackage{booktabs}
\usepackage{bbm}
\usepackage{xspace}
\usepackage{natbib}

 \makeatletter
 \DeclareRobustCommand\onedot{\futurelet\@let@token\@onedot}
 \def\@onedot{\ifx\@let@token.\else.\null\fi\xspace}
 \def\eg{e.g\onedot} 
 \def\ie{i.e\onedot}

 \makeatother

\DeclareRobustCommand{\Figsref}[1]{Figures~\ref{#1}}

\DeclareRobustCommand{\Eqnsref}[1]{Equations~(\ref{#1})}

\graphicspath{{./}}

\newcommand{\iccvArch}{Ask Your Neurons\xspace}
\newcommand{\newArch}{Refined Ask Your Neurons\xspace}

\newcommand{\AproachName}{Ask Your Neurons\xspace}
\newcommand{\daquarNew}{DAQUAR-Consensus\xspace}

\newcommand{\qenc}[1]{\Psi_{\text{#1}}(\bs{q})}
\newcommand{\ienc}{\Phi(\bs{x})\xspace}

\newcommand{\bs}[1]{\ensuremath{\boldsymbol{#1}}}

\DeclareMathOperator* {\argmax}{arg\, max}

\journalname{myjournal}
\begin{document}

\title{Ask Your Neurons:\\A Deep Learning Approach to Visual Question Answering}

\author{Mateusz Malinowski         \and
        Marcus Rohrbach \and
        Mario Fritz
}

\institute{Mateusz Malinowski \at
              Max Planck Institute for Informatics \\
              Saarbr{\"u}cken, Germany \\
              \email{mmalinow@mpi-inf.mpg.de}           %
           \and
           Marcus Rohrbach \at
              UC Berkeley EECS and ICSI \\
              Berkeley, CA, United States \\
              \email{rohrbach@berkeley.edu}
           \and
           Mario Fritz \at
             Max Planck Institute for Informatics \\
             Saarbr{\"u}cken, Germany \\
             \email{mfritz@mpi-inf.mpg.de}
}

\date{Received: date / Accepted: date}

\maketitle

\begin{abstract}
We propose a Deep Learning approach to the visual question answering task, where machines answer to questions about real-world images. %
By combining latest advances in image representation and natural language processing, we propose \AproachName, a scalable, jointly trained, end-to-end formulation to this problem.
In contrast to previous efforts, we are facing a multi-modal problem where the language output (answer) is conditioned on visual and natural language inputs (image and question).
We evaluate our approaches on the DAQUAR as well as the VQA dataset where we also report various baselines, including an analysis how much information is contained in the language part only.
To study human consensus, we propose two novel metrics and collect additional answers which extend the original DAQUAR dataset to \daquarNew.
Finally, we evaluate a rich set of design choices how to encode, combine and decode information in our proposed Deep Learning formulation.


\end{abstract}

\section{Introduction}\label{sec:intro}

With the advances of natural language processing and image understanding, more complex and demanding tasks have become within reach. Our aim is to take advantage of the most recent developments to push the state-of-the-art for answering natural language questions on real-world images. This task unites inference of a question intent and visual scene understanding with an answer generation task.

Recently, architectures based on the idea of layered, end-to-end trainable artificial neural networks have improved the state of the art across a wide range of diverse tasks. Most prominently Convolutional Neural Networks have raised the bar on image classification tasks \citep{krizhevsky2012imagenet} and Long Short Term Memory Networks \citep{hochreiter97nc} are dominating performance on a range of sequence prediction tasks such as machine translation \citep{sutskever14nips}.

\begin{figure}[t]
\begin{center}
  \includegraphics[width=\columnwidth]{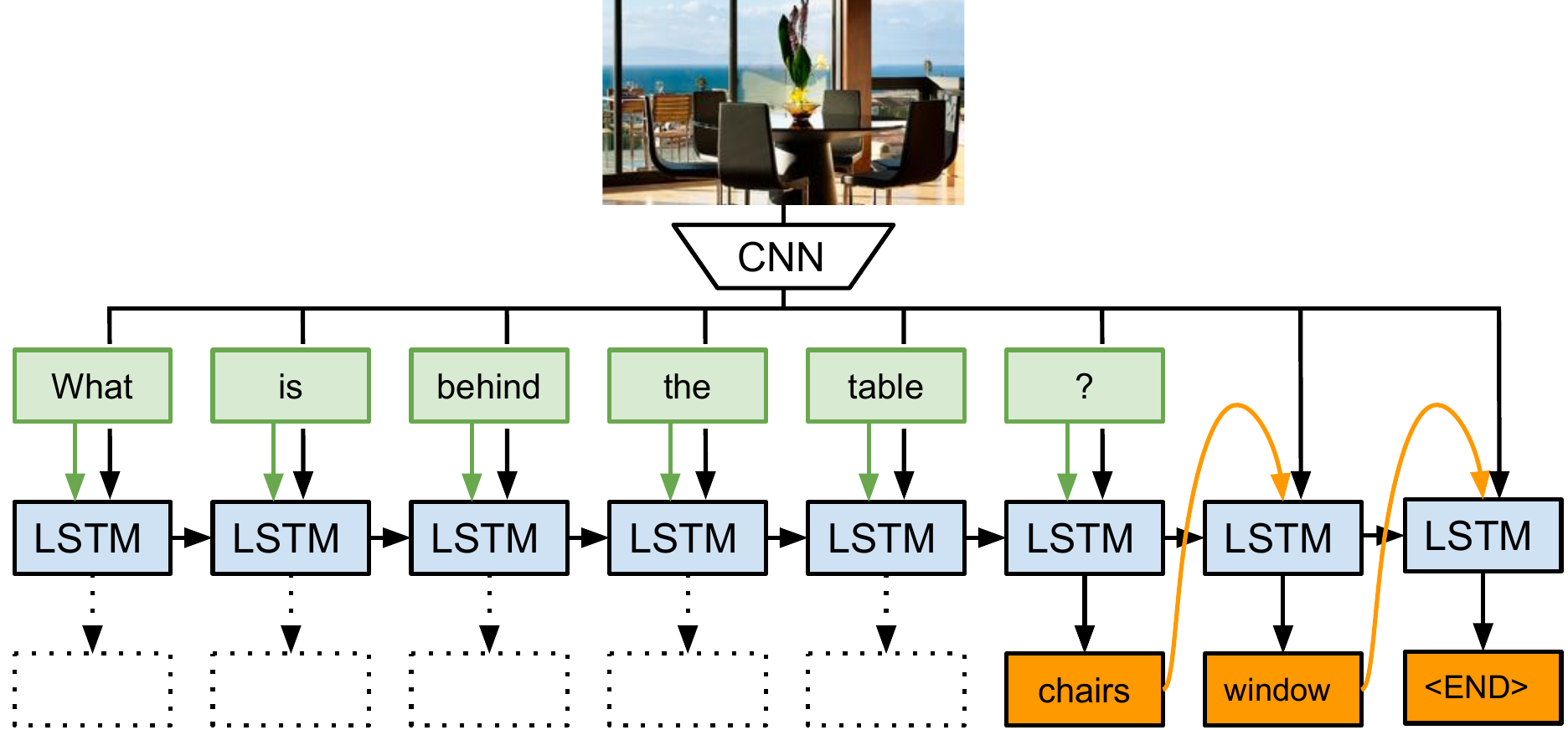}
  \caption[labelInTOC]{Our
 approach \emph{\iccvArch} to question answering with a Recurrent Neural Network using Long Short Term Memory (LSTM). To answer a question about an image, we feed in both, the image (CNN features) and the question (green boxes) into the LSTM. After the (variable length) question is encoded, we generate the answers (multiple words, orange boxes). During the answer generation phase the previously predicted answers are fed into the LSTM until the $\langle$END$\rangle$ symbol is predicted. See \autoref{sec:iccvArch} for more details.}
  \label{fig:teaser}
\end{center}
\end{figure}

Most recently, these two trends of employing neural architectures have been combined fruitfully with methods that can generate image \citep{karpathy15cvpr} and video descriptions \citep{venugopalan15iccv}. Both are conditioned on the visual features that stem from deep learning architectures and employ recurrent neural network approaches to produce descriptions.

To further push the boundaries and explore the limits of deep learning architectures, we propose an architecture for answering questions about images. In contrast to prior work, this task needs conditioning on language as well visual input. Both modalities have to be interpreted and jointly represented as an answer depends on inferred meaning of the question and image content.

While there is a rich body of work on natural language understanding that has addressed textual question answering tasks based on semantic parsing, symbolic representation and deduction systems, which also has seen applications to question answering about images \citep{malinowski14nips}, there is evidence that deep architectures can indeed achieve a similar goal
\citep{weston2014memory}.
This motivates our work to seek end-to-end architectures that learn to answer questions in a single, holistic model.

We propose \emph{\AproachName}, an approach to question answering with a recurrent neural network at its core.
An overview is given in \autoref{fig:teaser}. The image information is encoded via a Convolutional Neural Network (CNN) and the question together with the visual representation is fed into a Long Short Term Memory (LSTM) network.
The system is trained to produce the correct answer to the question about the image.
CNN and LSTM are trained jointly and end-to-end starting from words and pixels.

\paragraph{Outline.}
In \autoref{sec:method}, we present our novel approach based on recurrent neural networks for the challenging task of answering questions about images, which we presented originally in \citet{malinowski2015ask}. The approach combines a CNN with an LSTM into an end-to-end architecture that predicts answers conditioning on a question and an image. \autoref{sec:results} shows that the proposed approach
doubles performance  compared to a prior
symbolic approach
on this task. %
We collect additional data to study human consensus on this task, propose two new metrics sensitive to these effects, and provide a new baseline, by asking humans to answer the questions without observing the image.
We demonstrate a variant of our system that also answers question
without accessing any visual information, which beats the human baseline.
We also frame the multimodal approach to answer questions about images that combines LSTM with CNN \citep{malinowski2015ask} as a special instance of an encoder-decoder framework.  This modular perspective, shown in \autoref{sec:alternative_approaches}, allows us to study different design choices on a large scale visual question answering dataset VQA. \autoref{section:analysis_on_vqa} shows our analysis that leads to an improved visual question answering architecture. A deeper visual encoding together with several important design choices lead to a model that achieves stronger performance on VQA and DAQUAR datasets. %

\section{Related Work}\label{sec:related_work}
Since we have proposed a challenge and first methods for answering questions about real-world images
\citep{malinowski14nips, malinowski14visualturing,malinowski2015hard, malinowski2015ask}, frequently also referred to as ``Visual Question Answering'', numerous follow up works have appeared. In the following we first discuss related tasks and subtasks, then early approaches to tackle 
a broader
Visual Turing Test
and datasets proposed for it. Finally, we discuss the relations to our work.

\subsection{Convolutional neural networks for visual recognition}
One component to answer questions about images is to extract information from visual content.
Since the proposal of AlexNet \citep{krizhevsky2012imagenet}, Convolutional Neural Networks (CNNs) have become dominant and most successful approaches to extract relevant representation from the image.
CNNs directly learn the representation from the raw image data and are trained on large image corpora, typically ImageNet \citep{ILSVRCarxiv14}.
Interestingly, after these models are pre-trained on ImageNet, they can typically be adapted for other tasks. In this work, we evaluate how well the most dominant and successful CNN models can be adapted for the Visual Turing Test. Specifically, we evaluate \textit{AlexNet} \citep{krizhevsky2012imagenet}, \textit{VGG} \citep{simonyan2014very}, \textit{GoogleNet} \citep{szegedy2014going}, and \textit{ResNet} \citep{he2015deep}. These models, reportedly, achieve more and more accurate results on the ImageNet dataset, and hence, arguably, serve as increasingly stronger models of visual perception.

\subsection{Encodings for text sequence understanding}
The other important component to answer a question about an image is to understand the natural language question, which means here building a representation of a variable length sequence of words (or characters, but we will focus only on the words in this work).
The first approach is to encode all words of the question as a Bag-Of-Words \citep{manning1999foundations}, and hence ignoring an order in the sequence of words.
Another option is to use, similar to the image encoding, a CNN with pooling to handle variable length input \citep{kim2014convolutional,kalchbrenner2014convolutional}.
Finally, Recurrent Neural Networks (RNNs) are methods developed to directly handle sequences, and have shown recent success on natural language tasks such as machine translation \citep{cho2014learning,sutskever14nips}. In this work we investigate a Bag-Of-Words (BOW), a CNN, and two RNN variants  (LSTM \citep{hochreiter97nc} and GRU \citep{cho2014learning}) to encode the question.

\subsection{Combining RNNs and CNNs for description of visual content}
The task of describing visual content like still images as well as videos has been successfully addressed with a combination of encoding the image with CNNs and decoding, \ie predicting the sentence description with an RNN \citep{donahue15cvpr,karpathy15cvpr,venugopalan15naacl,vinyals2014show,zitnick2013learning}. This is achieved by using the RNN model that first gets to observe the visual content and is trained to afterwards predict a sequence of words that is a description of the visual content. Our work extends this idea to question answering, where we formulate a model trained to either generate or classify an answer based on visual as well as natural language input.

\subsection{Grounding of natural language and visual concepts}
Dealing with natural language input does involve the association of words with meaning. This is often referred to as the grounding problem - in particular if the ``meaning'' is associated with a sensory input. While such problems have been historically addressed by symbolic semantic parsing techniques \citep{krishnamurthy2013jointly,matuszek2012joint}, there is a recent trend of machine learning-based approaches \citep{kong2014you,karpathy2014deep,karpathy15cvpr,akata2016multi,hu16cvpr,rohrbach16eccv,mao16cvpr,wang2016cvpr,hu16eccv,plummer16arxiv}
to find the associations. These approaches have partially been enabled by recently proposed larger scale datasets \citep{kazemzadeh14emnlp,plummer15iccv,yu16eccv}, providing phrases or referential expressions which are associated with their corresponding image location.
Answering questions about images can be interpreted as first grounding the question in the image and then predicting an answer. Our approach thus is similar to the latter approaches in that we do not enforce or evaluate any particular representation of ``meaning'' on the language or image modality. We treat this as latent and leave it to the joint training approach to establish an appropriate hidden representation to link the visual and textual representations.

\subsection{Textual question answering}
Answering on purely textual questions has been studied in the NLP community \citep{berant2014semantic,liang2013learning} and state of the art techniques typically employ semantic parsing to arrive at a logical form capturing the intended meaning and infer relevant answers. Only recently, the success of the previously mentioned neural sequence models, namely RNNs, has carried over to this task \citep{iyyer2014neural,weston2014memory}.
More specifically \citet{iyyer2014neural} use dependency-tree Recursive NN instead of LSTM, and reduce the question-answering problem to a classification task. %
\citet{weston2014memory} propose different kind of network - memory networks - that is used to answer questions about short stories. In their work, all the parts of the story are embedded into different ``memory cells'', and next a network is trained to attend to relevant cells based on the question and decode an answer from that. A similar idea has also been applied to question answering about images, for instance by \citet{yang2015stacked}.

\subsection{Towards a Visual Turing Test} %
Recently, a large number architectures have been proposed to approach the Visual Turing Test \citep{malinowski14visualturing},
in which they mainly tackle a particularly important subtask that tests machines regarding their abilities to answer to questions about real-world images.
Such methods
range from symbolic to neural based.
There are also architectures that combine both symbolic and neural paradigms together. Some approaches use explicit visual representation in the form of bounding boxes surrounding objects of interest, while other use global full frame image representation, or soft attention mechanism. Yet others use an external knowledge base that helps in answering questions.
\paragraph{Symbolic based approaches.}
  In our first work
  towards a
  Visual Turing Test \citep{malinowski14nips}, we present a question answering system based on a semantic parser on a varied set of human question-answer pairs. Although it is the first attempt to handle question answering on DAQUAR, and despite its introspective benefits, it is a rule-based approach that requires a careful schema crafting, is not that scalable, and finally it strongly depends on the output of visual analysis methods as joint training in this model is not yet possible.
Due to such limitations, the community has rather shifted towards either neural based or combined approaches.

\paragraph{Deep Neural Approaches with full frame CNN.}
Several contemporary approaches use a global image representation, \ie they encode the whole image with a CNN. Questions are then encoded with an RNN \citep{malinowski2015ask,ren2015image,gao2015you} or a CNN \citep{learning_to_answer_questions}.
  In contrast to symbolic based approaches, neural based architectures offer scalable and joint end-to-end training that liberates  them from  ontological commitment that would otherwise be introduced by a semantic parser. Moreover, such approaches are not `hard' conditioned on the visual input and therefore can naturally take advantage of different language biases in question answer pairs, which can be interpret as learning common sense knowledge.

\paragraph{Attention-based Approaches.}
Following \citet{xu15icml}, who proposed to use spatial attention for image description, \citet{yang2015stacked,xu2015ask,zhu16cvpr,chen2015abc,shih2015look,fukui16emnlp} predict a latent weighting (attention) of spatially localized images features (typically a convolutional layer of the CNN) based on the question. The weighted image representation rather than the full frame feature representation is then  used as a basis for answering the question.
In contrast to the previous models using attention, Dynamic Memory Networks (DMN) \citep{kumar15arxiv,xiong16dynamic} first pass all spatial image features through a bi-directional GRU that captures spatial information from the neighboring image patches, and next retrieve an answer from a recurrent attention based neural network that allows to focus only on a subset of the visual features extracted in the first pass.
Another interesting direction has been taken by \citet{ilievski2016fda} who run state-of-the-art object detector of the classes extracted from the key words in the question. In contrast to other attention mechanisms, this approach offers a focused, question dependent, ``hard'' attention.

\paragraph{Answering with an external knowledge base.}
\citet{wu16cvpr} argue for an approach that first represents an image as an intermediate semantic attribute representation, and next query external knowledge sources based on the most prominent attributes and relate them to the question. With the help of such external knowledge bases, this approach captures richer semantic representation of the world, beyond what is directly contained in images.

\paragraph{Compositional approaches.}
A different direction is taken by \citet{andreas16naacl,andreas16cvpr} who predict the most important components to answer the question with a natural language parser. The components are then mapped to neural modules, which are composed to a deep neural network based on the parse tree. While each question induces a different network, the modules are trained jointly across questions. This work compares to \citet{malinowski14nips} by exploiting explicit assumptions about the compositionality of natural language sentences.
Related to the Visual Turing Test, \citet{malinowski2014pooling} have also combined a neural based representation with the compositionality of the language for the text-to-image retrieval task.

\paragraph{Dynamic parameters.}
 \citet{noh2015images} have an image recognition network and a Recurrent Neural Network (GRU) that dynamically change the parameters (weights) of visual representation based on the question.
More precisely, the parameters of its second last layer are dynamically predicted from the question encoder network and in this way changing for each question. While question encoding and image encoding is pre-trained, the network learns parameter prediction only from image-question-answer triples.

\subsection{Datasets for visual question answering}
Datasets are a driving force for the recent progress in visual question answering.
A large number of visual question answering datasets have recently been proposed. The first proposed datasets is DAQUAR \citep{malinowski14nips}, which contains about $12.5$ thousands manually annotated question-answer pairs about $1449$ indoor scenes \citep{silbermanECCV12}. While the dataset originally contained a single answer (that can consist of multiple words) per question, in this work we extend the dataset by collecting additional answers for each questions. This captures uncertainties
in evaluation.
We evaluate our approach on this dataset and discuss several consensus evaluation metrics that take the extended annotations into account.
In parallel to our work, 
\citet{geman2015visual} developed another variant of the Visual Turing Test. Their work, however, focuses on sequences of yes/no type of questions, and provides detailed object-scene annotations.

Shortly after the introduction of DAQUAR, three other large-scale datasets have been proposed. All are based on MS-COCO \citep{lin2014microsoft}.
\citet{gao2015you} have annotated about $158k$ images with $316k$  Chinese question answer pairs together with their corresponding English translations. \citet{ren2015image} have taken advantage of the existing annotations for the purpose of image description generation task and transform them into question answer pairs with the help of a set of hand-designed rules and a syntactic parser \citep{klein2003accurate}. This procedure has approximately generated $118k$  question answer pairs. Finally and currently  the most popular, large scale dataset on question answering about images is VQA \citep{antol2015vqa}. It has approximately $614k$  questions about the visual content of about $205k$   real-world images (the whole VQA dataset also contains $150k$ questions about $50k$ abstract scenes that are not considered in this work). Similarly to our Consensus idea, VQA provides $10$ answers per each image. For the purpose of the challenge the test answers are not publicly available. We perform one part of the experimental analysis in this paper on the VQA dataset, examining different variants of our proposed approach.

Although simple, automatic  performance evaluation metrics have been a part of building first visual question answering datasets \citep{malinowski14nips,malinowski14visualturing,malinowski2015hard}, \citet{yu2015visual} have simplified the evaluation even further by introducing Visual Madlibs - a multiple choice question answering by filling the blanks task. In this task, a question answering architecture has to choose one out of four provided answers for a given image and the prompt. Formulating question answering task in this way has wiped out ambiguities in answers, and just a simple accuracy metric can be used to evaluate different architectures on this task. Yet, the task requires holistic reasoning about the images, and despite of simple evaluation, it remains challenging for machines.

The Visual7W \citep{zhu16cvpr} extends canonical question and answer pairs with additional groundings of all objects appearing in the questions and answers to the image by annotating the correspondences. It contains natural language answers, but also answers which require to locate the object, which is then similar to the task of explicit grounding discussed above.
Visual7W builds question answer pairs based on the Visual Genome dataset \citep{krishna16arxiv}, and contains about $330k$ questions.
In contrast to others such as VQA \citep{antol2015vqa} or DAQUAR \citep{malinowski14nips} that has collected unconstrained question answer pairs, the Visual Genome focuses on the six, so called, Ws: \emph{what}, \emph{where}, \emph{when}, \emph{who}, \emph{why}, and \emph{how}, which can be answered with a natural language answer. An additional 7th question --~\emph{which}~-- requires a bounding box location as answer.
Similarly to Visual Madlibs \citep{yu2015visual}, Visual7W also contains multiple-choice answers.

Related to Visual Turing Test, \citet{sreyasi16icmr} have proposed collective memories and Xplore-M-Ego - a dataset of images with natural language queries, and a media retrieval system. This work focuses on a user centric, dynamic scenario, where the provided answers are conditioned not only on questions but also on the geographical position of the questioner.

Moving from asking questions about images to questions about video enhances typical questions with temporal structure. \citet{zhu2015uncovering} propose a task which requires to fill in blanks the captions associated with videos. The task requires inferring the past, describing the present and predicting the future in a diverse set of video description data ranging from cooking videos \citep{regneri13tacl} over web videos \citep{trecvid_med14} to movies \citep{rohrbach15cvpr}. \citet{tapaswi16cvpr} propose MovieQA, which requires to understand long term connections in the plot of the movie.
Given the difficulty of the data, both works provide multiple-choice answers.

\subsection{Relations to our work}
The original version of this work \citep{malinowski2015ask} belongs to the category of ``Deep Neural Approaches with full frame CNN'', and is among the very first methods of this kind (\autoref{sec:iccvArch}). We extend \citep{malinowski2015ask} by introducing a more general and modular encoder-decoder perspective (\autoref{sec:alternative_approaches}) that encapsulates a few different neural approaches. Next, we broaden our original analysis done on DAQUAR (\autoref{sec:results}) to the analysis of different neural based approaches on VQA showing the importance of getting a few details right together with benefits of a stronger visual encoder (\autoref{section:analysis_on_vqa}). Finally, we transfer lessons learnt from VQA \citep{antol2015vqa} to DAQUAR \citep{malinowski14nips}, showing a significant improvement on this challenging task (\autoref{section:analysis_on_vqa}).

\section{\AproachName}
\label{sec:method}

Answering questions about images can be formulated as the problem of predicting an answer %
$\bs{a}$ given an image $\bs{x}$ and a question $\bs{q}$ according to a parametric probability measure:
\begin{equation}
\label{eq:problem}
\bs{\hat{a}}=\argmax_{\bs{a}\in \mathcal{A}}p(\bs{a}|\bs{x},\bs{q};\bs{\theta})
\end{equation}
where $\bs{\theta}$ represent a vector of all parameters to learn and $\mathcal{A}$ is a set of all answers. %
The question $\bs{q}$ is a sequence of words, \ie
$\bs{q}=\left[\bs{q}_1,\ldots,\bs{q}_{n}\right]$, where each $\bs{q}_t$ is the $t$-th word question  with $\bs{q}_n= ``?"$ encoding the question mark - the end of the question.
In the following we describe how we represent $\bs{x}$, $\bs{a}$, $\bs{q}$, and $p(\cdot|\bs{x},\bs{q};\bs{\theta})$ in more details.
In a scenario of \textbf{multiple word answers}, we consequently decompose the problem to predicting a set of answer words $\bs{a}_{\bs{q},\bs{x}} = \left\{\bs{a}_1, \bs{a}_2, ..., \bs{a}_{\mathcal{N}(q,x)}\right\}$, where $\bs{a}_t$ are words from a finite vocabulary $\mathcal{V'}$, and $\mathcal{N}(q,x)$ is the number of answer words for the given question and image.
In our approach, named \AproachName, we propose to tackle the problem as follows.  To predict multiple words we formulate the problem as predicting a sequence of words from the vocabulary $\mathcal{V}:=\mathcal{V'}\cup\left\{\$\right\}$ where the extra token $\$$
indicates the end of the answer sequence, and points out that the question has been fully answered. %
We thus formulate the prediction procedure recursively:
\begin{equation}
\label{eq:recursivePred}
\bs{\hat{a}}_t=\argmax_{\bs{a}\in \mathcal{V}}p(\bs{a}|\bs{x},\bs{q},\hat{A}_{t-1};\bs{\theta})
\end{equation}
where $\hat{A}_{t-1}=\left\{\bs{\hat{a}}_1,\ldots,\bs{\hat{a}_{t-1}}\right\}$ is the set of previous words, with $\hat{A}_{0}=\left\{\right\}$ at the beginning, when our approach has not given any answer word so far. The approach is terminated when $\hat{a}_t=\$$.
We evaluate the method solely based on the predicted answer words ignoring the extra token $\$$.
To ensure uniqueness of the predicted answer words, as we want to predict the \emph{set} of answer words, the prediction procedure can be
be trivially changed
by maximizing over
$\mathcal{V}\setminus\hat{A}_{t-1}$. However, in practice, our algorithm learns to not predict any previously predicted words.

If we only have \textbf{single word answers}, or if we model each multi-word answer as a different answer (\ie vocabulary entry), we use \autoref{eq:problem} only once to pick the most likely answer.

In the following we first present a \iccvArch that models multi-word answers with a single recurrent network for question and image encoding and answer prediction (\autoref{sec:iccvArch}) and then present a more general and modular framework with question and image encoders, as well as answer decoder as  modules (\autoref{sec:newArch}).

\subsection{Method}
\label{sec:iccvArch}

\begin{figure}[t]
\begin{center}
  \includegraphics[width=.7\columnwidth]{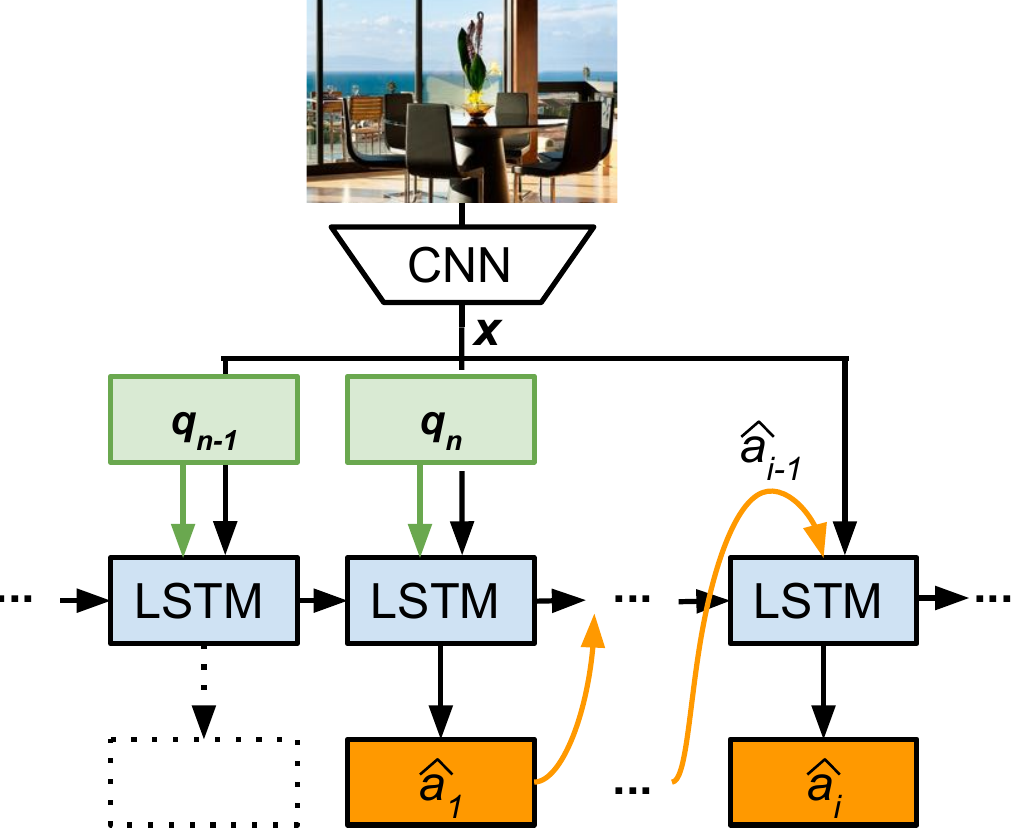}
  \caption{Our approach \emph{\AproachName},  see \autoref{sec:method} for details.}
  \label{fig:lstm-approach}
\end{center}
\end{figure}

As shown in \autoref{fig:teaser} and \autoref{fig:lstm-approach}, we  feed our approach \AproachName with a question as a sequence of words.
Since our problem is formulated as a variable-length input-output-sequence,
we decide to model the parametric distribution $p(\cdot|\bs{x},\bs{q};\bs{\theta})$ of  \AproachName with a recurrent neural network and a  softmax prediction layer. More precisely, \AproachName is a deep network built of CNN \citep{lecun1998gradient} and Long-Short Term Memory (LSTM) \citep{hochreiter97nc}. We decide on LSTM as it has been recently shown to be effective in learning a variable-length sequence-to-sequence mapping \citep{donahue15cvpr,sutskever14nips}.

Both question and answer words are represented with one-hot vector encoding (a binary vector with exactly one non-zero entry at the position indicating the index of the word in the vocabulary) and embedded in a lower dimensional space, using a jointly learnt latent linear embedding.
In the training phase, we augment the question words sequence $\bs{q}$ with the corresponding ground truth answer words sequence $\bs{a}$, \ie $\bs{\hat{q}} := \left[\bs{q}, \bs{a}\right]$. During the test time, in the prediction phase, at time step $t$, we augment $\bs{q}$ with previously predicted answer words $\hat{\bs{a}}_{1..t} := \left[\bs{\hat{a}}_1,\ldots,\bs{\hat{a}}_{t-1}\right]$, \ie $\bs{\hat{q}}_{t} := \left[\bs{q},\bs{\hat{a}}_{1..t}\right]$.
This means the question $\bs{q}$ and the previous answer words are encoded implicitly in the hidden states of the LSTM, while the latent hidden representation is learnt. We encode the image $\bs{x}$ using a CNN and provide it at every time step as input to the LSTM.
We set the input
$\bs{v}_t$
as a concatenation of
$\left[\Phi(\bs{x}), \bs{\hat{q}}_t\right]$,
where $\Phi(\cdot)$ is the CNN encoding.

\subsubsection{Long-Short Term Memory (LSTM)}
\label{sec:LSTM}
\begin{figure}[t]
\begin{center}
  \includegraphics[width=.64\columnwidth]{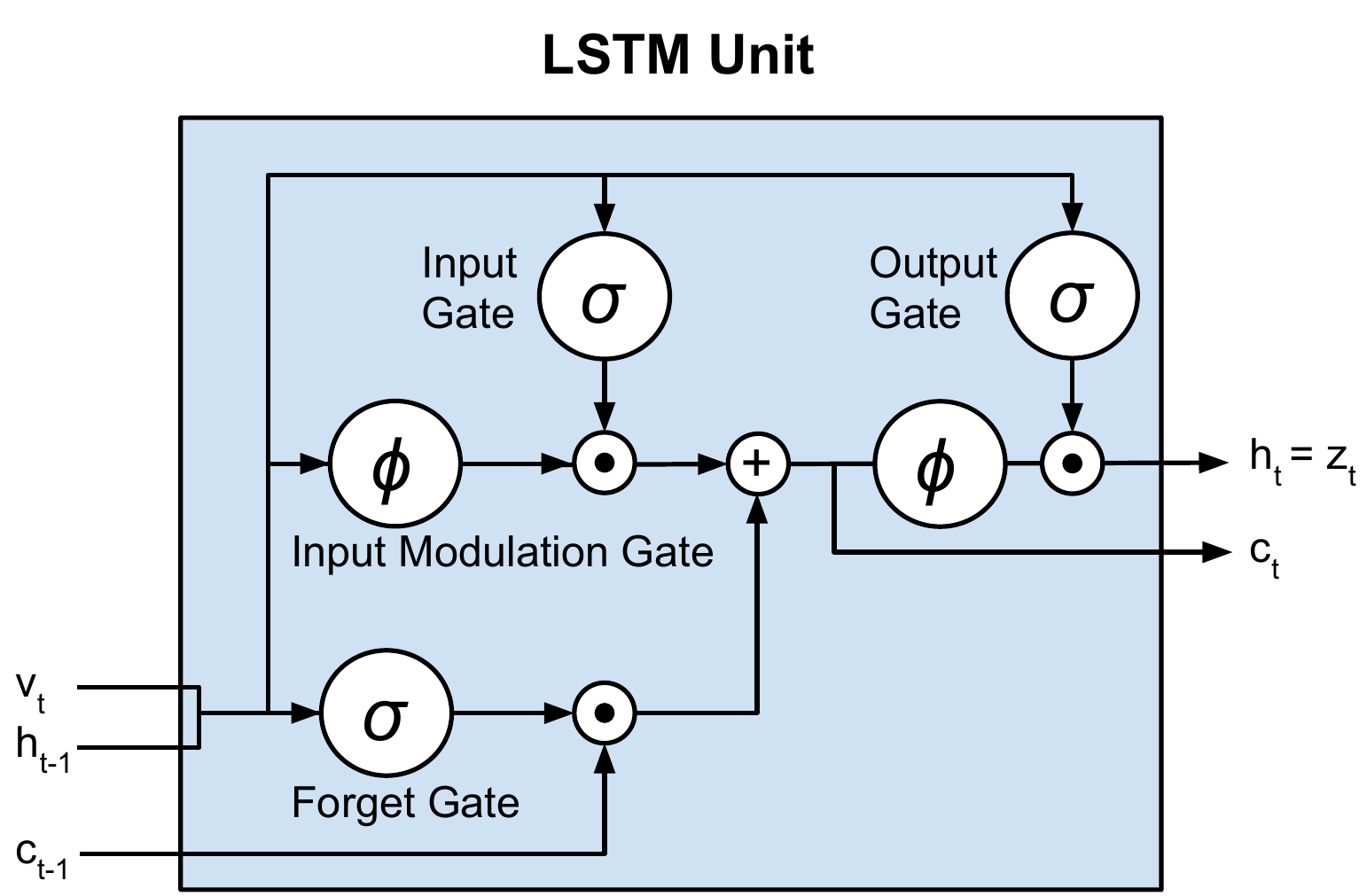}
  \caption[labelInTOC]{LSTM unit. See \autoref{sec:LSTM}, \Eqnsref{eq:i}-(\ref{eq:h}) for details.}
  \label{fig:lstm}
\end{center}
\end{figure}
As visualized in detail in \autoref{fig:lstm}, the LSTM unit takes an input vector $\bs{v}_t$ at each time step $t$ and predicts an output
word
 $\bs{z_t}$ which is equal to its latent hidden state
 $\bs{h}_t$. As discussed above, $\bs{z_t}$ is a linear embedding of the corresponding answer word $\bs{a_t}$.
In contrast to a simple RNN unit, the LSTM unit additionally maintains a memory cell $\bs{c}$. This allows to learn long-term dynamics more easily and significantly reduces the vanishing and exploding gradients problem~\citep{hochreiter97nc}.
More precisely, we use the LSTM unit as described in \citet{zaremba14arxiv}.
 With the  \textit{sigmoid} nonlinearity $\sigma:\mathbb{R} \mapsto [0, 1]$, $\sigma(v) = \left(1 + e^{-v}\right)^{-1}$ and the \textit{hyperbolic tangent} nonlinearity $\phi:\mathbb{R} \mapsto [-1, 1]$, $\phi(v) = \frac{e^v - e^{-v}}{e^v + e^{-v}} = 2\sigma(2v) - 1$, the LSTM updates for time step $t$ given inputs $\bs{v}_t$, $\bs{h}_{t-1}$, and the memory cell $\bs{c}_{t-1}$ as follows:
\begin{align}
\bs{i}_t &= \sigma(W_{vi}\bs{v}_t + W_{hi}\bs{h}_{t-1} + \bs{b}_i)\label{eq:i}\\
\bs{f}_t &= \sigma(W_{vf}\bs{v}_t + W_{hf}\bs{h}_{t-1} + \bs{b}_f)\label{eq:f} \\
\bs{o}_t &= \sigma(W_{vo}\bs{v}_t + W_{ho}\bs{h}_{t-1} + \bs{b}_o) \label{eq:o}\\
\bs{g}_t &=   \phi(W_{vg}\bs{v}_t + W_{hg}\bs{h}_{t-1} + \bs{b}_g)\label{eq:g} \\
\bs{c}_t &= \bs{f}_t \odot \bs{c}_{t-1} + \bs{i}_t \odot \bs{g}_t \label{eq:c}\\
\bs{h}_t &= \bs{o}_t \odot \phi(\bs{c}_t)\label{eq:h}
\end{align}
where $\odot$ denotes element-wise multiplication.
All the weights $W$ and biases $b$ of the network are learnt jointly with the cross-entropy loss. Conceptually, as shown in \autoref{fig:lstm},  \autoref{eq:i} corresponds to the input gate, \autoref{eq:g} the input modulation gate, and \autoref{eq:f} the forget gate, which determines how much to keep from the previous memory $c_{t-1}$ state.
As \Figsref{fig:teaser} and \ref{fig:lstm-approach} suggest, all the output predictions that occur before the question mark are excluded from the loss computation, so that the model is penalized solely based on the predicted answer words.

\subsection{\newArch}
\label{sec:alternative_approaches}
\label{sec:newArch}
\begin{figure}[t]
\begin{center}
  \includegraphics[width=.7\columnwidth]{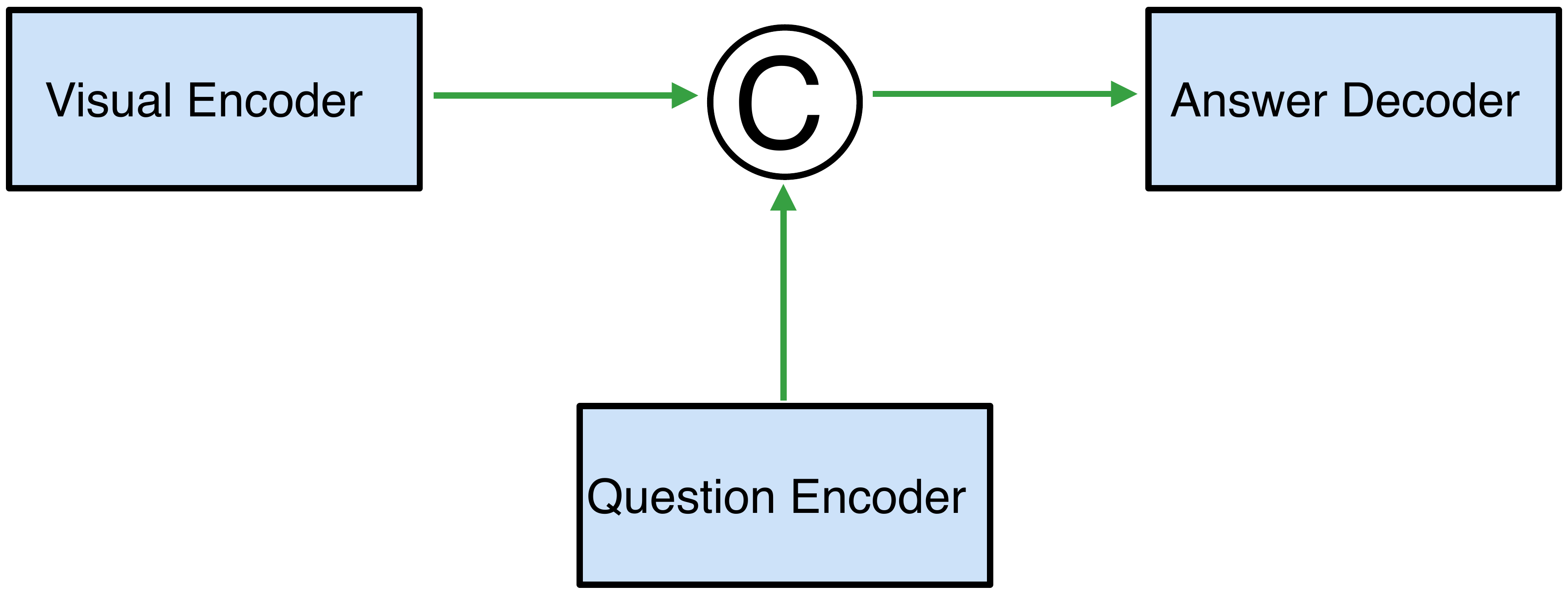}
  \caption{Our \emph{\newArch} architecture for answering questions about images that includes the following modules: visual and question encoders, and answer decoder. A multimodal embedding $C$ combines both encodings into a joint space that the decoder decodes from.  See \autoref{sec:alternative_approaches} for details.
  }
  \label{fig:vqa_encoder_decoder}
\end{center}
\end{figure}

In the previous section, we have described how to achieve visual question answering with a single recurrent network for question and image encoding and answering. In this section, we abstract away from the particular design choices taken and describe a modular framework,  where a question encoder has to be combined with a visual encoder in order to produce answers with an answer decoder (\autoref{fig:vqa_encoder_decoder}). %
This modular representation allows us to systematically investigate and evaluate a range of design choices of different encoders, multimodal embeddings, and decoders. 

\subsubsection{Question encoders}

\begin{figure}[t]
\begin{center}
  \includegraphics[width=0.9\columnwidth]{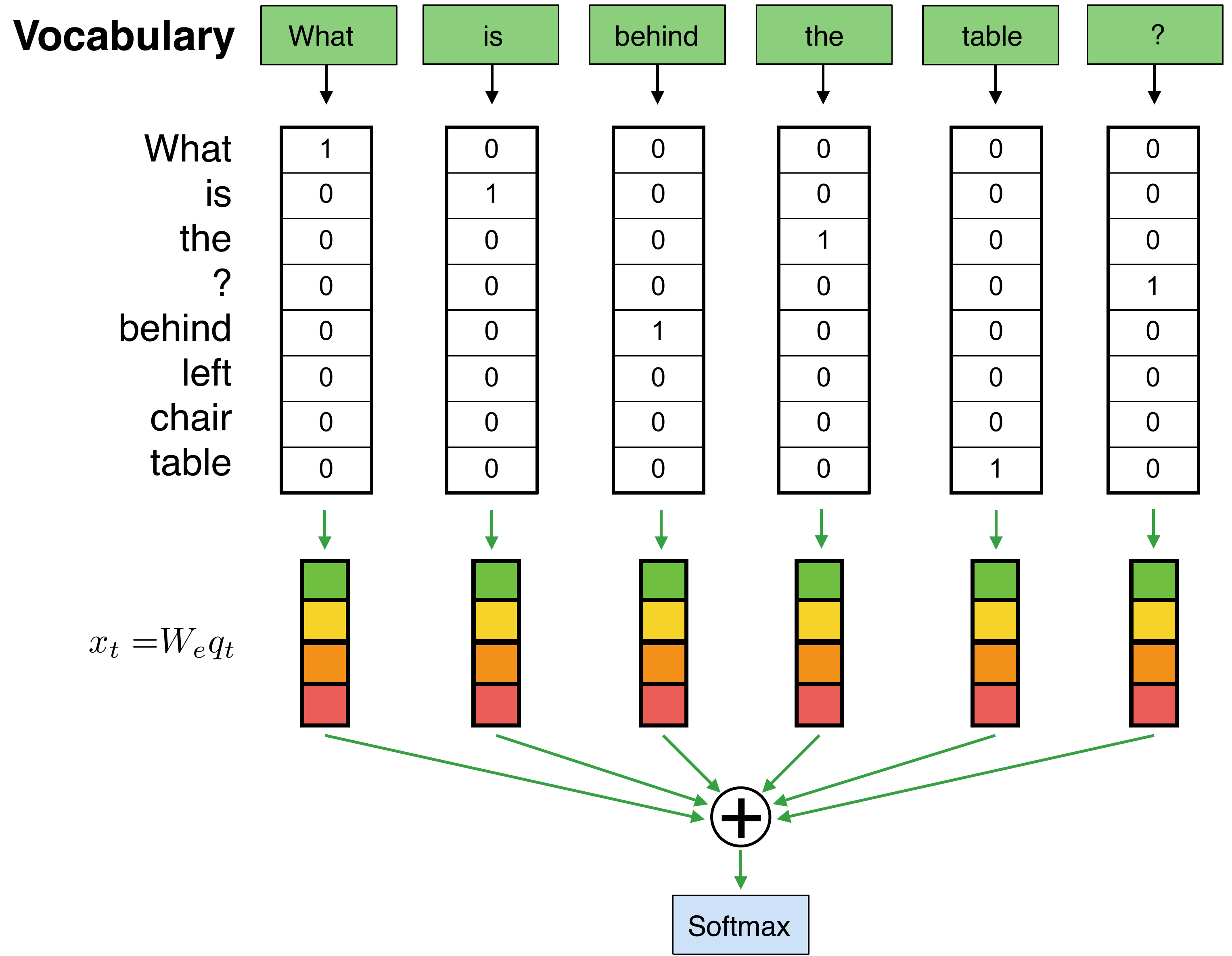}
  \caption{Bag-Of-Words (BOW) for encoding the question,  see \autoref{sec:bow} for details.}
  \label{fig:bow-approach}
\end{center}
\end{figure}

The main goal of a question encoder is to capture a meaning of the question, which we write here as $\qenc{}$.
Such an encoder can range from a very structured ones like Semantic Parser used in \citet{malinowski14nips} and \citet{liang2013learning} that explicitly model compositional nature of the question, to orderless Bag-Of-Word (BOW) approaches that merely compute a histogram over the question words (\autoref{fig:bow-approach}).
In this work, we investigate a few encoders within such spectrum that are compatible with the proposed Deep Learning approach:
Two recurrent question encoders, LSTM \citep{hochreiter97nc} (see \autoref{sec:LSTM}) and GRU \citep{cho2014learning}, that assume a temporal ordering in questions, as well as the aforementioned BOW.

\paragraph{Gated Recurrent Unit (GRU).}
GRU is a simplified variant of LSTM that also uses gates (a reset gate $\bs{r}$ and an update gate $\bs{u}$) in order to keep long term dependencies. GRU is expressed by the following set of equations:
\begin{align}
\bs{r}_t &= \sigma(W_{vr}\bs{v}_t + W_{hr}\bs{h}_{t-1} + \bs{b}_r)\\
\bs{u}_t &= \sigma(W_{vu}\bs{v}_t + W_{hu}\bs{h}_{t-1} + \bs{b}_u) \\
\bs{c}_t &= W_{vc}\bs{v}_t + W_{hc} (\bs{r}_t \odot \bs{h}_{t-1}) + \bs{b}_c \\
\bs{h}_t &= \bs{u}_t \odot \bs{h}_{t-1} + (\bs{1} - \bs{u}_t) \odot \phi(\bs{c}_t)
\end{align}
where $\sigma$ is the sigmoid function, $\phi$ is the hyperbolic tangent, and $\bs{v}_t$, $\bs{h}_t$ are input and hidden state at time $t$. The representation of the question $\bs{q}$ is the hidden vector at last time step, \ie $ \qenc{RNN} := \bs{h}_T$.

\begin{figure}[t]
\begin{center}
  \includegraphics[width=0.9\columnwidth]{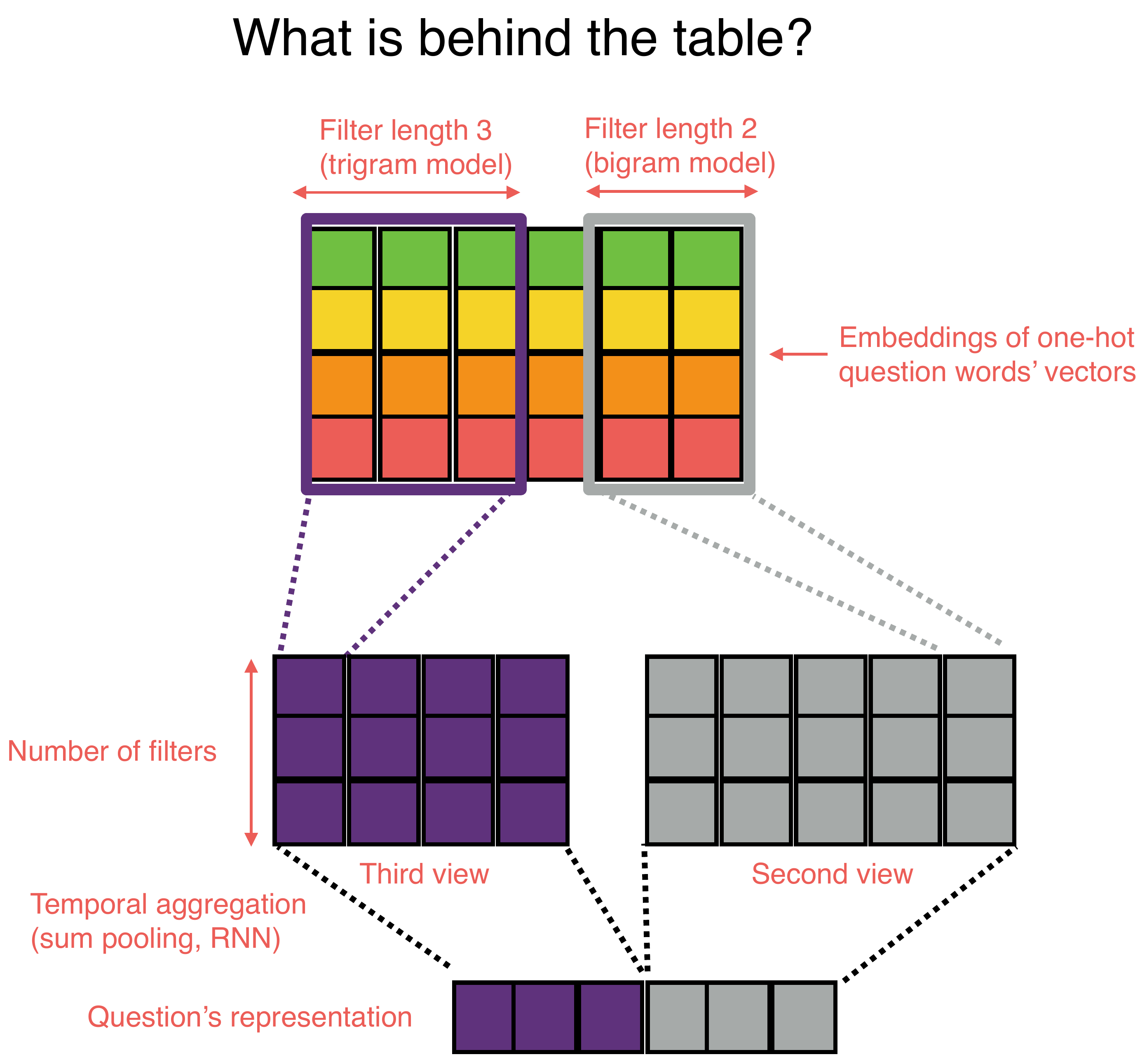}
  \caption{CNN for encoding the question that convolves word embeddings (learnt or pre-trained) with different kernels, second and third views are shown,  see \autoref{sec:question_cnn} and \citet{yang2015stacked} for details.}
  \label{fig:cnn_lang-approach}
\end{center}
\end{figure}

\paragraph{Bag-Of-Word (BOW).}
\label{sec:bow}
Conceptually the simplest encoder, the BOW  approach (\autoref{fig:bow-approach}) sums over the words embeddings:
\begin{align}
\label{eq:bow_representation}
\qenc{BOW} := \sum_{t}^{n} W_e(\bs{q}_t).
\end{align}
where $\bs{W}_e$ is a matrix and $\bs{q}_t$ is one-hot binary vector of the word with exactly one $1$ pointing to a place of the 'word' in the vocabulary (\autoref{fig:bow-approach}).
BOW does not encode the ordering of words in the question, so that especially questions with swapped arguments of spatial prepositions become indistinguishable, \ie\\
$\Psi_{\text{BOW}}(\text{red chair left of sofa}) =  \Psi_{\text{BOW}}(\text{red sofa left of chair} )$
in the BOW sentence representation.

\paragraph{Convolutional Neural Network (CNN).}
\label{sec:question_cnn}
 Convolutional Neural Networks (CNN) have been proposed to encode language \citep{kim2014convolutional,kalchbrenner2014convolutional,learning_to_answer_questions,yang2015stacked} and since have shown to be fast to compute and result in good accuracy. Since they consider a larger context, they arguably maintain more structure than BOW, but do not model such long term dependencies as recurrent neural networks. \autoref{fig:cnn_lang-approach} depicts our CNN architecture, which is very similar to \citet{learning_to_answer_questions} and \citet{yang2015stacked}, that convolves word embeddings with three convolutional kernels of length $1$, $2$ and $3$. For the sake of clarity, we only show two kernels in the figure. We either learn them jointly with the whole model or use GLOVE \citep{pennington2014glove} in our experiments. We call such architecture with $1$, ..., $n$ kernel lengths $n$ views CNN. At the end, the kernel's outputs are temporarily aggregated for the final question's representation. We use either sum pooling or a recurrent neural network (CNN-RNN) to accomplish this step.
\subsubsection{Visual encoders}
The second important component of the encoder-decoder architectures
is the visual representation. Convolutional Neural Networks (CNNs) have become the state-of-the-art framework that provide features from images. The typical protocol of using the visual models is to first pre-train them on the ImageNet dataset \citep{ILSVRCarxiv14}, a large scale recognition dataset, and next use them as an input for the rest of the architecture. Fine-tuning the weights of the encoder to the task at hand is also possible.
In our experiments, we use chronologically the oldest CNN architecture fully trained on ImageNet -- a Caffe implementation of AlexNet \citep{jia2014caffe,krizhevsky2012imagenet} -- as well as the recently introduced deeper networks -- Caffe implementations of GoogLeNet and VGG \citep{szegedy2014going, simonyan2014very} -- to the most recent extremely deep architectures -- a Facebook implementation of $152$ layered ResidualNet \citep{he2015deep}. As can be seen from our experiments in \autoref{section:analysis_on_vqa}, a strong visual encoder plays an important role in the overall performance of the architecture.
\subsubsection{Multimodal embedding}
The presented neural question encoders transform linguistic question into a vector space. Similarly visual encoders encode images as vectors. A multimodal fusion module combines both vector spaces into another vector based on which the answer is decoded.
Let $\qenc{}$ be a question representation (BOW, CNN, LSTM, GRU), and $\ienc{}$ be a representation of an image. Then $C(\qenc{} ,\ienc{})$ is a function which embeds both vectors.
In this work, we investigate three multimodal embedding techniques: Concatenation, element-wise multiplication, and summation. Since the last two techniques require compatibility in the number of feature components, we use additional visual embedding matrix $W_{ve} \in \mathbb{R}^{|\qenc{}| \times |\ienc{}|}$.
Let $W$ be weights of an answer decoder. Then we have $W C(\qenc{}, \ienc{})$, which is
\begin{align}
  W_{q} & \qenc{} + W_{v} \ienc{} \label{eq:concat_fusion}\\
  W (&\qenc{} \odot W_{ve} \ienc{}) \label{eq:piecewise_mult_fusion} \\
  W &\qenc{} + W W_{ve} \ienc{} \label{eq:summation_fusion}
\end{align}
in concatenation, element-wise multiplication, and summation fusion techniques respectively. In \autoref{eq:concat_fusion}, we decompose $W$ into two matrices $W_q$ and $W_v$, that is $W = \left[W_q; W_v\right]$. In \autoref{eq:piecewise_mult_fusion}, $\odot$ is an element-wise multiplication. Similarity between \autoref{eq:concat_fusion} and \autoref{eq:summation_fusion} is interesting as the latter is the former with weight sharing and additional decomposition into $W W_{ve}$.

\subsubsection{Answer decoders}

We consider two approach to decode the internal representation of our model into an answer.

\paragraph{Answer words generation.}
The last component of our architecture %
(\autoref{fig:vqa_encoder_decoder}) is an answer decoder. %
Inspired by the work on the image description task \citep{donahue15cvpr}, we uses an LSTM as decoder that shares the parameters with the encoder.

\paragraph{Classification.}
As alternative, we can  cast  the answering problem as a classification task, with answers as different classes. This approach has been widely explored, especially on VQA \citep{antol2015vqa}. 
\section{Analysis on DAQUAR}\label{sec:results}
In this section, we benchmark our method on a task of answering questions about images. We compare different variants of our proposed model to prior work in Section \ref{sec:experiments:eval}. In addition, in Section \ref{sec:experiments:evalNoImg}, we analyze how well questions can be answered without using the image in order to gain an understanding of biases in form of prior knowledge and common sense.
 We provide a new human baseline for this task.
In Section \ref{sec:experiments:humanConsensus} we discuss ambiguities in the question answering tasks and analyze them further by introducing metrics that are sensitive to these phenomena. In particular, the WUPS score \citep{malinowski14nips} is extended to a consensus metric that considers multiple human answers. All the material is available on our project webpage \footnote{\url{http://mpii.de/visual_turing_test}}.

\paragraph{Experimental protocol.}
We evaluate our approach from \autoref{sec:method} on the DAQUAR dataset \citep{malinowski14nips}, which provides $12,468$ human question answer pairs on images of indoor scenes \citep{silbermanECCV12} and follow the same evaluation protocol by providing results on  accuracy and the WUPS score at $\left\{0.9,0.0\right\}$.
We run experiments for the full dataset as well as their proposed reduced set that restricts the output space to only $37$ object categories and uses $25$ test images. In addition, we also evaluate the methods on different subsets of DAQUAR where only $1$, $2$, $3$ or $4$ word answers are present.

We use default hyper-parameters of LSTM \citep{donahue15cvpr} and CNN \citep{jia2014caffe}. All CNN models are first pre-trained on the ImageNet dataset \citep{ILSVRCarxiv14}, and next we randomly initialize and train the last layer together with the LSTM network on the task. We find this step crucial to obtain good results.
We have explored the use of a 2 layered LSTM model, but have consistently obtained worse performance.
In a pilot study, we have found that \textit{GoogleNet} architecture \citep{jia2014caffe,szegedy2014going} consistently outperforms the \textit{AlexNet} architecture \citep{jia2014caffe,krizhevsky2012imagenet} as a CNN model for our task and model. 

\begin{table}
\begin{center}
\begin{tabular}{lrrr}
\toprule
 & Accu- & WUPS & WUPS\\
 &racy&@0.9&@0.0\\
 \cmidrule(lr){1-1}\cmidrule(lr){2-4}
  \cite{malinowski14nips}  & $7.86$&$11.86$&$38.79$\\
  \cmidrule(lr){1-1}\cmidrule(lr){2-4}

\AproachName (ours)\\
\ - multiple words& $17.49$&$23.28$&$57.76$ \\
\ - single word& $\boldsymbol{19.43}$&$\boldsymbol{25.28}$&$\boldsymbol{62.00}$\\
Human answers \citeyear{malinowski14nips}  & $50.20$&$50.82$&$67.27$\\

 \cmidrule(lr){1-1}\cmidrule(lr){2-4}

  Question-only (ours)\\ 
  \ - multiple words& $17.06$&$22.30$&$56.53$  \\
\ - single word& $17.15$&$22.80$&$58.42$ \\
 Human answers, no images & $7.34$ & $13.17$ & $35.56$ \\

\bottomrule
\end{tabular}
\end{center}
\caption{
Results on DAQUAR, all classes, single reference, in \%.
}
\label{table:full_daquar}
\end{table}

\paragraph{WUPS scores.}
We base our experiments and the consensus metrics on WUPS scores \citep{malinowski14nips}. The metric is a generalization of the accuracy measure that accounts for word-level ambiguities in the answer words. For instance `carton' and `box' can be associated with a similar concept, and hence models should not be strongly penalized for this type of mistakes. Formally:

\begin{align*}
  \textrm{WUPS}(A,T) = \frac{1}{N} \sum_{i=1}^N\min\{ \prod_{a \in A^i} \max_{t\in T^i} \mu(a, t) ,\; \\ \prod_{t \in T^i} \max_{a \in A^i} \mu(a, t)
  \}
\end{align*}
To embrace the aforementioned ambiguities, \citet{malinowski14nips} suggest using a thresholded taxonomy-based Wu-Palmer similarity \citep{wu1994verbs} for $\mu$. Smaller thresholds yield more forgiving metrics. As in \citet{malinowski14nips}, we report WUPS at two extremes, $0.0$ and $0.9$.

\subsection{Evaluation of \AproachName}
\label{sec:experiments:eval}
We start with the evaluation of our \AproachName on the full DAQUAR dataset in order to study different variants and training conditions. Afterwards we evaluate on the reduced DAQUAR for additional points of comparison to prior work.

\paragraph{Results on full DAQUAR.}
\autoref{table:full_daquar} shows the results of our \AproachName method on the full set (``multiple words'') with $653$ images and $5673$ question-answer pairs available at test time. In addition, we evaluate a variant that is trained to predict only a single word (``single word'') as well as a variant that does not use visual features (``Question-only''). Note, however, that ``single word'' refers to a training procedure. All the methods in \autoref{table:full_daquar} are evaluated on the full DAQUAR dataset at test time that also contains multi-word answers.
In comparison to the prior work \citep{malinowski14nips} (shown in the first row in \autoref{table:full_daquar}), we observe strong improvements of over $9\%$ points in accuracy and over $11\%$ in the WUPS scores (second row in \autoref{table:full_daquar} that corresponds to ``multiple words''). Note that, we achieve this improvement despite the fact that the only published number available for the comparison
 on the full set uses ground truth object annotations \citep{malinowski14nips} -- which puts our method at a disadvantage.
Further improvements are observed when we train only on a single word answer, which doubles the accuracy obtained in prior work.
We attribute this to a joint training of the language and visual representations
and the dataset bias, where about $90\%$ of the answers contain only a single word.

We further analyze this effect in \autoref{fig:exactly_n_words}, where we show performance of our approach (``multiple words'') in dependence on the number of words in the answer (truncated at 4 words due to the diminishing performance). The performance of the ``single word'' variants on the one-word subset
are shown as horizontal lines. Although accuracy drops rapidly for longer answers, our model is capable of producing a significant number of correct two words answers. The ``single word'' variants have an edge on the single answers and benefit from the dataset bias towards these type of answers. Quantitative results of the ``single word'' model on the one-word answers subset of DAQUAR are shown in  \autoref{table:subset_single_word}.
While we have made substantial progress compared to prior work, there is still a $30\%$ points margin to human accuracy and $25$ in WUPS score (``Human answers'' in \autoref{table:full_daquar}).

Later on, in \autoref{sec:soa}, we will show improved results on DAQUAR with a stronger visual model and a pre-trained word embedding, with ADAM \citep{kingma2014adam} as the chosen optimization technique. We also put the method in a broader context, and compare with other approaches. 
\begin{table}
\begin{center}
\begin{tabular}{lrrr}
\toprule
 & Accu- & WUPS & WUPS\\
 &racy&@0.9&@0.0\\
 \cmidrule(lr){1-1}\cmidrule(lr){2-4}

\AproachName (ours)& $21.67$&$27.99$&$65.11
$\\
 \cmidrule(lr){1-1}\cmidrule(lr){2-4}
  Question-only (ours)& $19.13$ & $25.16$ & $61.51$ \\
\bottomrule
\end{tabular}
\end{center}
\caption{Results of the single word model on the one-word answers subset of DAQUAR, all classes, single reference, in $\%$.
}
\label{table:subset_single_word}
\end{table}

\pgfplotsset{
compat=1.5,
ymax=30
}
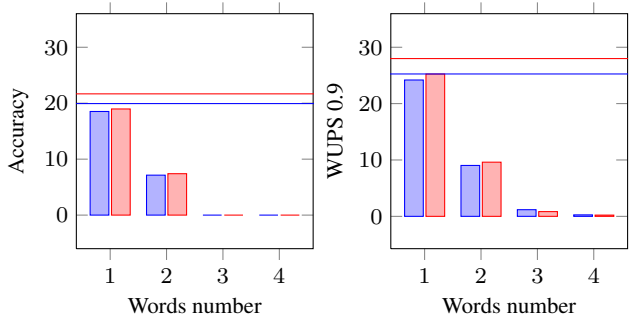
\begin{figure}
\hspace{-0.55cm}
\begin{tabular}{rl}
\begin{tikzpicture}
\begin{axis}[
  width=0.27\textwidth,
  height=0.27\textwidth,
	x tick label style={
		/pgf/number format/1000 sep=},
	ylabel=Accuracy,
  xlabel=Words number,
	enlargelimits=0.2,
  legend style={at={(0.5,-0.25)},
  anchor=north,legend columns=-1},
	ybar=1pt,
  bar width=7pt,
  xtick=data,
]
\addplot
	coordinates {(1,18.52) (2,7.14)
		 (3,0.0) (4,0.0)};

\addplot
	coordinates {(1,18.97) (2,7.40)
		(3,0.0) (4,0.0) };

\addplot[blue,sharp plot,update limits=false]
	coordinates {(0,19.93) (5,19.93)}
	node[below] at (axis cs:3,19.93) {};

\addplot[red,sharp plot,update limits=false]
	coordinates {(0,21.67) (5,21.67)}
	node[below] at (axis cs:3,21.67) {};

\end{axis}
\end{tikzpicture}
\begin{tikzpicture}
\begin{axis}[
  width=0.27\textwidth,
  height=0.27\textwidth,
	x tick label style={
		/pgf/number format/1000 sep=},
	ylabel=WUPS 0.9,
  xlabel=Words number,
	enlargelimits=0.2,
  legend style={at={(0.5,-0.25)},
    anchor=north,legend columns=-1},
	ybar=1pt,
  bar width=7pt,
  xtick=data,
]
\addplot
	coordinates {(1,24.19) (2,9.03)
		 (3,1.18) (4,0.26)};

\addplot
	coordinates {(1,25.25) (2,9.61)
		(3,0.84) (4,0.22) };

\addplot[blue,sharp plot,update limits=false]
	coordinates {(0,25.26) (5,25.26)}
	node[below] at (axis cs:3,25.26) {};

\addplot[red,sharp plot,update limits=false]
	coordinates {(0,27.99) (5,27.99)}
	node[below] at (axis cs:3,27.99) {};
\end{axis}
\end{tikzpicture}
\end{tabular}
\caption{Question-only (blue bar) and \AproachName (red bar) ``multi word'' models evaluated on different subsets of DAQUAR. We consider $1$, $2$, $3$, $4$ word subsets. The blue and red horizontal lines represent ``single word'' variants evaluated on the answers with exactly $1$ word.}
\label{fig:exactly_n_words}
\end{figure}
\paragraph{Results on reduced DAQUAR.}
In order to provide performance numbers that are comparable to the proposed Multi-World approach in \citet{malinowski14nips}, we also run our method on the reduced set with $37$ object classes and only $25$ images with $297$ question-answer pairs at test time.

\autoref{table:reduced_daquar} shows that \AproachName also improves on the reduced DAQUAR set, achieving $34.68\%$ Accuracy and $40.76\%$ WUPS at 0.9 substantially outperforming \citet{malinowski14nips} by  $21.95$ percent points of Accuracy and $22.6$ WUPS points. Similarly to previous experiments, we achieve the best performance using the ``single word'' variant of our method. Note that \citet{ren2015image}  reported $36.94\%$, $48.15\%$, and $82.68\%$ Accuracy, WUPS at 0.9, and WUPS at 0.0 respectively. However, they use another variant of our reduced DAQUAR dataset, where all the multiple word answers are removed. This roughly accounts for $98\%$ of the original reduced DAQUAR dataset. 

\subsection{Answering questions without looking at images}
\label{sec:experiments:evalNoImg}
In order to study how much information is already contained in questions, we train a version of our model that ignores the visual input.
The results are shown in \autoref{table:full_daquar} and \autoref{table:reduced_daquar} under ``Question-only (ours)''.
The best ``Question-only'' models with $17.15\%$ and $32.32\%$ compare very well in terms of accuracy to the best models that include vision. The latter achieve $19.43\%$ and $34.68\%$ on the full and reduced set respectively.

In order to further analyze this finding, we have collected a new human baseline ``Human answer, no image'', where we have asked participants to answer on the DAQUAR questions without looking at the images. It turns out that humans can guess the correct answer in
$7.86\%$
 of the cases by exploiting prior knowledge and common sense. Interestingly, our best ``Question-only'' model outperforms the human baseline by over $9$ percent points.
A substantial number of answers are plausible and resemble a form of common sense knowledge employed by humans to infer answers without having seen the image.

\subsection{Human Consensus}
\label{sec:experiments:humanConsensus}

\begin{table}
\begin{center}
\begin{tabular}{lrrr}
\toprule
 & Accu- & WUPS & WUPS\\
 &racy&@0.9&@0.0\\
 \cmidrule(lr){1-1}\cmidrule(lr){2-4}
  \citet{malinowski14nips}  & $12.73$&$18.10$&$51.47$\\
  \cmidrule(lr){1-1}\cmidrule(lr){2-4}
  \AproachName (ours)\\
\ - multiple words& $29.27$&$36.50$&$79.47$ \\
\ - single word& $\boldsymbol{34.68}$&$\boldsymbol{40.76}$&$79.54$\\
   \cmidrule(lr){1-1}\cmidrule(lr){2-4}
Question-only (ours)\\ 
  \ - multiple words& $32.32$&$38.39$&$80.05$  \\
\ - single word& $31.65$&$38.35$&$\boldsymbol{80.08}$ \\
\bottomrule
\end{tabular}
\end{center}
\caption{
Results on reduced DAQUAR, single reference, with a reduced set of $37$ object classes and $25$ test images with $297$ question-answer pairs, in $\%$ 
}
\label{table:reduced_daquar}
\end{table}

We observe that in many cases there is an inter human agreement in the answers for a given image and question and this is also reflected by the human baseline performance on the question answering task of $50.20\%$ (``Human answers'' in \autoref{table:full_daquar}).
We study and analyze this effect further by extending our dataset to multiple human reference answers in \autoref{sec:extended_consensus_annotation}, and proposing a new measure -- inspired by the work in psychology
 \citep{cohen1960coefficient,fleiss1973equivalence,nakashole2013fine} -- that handles agreement in \autoref{sec:consensus_measure}, as well as conducting additional experiments in \autoref{sec:consensus_results}.
 
\subsubsection{\daquarNew}\label{sec:extended_consensus_annotation}
In order to study the effects of consensus in the question answering task, we have asked multiple participants to answer the same question of the DAQUAR dataset given the respective image.
We follow the same scheme as in the original data collection effort, where the answer is a set of words or numbers. We do not impose any further restrictions on the answers.
This extends the original data \citep{malinowski14nips} to an average of $5$ test answers per image and question collected from $5$ in-house annotators. 
The annotators were first tested for their English proficiency so they would be able to accomplish the task. They were instructed verbally and were given the image and  entered an answer for a given question in a text editor.
 Regular quality checks were performed with a random set of question-answer-image triplets.
 We refer to this dataset as \daquarNew.

\subsubsection{Consensus Measures}\label{sec:consensus_measure}
While we have to acknowledge inherent ambiguities in our task, we seek a metric that prefers an answer that is commonly seen as preferred.
We make two proposals:

\paragraph{Average Consensus.}
We use our new annotation set that contains multiple answers per question in order to compute an expected score in the evaluation:
\begin{align}
\label{eq:consensus_metric}
\frac{1}{N K} \sum_{i=1}^N \sum_{k=1}^K \min\{ \prod_{a \in A^i} \max_{t\in T^i_k} \mu(a, t) ,\; \prod_{t \in T^i_k} \max_{a \in A^i} \mu(a, t)\}
\end{align}
where for the $i$-th question $A^i$ is the answer generated by the architecture and $T^i_k$ is the $k$-th possible human answer corresponding to the $k$-th interpretation of the question.
Both answers $A^i$ and $T^i_k$ are sets of the words, and $\mu$ is a membership measure, for instance WUP \citep{wu1994verbs}.
We call this metric
``Average Consensus Metric (ACM)'' since, in the limits, as $K$ approaches the total number of humans, we truly measure the inter human agreement of every question.

\begin{figure}
\begin{tabular}{rl}
\begin{tikzpicture}
\begin{axis}[
  width=0.25\textwidth,
  height=0.25\textwidth,
	x tick label style={
		/pgf/number format/1000 sep=},
  ylabel=Fraction of data,
  xlabel=Human agreement,
	enlargelimits=0.2,
  legend style={at={(0.5,-0.25)},
    anchor=north,legend columns=-1},
	ybar=1pt,
  bar width=7pt,
  xtick=data,
  ymax=100,
]
\addplot
  coordinates {(0,19.65) (50,56.58) (100, 15.27) };
\end{axis}
\end{tikzpicture}
\begin{tikzpicture}
\begin{axis}[
  width=0.25\textwidth,
  height=0.25\textwidth,
	x tick label style={
		/pgf/number format/1000 sep=},
  ylabel=Fraction of data,
  xlabel=Human agreement,
	enlargelimits=0.2,
  legend style={at={(0.5,-0.25)},
    anchor=north,legend columns=-1},
	ybar=1pt,
  bar width=7pt,
  xtick=data,
  ymax=100,
]
\addplot
  coordinates {(0,19.5) (50,37) (100, 15.35) };
\end{axis}
\end{tikzpicture}
\end{tabular}
\caption{Study of inter human agreement. At $x$-axis: no consensus ($0\%$), at least half consensus ($50\%$), full consensus ($100\%$). Results in $\%$. Left: consensus on the whole data, right: consensus on the test data.}
\label{fig:consensus_quality}
\end{figure}
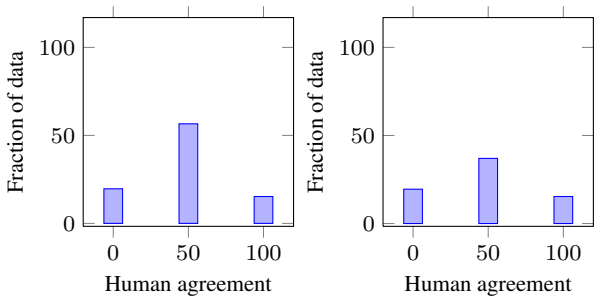
\paragraph{Min Consensus.}
The Average Consensus Metric puts more weights on more ``mainstream'' answers due to the summation over possible answers given by humans. In order to measure if the result was at least with one human in agreement, we propose a ``Min Consensus Metric (MCM)'' by replacing the averaging in \autoref{eq:consensus_metric} with a max operator.
We call such metric Min Consensus and suggest using both metrics in the benchmarks. We will make the implementation of both metrics publicly available.
\begin{align}
\label{eq:interpretation_metric}
\frac{1}{N} \sum_{i=1}^N \max_{k=1}^K \left( \min\{ \prod_{a \in A^i} \max_{t\in T^i_k} \mu(a, t) ,\; \prod_{t \in T^i_k} \max_{a \in A^i} \mu(a, t)\} \right)
\end{align}
Intuitively, the max operator uses in evaluation a human answer that is the closest to the predicted one -- which represents a minimal form of consensus.

\begin{table}
\begin{center}
\begin{tabular}{lrrr}
\toprule
 & Accu- & WUPS & WUPS\\
 &racy&@0.9&@0.0\\
 \cmidrule(lr){1-1}\cmidrule(lr){2-4}
   {\bf Subset: No agreement }\\
     Question-only (ours)\\ 
     \ - multiple words& $8.86$&$12.46$&$38.89$  \\
   \ - single word& $8.50$&$12.05$&$40.94$ \\
    \cmidrule(lr){1-1}\cmidrule(lr){2-4}
   \AproachName (ours)\\
   \ - multiple words& $\boldsymbol{10.31}$&$\boldsymbol{13.39}$&$40.05$ \\
   \ - single word&  $9.13$&$13.06$&$\boldsymbol{43.48} $
   \\
 \bottomrule
  {\bf Subset: $\ge 50\%$ agreement }\\
    Question-only (ours)\\ 
    \ - multiple words& $21.17$&$27.43$&$66.68$  \\
  \ - single word& $20.73$&$27.38$&$67.69$ \\
   \cmidrule(lr){1-1}\cmidrule(lr){2-4}
  \AproachName (ours)\\
  \ - multiple words& $20.45$&$27.71$&$67.30$ \\
  \ - single word&  $\boldsymbol{24.10}$&$\boldsymbol{30.94}$&$\boldsymbol{71.95} $
  \\
  \bottomrule
   {\bf Subset: Full Agreement} \\
     Question-only (ours)\\ 
     \ - multiple words& $27.86$&$35.26$&$78.83$  \\
   \ - single word& $25.26$&$32.89$&$79.08$ \\
    \cmidrule(lr){1-1}\cmidrule(lr){2-4}
   \AproachName (ours)\\
   \ - multiple words& $22.85$&$33.29$&$78.56$ \\
   \ - single word&  $\boldsymbol{29.62}$&$\boldsymbol{37.71}$&$\boldsymbol{82.31} $
   \\
   \bottomrule
\end{tabular}
\end{center}
\caption{
Results on DAQUAR, all classes, single reference in \% (the subsets are chosen based on \daquarNew).
}
\label{table:results_agreement_daquar}
\end{table}   

\subsubsection{Consensus results}\label{sec:consensus_results}
\begin{table}
\begin{center}
\begin{tabular}{lrrr}
\toprule
 & Accu- & WUPS & WUPS\\
 &racy&@0.9&@0.0\\
 \cmidrule(lr){1-1}\cmidrule(lr){2-4}
  {\bf Average Consensus Metric} \\
  Question-only (ours)\\ 
  \ - multiple words& $11.60$&$18.24$&$52.68$  \\
\ - single word& $11.57$&$18.97$&$54.39$ \\
 \cmidrule(lr){1-1}\cmidrule(lr){2-4}
\AproachName (ours)\\
\ - multiple words& $11.31$&$18.62$&$53.21$ \\
\ - single word& $\boldsymbol{13.51}$&$\boldsymbol{21.36}$&$\boldsymbol{58.03} $
\\
\bottomrule
 {\bf Min Consensus Metric}\\
  Question-only (ours)\\ 
  \ - multiple words& $22.14$&$29.43$&$66.88$  \\
\ - single word& $22.56$&$30.93$&$69.82$ \\
 \cmidrule(lr){1-1}\cmidrule(lr){2-4}
\AproachName (ours)\\
\ - multiple words& $22.74$&$30.54$&$68.17$ \\
\ - single word& $\boldsymbol{26.53}$&$\boldsymbol{34.87}$&$\boldsymbol{74.51} $\\
\bottomrule
\end{tabular}
\end{center}
\caption{
Results on \daquarNew, all classes, consensus in \%.
}
\label{table:aconsensus_daquar}
\end{table}

Using the multiple reference answers in \daquarNew we can show a more detailed analysis of inter human agreement. \autoref{fig:consensus_quality} shows the fraction of the data where the answers agree between all available questions (``100''), at least $50\%$ of the available questions and do not agree at all (no agreement - ``0''). We observe that for the majority of the data, there is a partial agreement, but even full disagreement is possible. We split the dataset into three parts according to the above criteria ``No agreement'', ``$\ge 50\%$ agreement'' and ``Full agreement'' and evaluate our models on these splits (\autoref{table:results_agreement_daquar} summarizes the results).
On subsets with stronger agreement, we achieve substantial gains of up to $10\%$ and $20\%$ points in accuracy over the full set (\autoref{table:full_daquar}) and the \textbf{Subset: No agreement} (\autoref{table:results_agreement_daquar}), respectively.
These splits can be seen as curated versions of DAQUAR, which allows studies with factored out ambiguities.

The aforementioned ``Average Consensus Metric'' generalizes the notion of the agreement, and encourages predictions of the most agreeable answers. On the other hand ``Min Consensus Metric'' has a desired effect of providing a more optimistic evaluation.  \autoref{table:aconsensus_daquar} shows the application of both measures to our data and models.

Moreover,  \autoref{table:consensus_human_baseline} shows that ``MCM'' applied to human answers at test time captures ambiguities in interpreting questions by improving the score of the human baseline from \citet{malinowski14nips}
(here, as opposed to \autoref{table:aconsensus_daquar}, we exclude the original human answers from the measure). It  cooperates well with WUPS at $0.9$, which takes word ambiguities into account, gaining 
 an $18\%$ higher score.%

\subsection{Qualitative results}
\vspace{-0.1cm}
We show predicted answers of different architecture variants in Tables \ref{fig:vision_vs_language}, \ref{fig:multiple_answers}, and \ref{fig:mix_predictions}.
We chose the examples to highlight differences between \AproachName and the ``Question-only''.
We use a ``multiple words'' approach only in \autoref{fig:multiple_answers}, otherwise the ``single word'' model is shown. Despite some failure cases, ``Question-only'' makes ``reasonable guesses'' like predicting that the largest object could be table or an object that could be found on the bed is a pillow or doll.

\begin{table}
\center
\scalebox{0.95}{
\begin{tabular}{lrrr}
\toprule
 & Accuracy & WUPS & WUPS\\
 &&@0.9&@0.0\\
 \cmidrule(lr){1-1}\cmidrule(lr){2-4}
  WUPS \citep{malinowski14nips}  & $50.20$&$50.82$&$67.27$\\
  \cmidrule(lr){1-1}\cmidrule(lr){2-4}
  ACM (ours) &$36.78$ & $45.68$ & $64.10$ \\
  MCM (ours) &$60.50$ & $69.65$ & $82.40$ \\
\bottomrule

\end{tabular}
}
\caption{
Min and Average Consensus on human answers from DAQUAR, as reference sentence we use all answers in \daquarNew which are not in DAQUAR, in $\%$ 
}
\label{table:consensus_human_baseline}
\end{table}
\subsection{Failure cases}
While our method answers correctly on a large part of the challenge (e.g. $\approx 35 $ WUPS at $0.9$ on ``what color'' and ``how many'' question subsets),  spatial relations ($\approx 21$ WUPS at $0.9$) which account for a substantial part of DAQUAR remain challenging.  Other errors involve questions with small objects, negations, and shapes (below $12$ WUPS at $0.9$). Too few training data points for the aforementioned cases may contribute to these mistakes.
\autoref{fig:mix_predictions} shows examples of failure cases that include (in order) strong occlusion, a possible answer not captured by our ground truth answers, and unusual instances (red toaster).

\subsection{Common Sense Knowledge}
\label{sec:common_sense}
\begin{figure}[t]
\begin{center} \includegraphics[width=0.8\linewidth]{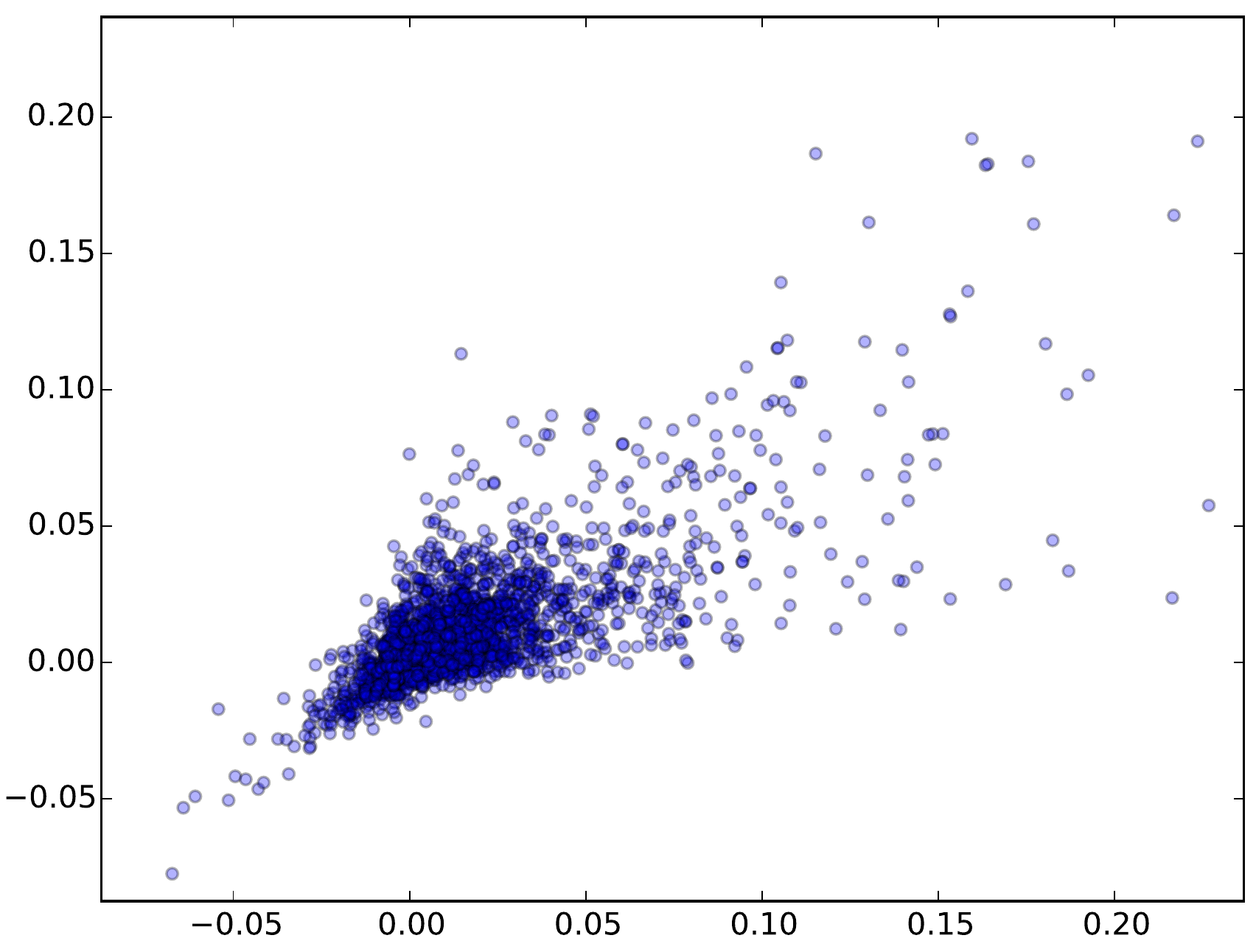}
\end{center}
\caption{
Figure showing correlation between question and answer words of the ``Question-only'' model (at x-axis), and a similar correlation of the ``Human-baseline'' \citep{malinowski14nips} (at y-axis).
}
\label{fig:correlation_plots}
\end{figure}
 Although ``Question-only'' ignores the image, it is still able to make ``reasonable guesses''
 by exploiting  biases captured by the dataset.
Some of such biases we interpret as a type of a common sense knowledge.
  For instance,
 ``tea kettle'' often sits on the oven, cabinets are usually ``brown'', ``chair'' is typically placed in front of a table, and we commonly keep a ``photo'' on a cabinet (\autoref{fig:compound_answers}, \ref{fig:colors}, \ref{fig:objects1}, \ref{fig:objects4}).
On the other hand, some other biases are hardly related to  the common sense knowledge. For instance, ``11'' as the answer to the question ``How many bottles are on the desk?'' or a ``clock'' as the answer to the question ``what is the black and white object on the top right of the brown board?''.
This effect is analysed in \autoref{fig:correlation_plots}. Each data point in the plot represents the correlation between a question and a predicted answer words for our ``Question-only'' model (x-axis) versus the correlation
in the human answers (y-axis). 
Despite the reasonable guesses of the ``Question-only'' architecture, the ``\AproachName'' predicts in average better answers (shown in \autoref{table:full_daquar}).
For instance in \autoref{fig:objects2}  the ``Question-only'' model incorrectly answers ``6'' on the question ``How many burner knobs are there ?'' because it has seen only this answer during the training with exactly the same question but on different image.

\section{Extended Experiments on DAQUAR} 
In this section, we first extend our experiments with other baseline methods, and next, guided by our findings on VQA shown in \autoref{sec:soa}, we show the results with the refined model in the context of a larger body of results on DAQUAR.
\begin{table*}
  \centering
  \begin{tabular}{cc}
  \toprule
\multicolumn{1}{c}{Type}  & \multicolumn{1}{c}{Regular expression} \\
 \cmidrule(l){1-2}
 color & what \textbackslash(is \textbackslash\textbackslash b\textbar the \textbackslash)\textbackslash?\textbackslash(the \textbackslash)\textbackslash?colo[u]?r \\
 count & how many \\
 size & \textbackslash blargest\textbackslash b\textbackslash\textbar\textbackslash bsmallest\textbackslash b\textbackslash\textbar\textbackslash blarge\textbackslash b\textbackslash\textbar\textbackslash bsmall\textbackslash b\textbackslash\textbar\textbackslash bbig\textbackslash b\textbackslash\textbar\textbackslash bbiggest\textbackslash b \\
 spatial & \textbackslash b\textbackslash bfront\textbackslash b\textbackslash\textbar\textbackslash bleft\textbackslash b\textbackslash\textbar\textbackslash bright\textbackslash b\textbackslash\textbar\textbackslash bbelow\textbackslash b\textbackslash\textbar\textbackslash babove\textbackslash b\textbackslash\textbar\textbackslash bbeneath\textbackslash b
 \\ 
&\textbackslash\textbar\textbackslash bunder\textbackslash  b\textbackslash\textbar\textbackslash bbehind\textbackslash b\textbackslash\textbar\textbackslash bbeside\textbackslash b\textbackslash\textbar\textbackslash bacross\textbackslash b\textbackslash\textbar\textbackslash bahead\textbackslash b\textbackslash\textbar\textbackslash baround\textbackslash b \\ 
 other & - \\
 \bottomrule
\end{tabular}
\caption{
We start with a regular expression from the top of this table to define the question type and extract the corresponding subset of  questions. Next, such a subset is removed from the set of all questions to form the remaining set. Subsequently, we replace the whole set with the remaining set, and continue the procedure. Therefore, a subset of questions corresponding to the  ``count'' type does not contain the ``color'' question. We will provide the splits in our project web-page under: \url{ http://mpii.de/visual_turing_test}.
}
\label{table:question_types}
\end{table*}
\begin{table*}
  \centering
  \begin{tabular}{lccc}
  \toprule
& \multicolumn{1}{c}{Accuracy}  & \multicolumn{1}{c}{WUPS@0.9} & \multicolumn{1}{c}{WUPS@0 on subset} \\
\cmidrule(l){1-4} %
 \cmidrule(l){1-1}
 \iccvArch \citep{malinowski2015ask} & 19.43 & 21.67 &	25.28  \\
Question-only of \citet{malinowski2015ask} & 17.15 & 22.80 &	58.42  \\
 \cmidrule(l){1-1}
 Constant & 4.97 & 7.13 & 26.89 \\
 Constant, per question type & 11.35 & 19.55 & 61.66 \\
 Look-up table & 15.44 & 18.45 & 36.87 \\
 Nearest Neighbor, Question-only & 13.15 & 18.39 & 51.80 \\
 Nearest Neighbor & 14.58 & 19.65 & 54.01 \\
 \bottomrule
\end{tabular}
\caption{
Results of various baselines on DAQUAR. All the results are shown on the whole dataset.
}
\label{table:daquar_baselines}
\end{table*}
\begin{table*}
  \centering
  \begin{tabular}{lcccccc}
  \toprule
&\multicolumn{2}{c}{Accuracy on subset}  & \multicolumn{2}{c}{WUPS@0.9 on subset} & \multicolumn{2}{c}{WUPS@0 on subset} \\
& all & single word & all & single word & all & single word \\
\cmidrule(l){1-7} %
 Global & & & & & & \\
 \cmidrule(l){1-1}
 \iccvArch \citep{malinowski2015ask} & 19.43 & 21.67 &	25.28 & 27.99 &	62.00 & 65.11 \\
 {\bf\newArch} & 24.48 & 26.67 &	29.78 & 32.55 &	62.80 & 66.25 \\
 {\bf\newArch$\ast$} & 25.74 & 27.26 &  31.00 & 33.25 &  63.14 & 66.79 \\
  IMG-CNN \citep{learning_to_answer_questions} & 21.47 & 24.49 &	27.15 & 30.47 &	59.44 & 66.08 \\
 & & & & & &  \\
 Attention & & & & & & \\
 \cmidrule(l){1-1}
 SAN (2, CNN) \citep{yang2015stacked} & - & $29.30$ & - & $35.10$ & - & $68.60$ \\
 DMN+ \citep{xiong16dynamic} & - & 28.79 & - & - & - & - \\
 ABC-CNN \citep{chen2015abc} & - & 25.37 & - & 31.35 & - & 65.89 \\
 Comp. Mem. \citep{jiang2015compositional} & 24.37 & - & 29.77 & - & 62.73 & - \\
 & & & & & &  \\
 Question prior & & & & & & \\
 \cmidrule(l){1-1}
 Hybrid & - & 28.96 & - & 34.74 & - & 67.33  \\
 \bottomrule
\end{tabular}
\caption{Comparison with state-of-the-art on DAQUAR. \newArch architecture: LSTM + Vision with GLOVE and ResNet-152. \iccvArch architecture: originally presented in \citet{malinowski2015ask}, results in \%. In the comparison, we use original data (all), or a subset with only single word answers (single word) that covers about $90\%$ of the original data. Asterisk `$\ast$' after the method denotes using a box filter that smooths the otherwise noisy validation accuracies. Dash `-' denotes lack of data.
}
\label{table:daquar_soa_results}
\end{table*}
\paragraph{Baseline methods.}
\label{sec:baselines:daquar}
To gain a better understanding of the effectiveness of our neural-based approach, we relate the obtained results to other baseline techniques. Similar baselines were also introduced in \citet{antol2015vqa} and \citet{ren2015image}, although they target either a different dataset or another variant of DAQUAR.
The \textit{Constant} technique uses the most frequent answer in the training set to answer to every question in the test set. In \textit{Constant, per question type}, we first break all questions into a few categories and then use the most frequent answer per category to answer to every question in that category at the test time. \autoref{table:question_types} provides more details on the chosen categories. \textit{Look-up table} builds a hash map from a textual question into the most frequent answer for that question at the training time. At the test time, the method just looks up the answer for the question in the hash-map. If the question exists then the most popular answer for that question is provided, otherwise an `empty' answer is given. In addition, we also remove articles, such as `the' and `a', from all the questions. However, this brings only a minor improvement. 
Finally, we experiment with two nearest-neighbor methods. Both methods rely on a Bag-of-Words representation of questions, where every question word is encoded by GLOVE \citep{pennington2014glove}. 
In the following, we call this the representation the semantic space. \textit{Nearest Neighbor, Question-only} searches at test time  for the most similar question in the semantic space  from the training set. Then it takes the answer that corresponds to this question. \textit{Nearest Neighbor} is inspired by a similar baseline introduced in \citet{antol2015vqa}. 
At the test time we first search for the $4$ most similar questions in the semantic space available in the training set. Next, we form candidate images that correspond to the aforementioned $4$ questions. At the last step, we choose an answer that is associated with the best match in the visual space. The latter is done by a cosine similarity between global CNN representations of the test image and every candidate image. %
We experiment with several CNN representations (VGG-19, GoogLeNet, ResNet-152)  but to our surprise there is little performance difference between them. We decide to use GoogLeNet as the results are slightly better.
\paragraph{Baseline results.}
\textit{Constant} shows how the dataset is biased w.r.t. the most frequent answer. This answer turns out to be the number ``$2$'', which also explain a good performance of \iccvArch on the ``how many'' question subset in DAQUAR. \textit{Constant, per question type} shows that question types provide quite strong clues for answering questions. \citet{kafle2016answer} take advantage of such clues in their Bayesian and Hybrid models (in \autoref{table:daquar_soa_results}, we only show a better performing Hybrid model). Our next baseline, \textit{Look-up table} can be seen as an extreme case of the \textit{Constant, per question type} model. It also gives surprisingly good results even though it cannot answer to novel questions. This result also confirms our intuitions that ``Question-only'' biases are important in the `question answering about images' datasets. Finally,  next nearest-neighbor baselines show that a visual representation still helps.

\paragraph{State-of-the-Art.}
\label{sec:soa:daquar}
Based on our further analysis on VQA (for more details we refer to \autoref{section:analysis_on_vqa} and \autoref{sec:soa}), 
we have also applied the improved model to DAQUAR, and  we significantly outperform \citet{malinowski2015ask} presented in \autoref{sec:results}. In the experiments, we first choose last $10\%$ of training set as a validation set in order to determine a number of training epochs $K$, and next we train the model for $K$ epochs. We evaluate model on two variants of DAQUAR: all data points (`all' in \autoref{table:daquar_soa_results}), and a subset (`single word' in \autoref{table:daquar_soa_results}) containing only single word answers, which consists of about $90\%$ of the original dataset. As \autoref{table:daquar_soa_results} shows, our model, Vision + Language with GLOVE and Residual Net that sums visual and question representations, outperforms the model of \citet{malinowski2015ask} by $5.05$, $4.50$, $0.80$
of Accuracy, WUPS at 0.9, and WUPS at 0.0 respectively. This shows how important a strong visual model is, as well as the aforementioned details used in training.
Likewise to our conclusions on VQA, we are also observing an improvement with attention based models (comparison in \textit{Attention} and \textit{Global} sections in \autoref{table:daquar_soa_results}).

\section{Analysis on VQA}
\label{section:analysis_on_vqa}

\newcommand{\resultsOnVQAval}{Results on VQA validation set}
\newcommand{\resultsOnVQAtest}{Results on VQA test set}

While \autoref{sec:results} analyses our original architecture \citep{malinowski2015ask} on the DAQUAR dataset, in this section, we analyze different variants and design choices for neural question answering on the large-scale Visual Question Answering (VQA) dataset \citep{antol2015vqa}. It is currently one of the largest and most popular visual question answering dataset with human question answer pairs.
In the following, after describing the experimental setup (\autoref{sec:vqa:setup}), we first describe several experiments which examine the different variants of question encoding, only looking at language input to predict the answer (\autoref{sec:vqa:setup}), and then, we examine the full model (\autoref{sec:vqa:vision_language}).

\subsection{Experimental setup}
\label{sec:vqa:setup}

We evaluate on the VQA dataset \citep{antol2015vqa}, which is built on top of the MS-COCO dataset \citep{lin2014microsoft}. Although VQA offers a different challenge tasks, we focus our efforts on the Real Open-Ended Visual Question Answering challenge. The challenge consists of $10$ answers per question with about $248k$  training questions, about $122k$  validation questions, and about $244k$  test questions.

As VQA consist mostly of single word answers (over 89\%), we treat the question answering problem as a classification problem of the most frequent answers in the training set. For the evaluation of the different model variants and design choices, we train on the training set and test on the validation set. Only the final evaluations (\autoref{table:vqa_test}) are evaluated on the test set of the VQA challenge, we evaluate on both parts test-dev and test-standard, where  the answers are not publicly available. As a performance measure we use a consensus variant of accuracy introduced in \citet{antol2015vqa}, where the predicted answer gets score between $0$ and $1$, with $1$ if it matches with at least three human answers. 
 We use ADAM \citep{kingma2014adam} throughout our experiments as we found out it performs better than SGD with momentum.
We keep default hyper-parameters for ADAM. Employed Recurrent Neural Networks maps input question into $500$ dimensional vector representation.
All the CNNs for text are using $500$ feature maps in our experiments, but the output dimensionality also depends on the number of views.
In preliminary experiments, we found that removing question mark `?' in the questions slightly improves the results, and we report the numbers only with this setting. Since VQA has $10$ answers associated with each question, we need to consider a suitable training strategy that takes this into account. We have examined the following strategies: (1) picking an answer  randomly,  (2) randomly but if possible annotated as confidently answered, (3) all answers, or (4) choosing the most frequent answer. 
In the following, we only report the results using the last strategy as we have found out little difference in accuracy between the strategies.
To allow training and evaluating many different models with limited time and computational power, we %
do not fine-tune the visual representations in these experiments, although our model would allow us to do so. All the models, which are publicly available together with our tutorial \citep{malinowski2016tutorial} 
\footnote{\url{https://github.com/mateuszmalinowski/visual_turing_test-tutorial}}, are implemented in Keras \citep{chollet2015} and Theano \citep{Bastien-Theano-2012}.

\subsection{Question-only}
\label{section:language_only}
\label{sec:vqa:language_only}

\begin{table}[t]
\center
\begin{tabular}{ccc}
\toprule
kernel length & single view & multi view \\
 $k$ &   $=k$ & $\leq k$\\
  \cmidrule(l){1-1}\cmidrule(r){2-3}
  $1$ & $47.43$ & $47.43$ \\
  $2$ & $48.11$ & $48.06$\\
  $3$ & $\boldsymbol{48.26}$ & $48.09$ \\
  $4$ & $\boldsymbol{48.27}$& $47.86$ \\
\bottomrule
\end{tabular}
\caption{\resultsOnVQAval, ``Question-only'' model:
Analysis of CNN questions encoders with different filter lengths, accuracy in \%, see \autoref{sec:vqa:lang:cnn} for discussion.
}
\label{table:vqa_cnn_filter_length}
\end{table}

\begin{table}[t]
\center
\begin{tabular}{lcc}
\toprule
Question & \multicolumn{2}{c}{Word embedding}\\
  encoder & learned & GLOVE \\
  \cmidrule(r){1-1}\cmidrule(lr){2-3}
  BOW & $47.41$ & $47.91$ \\ %
 CNN & $48.26$ &  $48.53 $  \\
 GRU & $47.60$ & $48.11$ \\
 LSTM & $47.80$  & $\boldsymbol{48.58}$\\
\bottomrule
\end{tabular}
\caption{\resultsOnVQAval, ``Question-only'' model:
Analysis of different questions encoders, accuracy in \%, see \autoref{sec:vqa:language_only} for discussion.
}
\label{table:question_encoders}
\end{table}
\begin{table}[t]
\center
\begin{tabular}{lccc}
\toprule
&  \multicolumn{3}{c}{top frequent answers}\\
  Encoder & 1000 & 2000 & 3000 \\
  \cmidrule(r){1-1}\cmidrule(lr){2-4}
  BOW & $47.91$& $48.13$ & $47.94$  \\
  CNN & $48.53$  & $48.67$ & $48.57$ \\
  LSTM & $48.58$  & $\boldsymbol{48.86}$ & $48.65$\\
\bottomrule
\end{tabular}
\center
\caption{\resultsOnVQAval, ``Question-only'' model:
Analysis of the number of top frequent answer classes, with different question encoders. All using GLOVE; accuracy in \%; see \autoref{sec:vqa:lang:top_answers} for discussion.
}
\label{table:top_frequent_words}
\end{table}

We start our analysis from ``Question-only'' models that do not use images to answer on questions. Note that the ``Question-only'' baselines play an important role in the question answering about images tasks since how much performance is added by vision. Hence, better overall performance of the model is not obscured by a better language model.
To understand better different design choices, we have conducted our analysis along the different `design' dimensions.

\subsubsection{CNN questions encoder}
\label{sec:vqa:lang:cnn}
We first examine different hyper-parameters for CNNs to encode the question. We first  consider the filter's length of the convolutional kernel. We run the model over different kernel lengths ranging from $1$ to $4$ (\autoref{table:vqa_cnn_filter_length}, left column). We notice that increasing the kernel lengths improves performance up to length $3$ were the performance levels out, we thus use kernel length $3$ in the following experiments for, such CNN can be interpreted as a trigram model.
We also tried to run %
simultaneously a few kernels with different lengths. In \autoref{table:vqa_cnn_filter_length} (right column) one view corresponds to a kernel length $1$, two views correspond to two kernels with length $1$ and $2$, three views correspond to length $1$, $2$ and $3$, etc. However, we find that the best performance still achieve with a single view and kernel length $3$ or $4$.

\subsubsection{BOW questions encoder}
Alternatively to neural network encoders, we consider Bag-Of-Words (BOW) approach where one-hot representations of the question words are first mapped to a shared embedding space, and subsequently summed over (\autoref{eq:bow_representation}), 
\ie $\Psi(\text{question}) := \sum_{\text{word}} W_e(word)$.
Surprisingly, such a simple approach gives very competitive results (first row in \autoref{table:question_encoders}) compared to the CNN encoding discussed in the previous section  (second row). %
\paragraph{Recurrent questions encoder.}
We examine two recurrent questions encoders, LSTM \citep{hochreiter97nc} and a simpler GRU \citep{cho2014learning}.
The last two rows of \autoref{table:question_encoders} show a slight advantage of using LSTM.%

\subsubsection{Pre-trained words embedding}
In all the previous experiments, we jointly learn the embedding transformation $\bs{W}_e$ together with the whole architecture only on the VQA dataset. This means we do not have any means for dealing with unknown words in questions at test time apart from using a special token $\left<\text{UNK}\right>$ to indicate such class.
To address such shortcoming, we investigate the pre-trained word embedding transformation GLOVE \citep{pennington2014glove} that encodes question words (technically it maps one-hot vector into a $300$ dimensional real vector). This choice naturally extends the vocabulary of the question words to about $2$ million words extracted a large corpus of web data -- Common Crawl \citep{pennington2014glove} --
that is used to train the GLOVE embedding.
Since the BOW architecture in this scenario becomes shallow (only classification weights are learnt),  we add an extra hidden layer between pooling and classification (without this embedding, accuracy drops by $5\%$). %
\autoref{table:question_encoders} (right column) summarizes our experiments with GLOVE. For all question encoders, the word embedding consistently improves performance which confirms that using a word embedding model learnt from a larger corpus helps. LSTM benefits most from GLOVE embedding, archiving the overall best performance with 48.58\% accuracy.

\begin{table}[t]
\center
\begin{tabular}{lcc}
\toprule
  & no norm & L2 norm  \\
  \cmidrule(l){1-1} \cmidrule(r){2-3}
 Concatenation &  $47.21$  & $52.39$ \\
 Summation     &  $40.67$  & $\boldsymbol{53.27}$ \\
 Element-wise multiplication & $49.50$ & $52.70$ \\
\bottomrule
\end{tabular}
\caption{\resultsOnVQAval, vision and language:
Analysis of different multimodal techniques that combine vision with language on BOW (with GLOVE word embedding and VGG-19 fc7), accuracy in \%, see \autoref{sec:vqa:fusion}.
}
\label{table:multimodal_fusion}
\end{table}

\begin{table}[t]
\center
\begin{tabular}{lc}
\toprule
 Method & Accuracy \\
  \cmidrule(l){1-1}\cmidrule(r){2-2}
 BOW & $53.27$  \\
 CNN & $54.23$ \\
 GRU & $54.23$ \\
 LSTM & $\boldsymbol{54.29}$ \\
\bottomrule

\end{tabular}
\caption{\resultsOnVQAval, vision and language:
Analysis of different language encoders with GLOVE word embedding, VGG-19, and Summation to combine vision and language. Results in \%, see \autoref{sec:vqa:vision:questionenc} for discussion.
}
\label{table:multimodal_methods}
\end{table}

\subsubsection{Top most frequent answers}
\label{sec:vqa:lang:top_answers}
Our experiments reported in \autoref{table:top_frequent_words} investigate predictions using different number of answer classes. We experiment with a truncation of $1000$, $2000$, or $4000$ most frequent classes. For all question encoders (and always using GLOVE word embedding), we find that a truncation at $2000$ words is best, being apparently a good compromise between answer frequency and missing recall.
\subsubsection{Summary Question-only}
We achieve the best ``Question-only'' accuracy with GLOVE word embedding, LSTM sentence encoding, and using the top $2000$ most frequent answers. This achieves an performance of 48.86\% accuracy.  In the remaining experiments, we use these settings for language and answer encoding.

\subsection{Vision and Language}
\label{sec:vqa:vision_language}

\begin{table}[t]
\center
\begin{tabular}{lc}
\toprule
 Method & Accuracy \\
  \cmidrule(l){1-1}\cmidrule(r){2-2}
 AlexNet & $ 53.69$  \\
 GoogLeNet & $54.52$ \\
 VGG-19 & $54.29$ \\
 ResNet-152 & $\boldsymbol{55.52}$ \\
\bottomrule

\end{tabular}
\caption{\resultsOnVQAval, vision and language:
Different visual encoders (with LSTM, GLOVE, the summation technique, l2 normalized features). Results in \%, see \autoref{sec:vqa:visual} for discussion.
}
\label{table:visual_encoders}
\end{table}

\begin{table}[t]
\center
\begin{tabular}{lcccc}
\toprule
 & \multicolumn{3}{c}{Question only} & + Vision\\
 \cmidrule(r){2-4}\cmidrule(r){5-5}
 & Learnt - & \multicolumn{3}{c}{GLOVE - word embedding} \\
 \cmidrule(r){2-2}\cmidrule(r){3-5}
Question encoding  $\downarrow$  & \multicolumn{2}{c}{Top 1000 answers} & \multicolumn{2}{c}{Top 2000 answers} \\
  \cmidrule(r){1-1}\cmidrule(r){2-3}\cmidrule(r){4-5}
 BOW & $47.41$  & $47.91$  & $48.13$  & $54.45$  \\
 CNN & $48.26$  & $48.53$  & $48.67$  & $55.34$  \\
 LSTM & $47.80$  & $48.58$  & $48.86$  & $\boldsymbol{55.52}$  \\
\bottomrule

\end{tabular}
\caption{\resultsOnVQAval, vision and language:
Summary of our results, results in \%, see \autoref{sec:vqa:summary} for discussion. Columns denote, from the left to right, word embedding learnt together with the architecture, GLOVE embedding that replaces learnt word embedding, truncating the dataset to $2000$ most frequent answer classes, and finally added visual representation to the model (\textit{ResNet-152}).}
\label{table:vqa_summary}
\end{table}

Although Question-only models can answer on a substantial number of questions as they arguably capture common sense knowledge, in order to address the full problem we will now also observing the image the question is based on.

\begin{table*}[t]
\center
\begin{tabular}{lcccccccc}
\toprule
  & \multicolumn{4}{c}{Test-dev} & \multicolumn{4}{c}{Test-standard} \\
 Trained on & Yes/No & Number & Other  & All & Yes/No & Number & Other  & All \\
  \cmidrule(r){1-1}\cmidrule(l){2-5}\cmidrule(l){6-9}
 Training set & 78.06 & 36.79& 44.59& $57.48$  & -&-&-&$57.55$  \\
 Training + Val set & 78.39 & 36.45 &46.28 &$\boldsymbol{58.39}$  & 78.24&36.27&46.32&$\boldsymbol{58.43}$  \\
\bottomrule

\end{tabular}
\caption{\resultsOnVQAtest, our best vision and language model chosen based on the validation set: accuracy in \%, from the challenge test server. Dash `-' denotes lack of data}
\label{table:vqa_test}
\end{table*}

\subsubsection{Multimodal fusion}
\label{sec:vqa:fusion}
\autoref{table:multimodal_fusion} investigates different techniques that combine visual and language representations. To speed up training, we combine the last unit of the question encoder with the visual encoder, as it is explicitly shown in \autoref{fig:vqa_encoder_decoder}. In the experiments we use concatenation, summation,
and element-wise multiplication on the BOW language encoder with GLOVE word embedding and features extracted from the VGG-19 net. In addition, we also investigate using L2 normalization of the visual features, which divides every feature vector by its L2 norm. The experiments show that the normalization is crucial in obtaining good performance, especially for Concatenation and Summation. In the remaining experiments, we use Summation.
\subsubsection{Questions encoders}
\label{sec:vqa:vision:questionenc}
\autoref{table:multimodal_methods} shows how well different questions encoders combine with the visual features. We can see that LSTM slightly outperforms two other encoders GRU and CNN, while BOW remains the worst, confirming our findings in our language-only experiments with GLOVE and 2000 answers (\autoref{table:top_frequent_words}, second column). %
\subsubsection{Visual encoders}
\label{sec:vqa:visual}
Next we fix the question encoder to LSTM and vary different visual encoders: Caffe variant of \textit{AlexNet} \citep{krizhevsky2012imagenet}, \textit{GoogLeNet} \citep{szegedy2014going}, \textit{VGG-19} \citep{simonyan2014very}, and recently introduced 152 layered \textit{ResNet} (we use the Facebook implementation of \citet{he2015deep}). \autoref{table:visual_encoders} confirms our hypothesis that stronger visual models perform better.
\subsubsection{Qualitative results}
We show predicted answers using our best model on VQA test set in Tables \ref{fig:vqa-image_qa-yes_no}, \ref{fig:vqa-image_qa-counting} ,\ref{fig:vqa-image_qa-what}, \ref{fig:vqa-image_qa-compound}. We show chosen examples with `yes/no', `counting', and `what'  questions, where our model, according to our opinion, makes valid predictions. Moreover, \autoref{fig:vqa-image_qa-compound} shows predicted compound answers.

\subsection{Summary VQA results}
\label{sec:vqa:summary}
\autoref{table:vqa_summary} summarises our findings on the validation set. We can see that on one hand methods that use contextual language information such as CNN and LSTM are performing better, on the other hand adding strong vision becomes crucial.
Furthermore, we use the best found models to run experiments on the VQA test sets: test-dev2015 and test-standard. To prevent overfitting, the latter restricts the number of submissions to 1 per day and 5 submissions in total. Here, we also study the effect of larger datasets where first we train only on the training set, and next we train for $20$ epochs on a joint, training and validation, set. When we train on the join set, we consider question answer pairs with answers among $2000$ the most frequent answer classes from the training and validation sets.  Training on the joint set
have gained us about $0.9\%$.  This implies that on one hand having more data indeed helps, but arguably we also need better models that exploit the current training datasets more effectively. Our findings are summarized in \autoref{table:vqa_test}.

\section{State-of-the-art on VQA}
\label{sec:soa}

\begin{table*}[t]
\center
\begin{tabular}{lcccccccc}
\toprule
  & \multicolumn{4}{c}{Test-dev} & \multicolumn{4}{c}{Test-standard} \\
  & Yes/No & Number & Other  & All & Yes/No & Number & Other  & All \\
  \cmidrule(r){1-1}\cmidrule(l){2-5}\cmidrule(l){6-9}
  SNUBI \citep{kim2016hadamard} & - & - & - & - & 84.6 & 39.1 & 57.8 & 66.9 \\
    MCB \citep{fukui16emnlp} &      83.4 & 39.8 & 58.5 &  66.7 & 83.2 & 39.5 & 58.0 & 66.5 \\ %
    DLAIT (unpublished) & 83.7 & 40.7 & 52.3 & 63.9 & 83.2 & 40.8 & 54.3 & 64.8 \\
    Naver Labs (unpublished)    & 83.5 & 39.8 & 54.8 & 64.9 & 83.3 & 38.7 & 54.6 & 64.8  \\
    POSTECH (unpublished) & - & - & - & - & 83.3 & 38.0 & 53.4 & 64.1 \\
    Brandeis \citep{prakashhighway} & 82.6 & 38.1 & 51.3 & 62.7 & 82.1 & 37.7 & 51.9 & 62.9 \\
   HieCoAtt \citep{lu2016hiecoatt} & 79.7 & 38.7 & 51.7 & 61.8 &  79.9 & 38.2 & 51.9 & 62.1  \\
   DualNet \citep{saito2016dualnet} & 82.0 & 37.9 & 49.2 & 61.5 & 82.0 & 37.6 & 49.7 & 61.8 \\
   klab \citep{kafle2016answer} & 81.7 & 39.1 & 49.2 & 61.5 & 81.5 & 39.3 & 49.6 & 61.7 \\
   SHB\_1026 (unpublished) & 82.3 & 37.0 & 47.7 & 60.7 & 82.1 & 36.8 & 47.8 & 60.8 \\
  DMN+ \citep{xiong16dynamic}      & 80.5 & 36.8 & 48.3 & 60.3 &  80.7 & 37.0 & 48.2 & 60.4  \\
   VT\_CV\_Jiasen (unpublished) & 80.5 & 38.5 & 47.5 & 60.1 & 80.6 & 38.1 & 47.9 & 60.3 \\ 
  FDA \citep{ilievski2016fda}      & 81.1 & 36.2 & 45.8 & 59.2 &  81.3 &  35.7 & 46.1 & 59.5 \\
  D-NMN \citep{andreas16naacl}     & 81.1 & 38.6 & 45.5 & 59.4 &  81.0 & 37.5 & 45.8 & 59.4 \\
  AMA \citep{wu16cvpr}             & 81.0 & 38.4 & 45.2 & 59.2 & 81.1 & 37.1 & 45.8 & 59.4  \\
  SAN \citep{yang2015stacked}      & 79.3 & 36.6 & 46.1 & 58.7 & 79.1 & 36.4 & 46.4 & 58.8 \\
NMN \citep{andreas16cvpr}          & 81.2 & 38.0 & 44.0 & 58.6 &  81.2 & 37.7 & 44.0 & 58.7   \\
  {\bf\newArch} & 78.4 & 36.4 &46.3 &58.4  & 78.2& 36.3 & 46.3 & 58.4  \\
    SMem \citep{xu2015ask}         & 80.9 & 37.3 & 43.1 & 58.0 & 80.8 & 37.5 & 43.5 & 58.2  \\
  VQA team \citep{antol2015vqa}    & 80.5 & 36.8 & 43.1 & 57.8 & 80.6 & 36.5 & 43.7 & 58.2  \\
  DPPnet \citep{noh2015images}     & 80.7 & 37.2 & 41.7 & 57.2 & 80.3 & 36.9 & 42.2 & 57.4 \\
  iBOWIMG \citep{zhou2015simple}   & 76.5 & 35.0 & 42.6 & 55.7 & 76.8 & 35.0 & 42.6 & 55.9  \\
  LSTM Q+I \citep{antol2015vqa} & 78.9 & 35.2 & 36.4 & 53.7 & - & - & - & 54.1\\
  Comp. Mem. \citep{jiang2015compositional}  & 78.3 & 35.9 & 34.5 & 52.7 & - & - & - & - \\
\bottomrule
\end{tabular}
\caption{\resultsOnVQAtest, comparison with state-of-the-art: accuracy in \%, from the challenge test server. Dash '-' denotes lack of data. The leaderboard can be found under \url{http://www.visualqa.org/roe.html}. Baselines from \citet{antol2015vqa} are not considered.}
\label{table:vqa_soa_test}
\end{table*}
In this section, we first put our findings on VQA in a broader context, where we compare our refined version of \textit{Ask Your Neurons} with other, publicly available, approaches. 
\autoref{table:vqa_soa_test} compares our \textit{Refined Ask Your Neurons} model with other approaches. Some methods, likewise to our approach, use global image representation, other attention mechanism, yet other dynamically predict question dependent weights, external textual sources, or fuse compositional  question's representation with neural networks.
\autoref{table:vqa_soa_test} shows a few trends.
First of all, a better visual representation significantly helps (\autoref{table:visual_encoders}). Most of the leading approaches to VQA also uses variants of ResNet, which is among the strongest approaches to the image classification task. It is, however, important to normalize the visual features (\autoref{table:multimodal_fusion}). Additionally, %
all the best models use an explicit attention mechanism (\eg \textit{DMN+}, \textit{FDA}, \textit{SAN}, \textit{POSTECH}, \textit{MCB}, \textit{HieCoAtt})

In this work, however, we focus on the extension of the plain ``Ask Your Neurons'' model that uses a global, full-frame image representation. A similar representation is used in  \textit{Refined Ask Your Neurons}, \textit{iBOWIMG}, \textit{VQA team}, and \textit{LSTM Q+I}. The best performing approaches also use different variants of the Recurrent Neural Networks (LSTM and GRU are the most popular). Such a question encoding outperforms Bag-of-Words representations (\textit{iBOWIMG} in \autoref{table:vqa_soa_test}, and
\textit{BOW} in \autoref{table:multimodal_methods}). As we hypothesize, a multimodal embedding plays an important role. This is not only shown in
 \autoref{table:multimodal_fusion}, but also emphasized in two leading approaches to VQA (\textit{MCB} and \textit{SNUBI}). Both methods use novel multimodal embedding techniques that build upon the element-wise multiplication.
  Finally, using external textual resources also seems to be beneficial (\textit{AMA}).

\begin{table*}[p!]
\begin{center}
\begin{tabular}{l@{\ }c@{\ }c@{\ }c}
\toprule
\multicolumn{2}{c}{\includegraphics[width=0.3\linewidth]{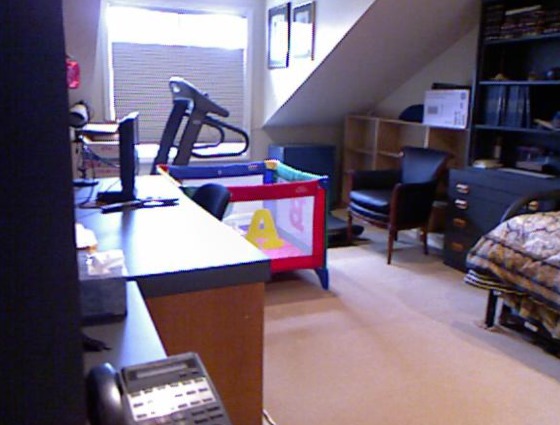}} &
\includegraphics[width=0.3\linewidth]{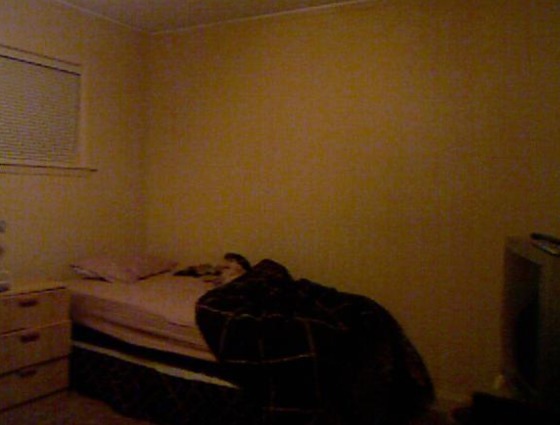} &
\includegraphics[width=0.3\linewidth]{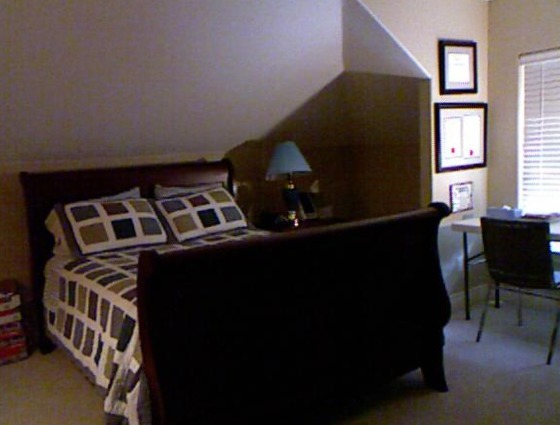} 
\\
\multicolumn{2}{c}{What is on the right side of the cabinet?} & \multicolumn{1}{c}{How many drawers are there?} & 
\multicolumn{1}{c}{What is the largest object?}
\\\midrule
\textit{\AproachName:}&\multicolumn{1}{c}{\textcolor{green}{bed}} & \multicolumn{1}{c}{\textcolor{green}{3}} & \multicolumn{1}{c}{\textcolor{green}{bed}}
\\\midrule
 \textit{Question-only:} &\multicolumn{1}{c}{\textcolor{green}{bed}} & 
\multicolumn{1}{c}{\textcolor{red}{6}} & 
\multicolumn{1}{c}{\textcolor{red}{table}}\\
\bottomrule
\end{tabular}
\end{center}
\caption{Examples of questions and answers on DAQUAR.  Correct predictions are colored in green, incorrect in red.}

\label{fig:vision_vs_language}
\end{table*}

\begin{table*}[tp]
\begin{center}
\begin{tabular}{l@{\ }c@{\ }c@{\ }c}
 \toprule
\multicolumn{2}{c}{\includegraphics[width=0.3\linewidth]{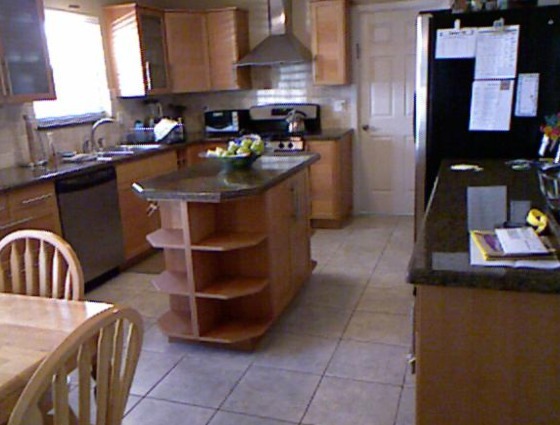}} &
\includegraphics[width=0.3\linewidth]{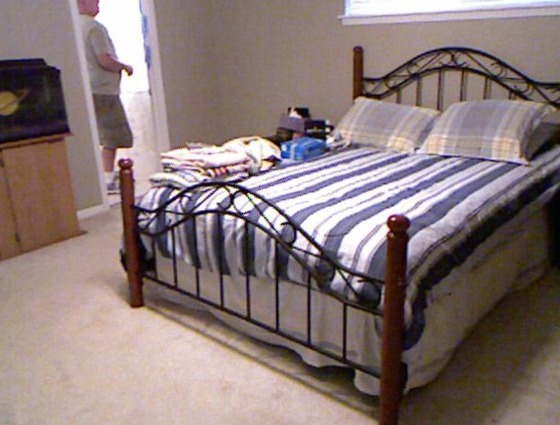} &
\includegraphics[width=0.3\linewidth]{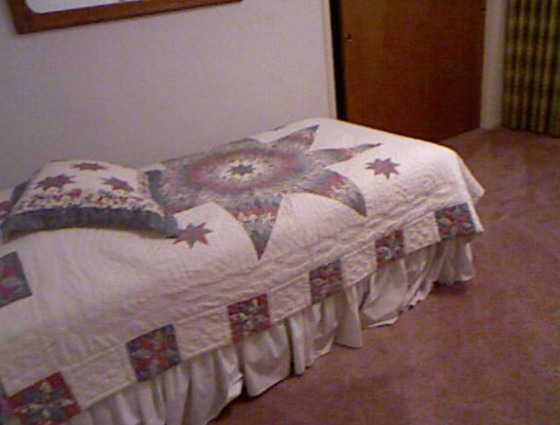} 
\\
\multicolumn{2}{c}{What is on the refrigerator?} & 
\multicolumn{1}{c}{What is the colour of the comforter?} & 
\multicolumn{1}{c}{What objects are found on the bed?}
\\\midrule
\textit{\AproachName:}&\multicolumn{1}{c}{\textcolor{green}{magnet, paper}} & 
\multicolumn{1}{c}{\textcolor{green}{blue, white}} & 
\multicolumn{1}{c}{\textcolor{green}{bed sheets, pillow}}
\\\midrule
\textit{Question-only:}&\multicolumn{1}{c}{\textcolor{green}{magnet, paper}} & 
\multicolumn{1}{c}{\textcolor{red}{blue, green, red, yellow}} & 
\multicolumn{1}{c}{\textcolor{red}{doll, pillow}}\\
\bottomrule
\end{tabular}
\end{center}
\caption{Examples of questions and answers on DAQUAR with multiple words. Correct predictions are colored in green, incorrect in red.
}

\label{fig:multiple_answers}
\end{table*}

\begin{table*}[tp]
\begin{center}
\begin{tabular}{l@{\ }c@{\ }c@{\ }c}
 \toprule
\multicolumn{2}{c}{\includegraphics[width=0.3\linewidth]{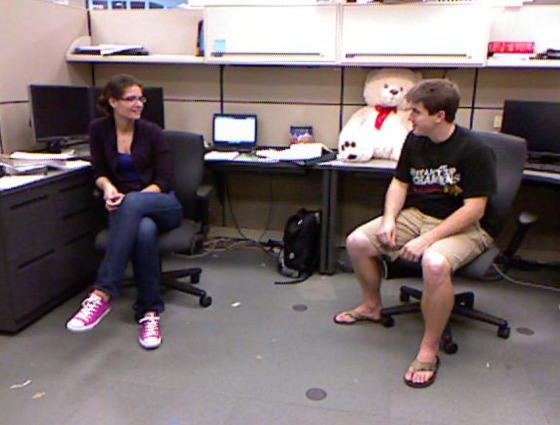}} &
\includegraphics[width=0.3\linewidth]{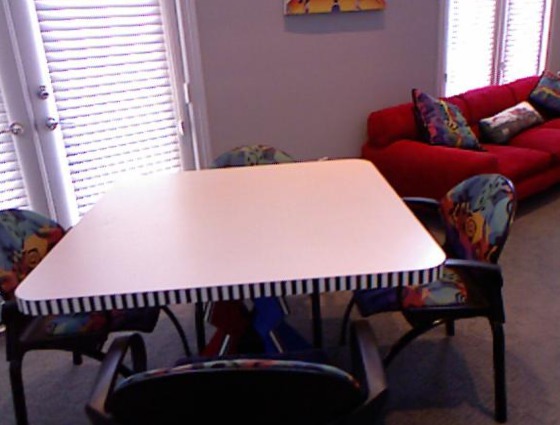} &
\includegraphics[width=0.3\linewidth]{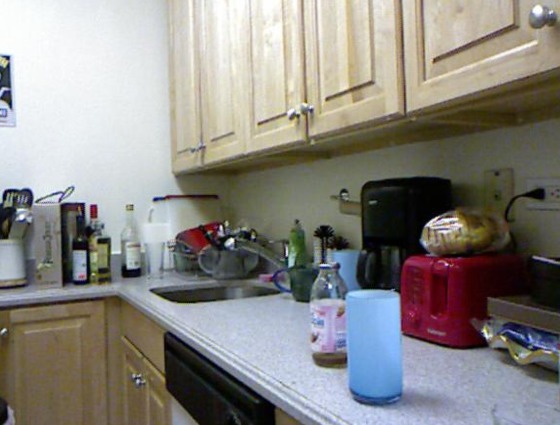} 
\\
\multicolumn{2}{c}{How many chairs are there?} & 
\multicolumn{1}{c}{What is the object fixed on the window?} & 
\multicolumn{1}{c}{Which item is red in colour?}
\\\midrule
\textit{\AproachName:}&\multicolumn{1}{c}{\textcolor{red}{1}} & 
\multicolumn{1}{c}{\textcolor{red}{curtain}} & 
\multicolumn{1}{c}{\textcolor{red}{remote control}}
\\\midrule
\textit{Question-only:}&\multicolumn{1}{c}{\textcolor{red}{4}} & 
\multicolumn{1}{c}{\textcolor{red}{curtain}} & 
\multicolumn{1}{c}{\textcolor{red}{clock}}
\\\midrule
\textit{Ground truth answers:}&\multicolumn{1}{c}{\textcolor{black}{2}} & 
\multicolumn{1}{c}{\textcolor{black}{handle}} & 
\multicolumn{1}{c}{\textcolor{black}{toaster}}\\
\bottomrule
\end{tabular}
\end{center}
\caption{Examples of questions and answers on DAQUAR - failure cases.}

\label{fig:mix_predictions}
\end{table*}

\begin{table*}[t]

\begin{center}
\begin{tabular}{l@{\ }c@{\ }c@{\ }c}

 \toprule
\multicolumn{2}{c}{\includegraphics[width=0.3\linewidth]{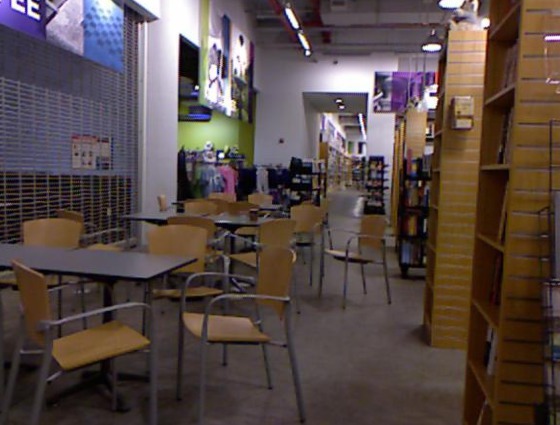}} &
\includegraphics[width=0.3\linewidth]{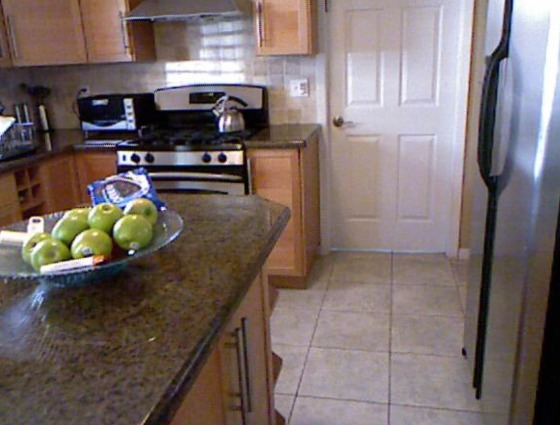} &
\includegraphics[width=0.3\linewidth]{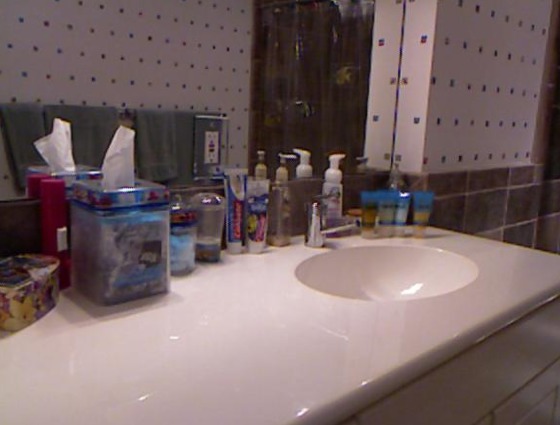}
\\
\multicolumn{2}{c}{What are the objects close to the wall?}  &
\multicolumn{1}{c}{What is on the stove?} &
\multicolumn{1}{c}{What is left of sink?}
\\\midrule
\textit{\AproachName:}&\multicolumn{1}{c}{\textcolor{green}{wall decoration}} &
\multicolumn{1}{c}{\textcolor{green}{tea kettle}} &
\multicolumn{1}{c}{\textcolor{green}{tissue roll}}
\\\midrule
\textit{Question-only:}&\multicolumn{1}{c}{\textcolor{red}{books}} &
\multicolumn{1}{c}{\textcolor{green}{tea kettle}} &
\multicolumn{1}{c}{\textcolor{red}{towel}}
\\\midrule
\textit{Ground truth answers:}&\multicolumn{1}{c}{\textcolor{black}{wall decoration}} &
\multicolumn{1}{c}{\textcolor{black}{tea kettle}} &
\multicolumn{1}{c}{\textcolor{black}{tissue roll}}\\
\bottomrule
\end{tabular}
\end{center}
\caption{Examples of compound answer words on DAQUAR.}
\label{fig:compound_answers}
\end{table*}
\begin{table*}[t]

\begin{center}
\begin{tabular}{l@{\ }c@{\ }c@{\ }c}

 \toprule
\multicolumn{2}{c}{\includegraphics[width=0.3\linewidth]{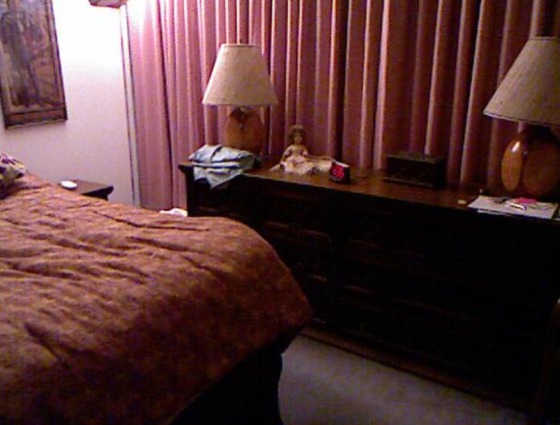}} &
\includegraphics[width=0.3\linewidth]{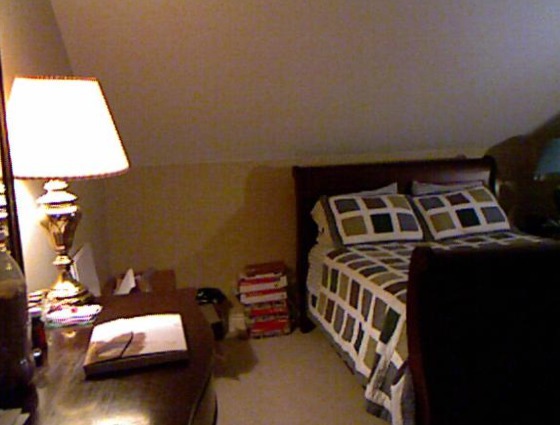} &
\includegraphics[width=0.3\linewidth]{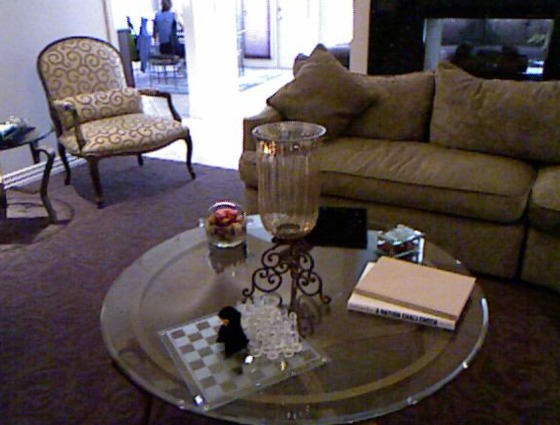}
\\
\multicolumn{2}{c}{How many lamps are there?}  &
\multicolumn{1}{c}{How many pillows are there on the bed?} &
\multicolumn{1}{c}{How many pillows are there on the sofa?}
\\\midrule
\textit{\AproachName:}&\multicolumn{1}{c}{\textcolor{green}{2}} &
\multicolumn{1}{c}{\textcolor{green}{2}} &
\multicolumn{1}{c}{\textcolor{green}{3}}
\\\midrule
\textit{Question-only:}&\multicolumn{1}{c}{\textcolor{green}{2}} &
\multicolumn{1}{c}{\textcolor{red}{3}} &
\multicolumn{1}{c}{\textcolor{green}{3}}
\\\midrule
\textit{Ground truth answers:}&\multicolumn{1}{c}{\textcolor{black}{2}} &
\multicolumn{1}{c}{\textcolor{black}{2}} &
\multicolumn{1}{c}{\textcolor{black}{3}}\\
\bottomrule
\end{tabular}
\end{center}
\caption{Counting questions on DAQUAR.}

\label{fig:counting}
\end{table*}
\begin{table*}[t]

\begin{center}
\begin{tabular}{l@{\ }c@{\ }c@{\ }c}

 \toprule
\multicolumn{2}{c}{\includegraphics[width=0.3\linewidth]{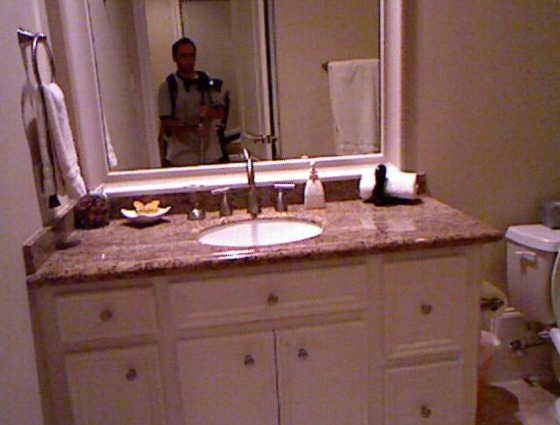}} &
\includegraphics[width=0.3\linewidth]{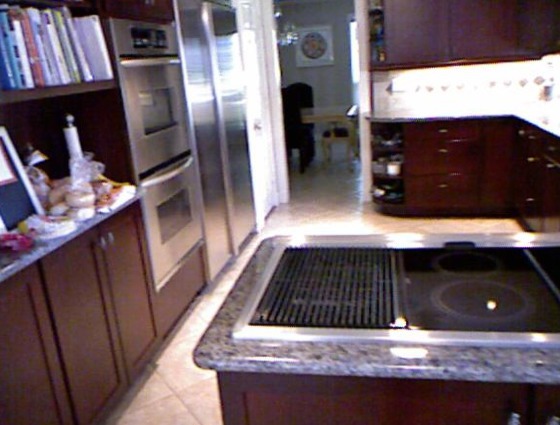} &
\includegraphics[width=0.3\linewidth]{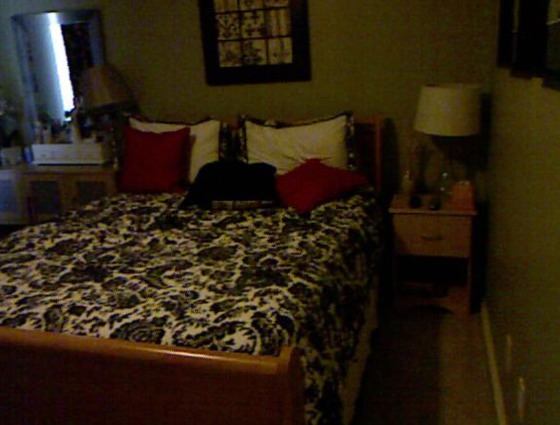}
\\
\multicolumn{2}{c}{What color is the towel?}  &
\multicolumn{1}{c}{What color are the cabinets?} &
\multicolumn{1}{c}{What is the colour of the pillows?}
\\\midrule
\textit{\AproachName:}&\multicolumn{1}{c}{\textcolor{red}{brown}} &
\multicolumn{1}{c}{\textcolor{green}{brown}} &
\multicolumn{1}{c}{\textcolor{green}{black}, \textcolor{green}{white}}
\\\midrule
\textit{Question-only:}&\multicolumn{1}{c}{\textcolor{green}{white}} &
\multicolumn{1}{c}{\textcolor{green}{brown}} &
\multicolumn{1}{c}{\textcolor{red}{blue}, \textcolor{red}{green}, \textcolor{green}{red}}
\\\midrule
\textit{Ground truth answers:}&\multicolumn{1}{c}{\textcolor{black}{white}} &
\multicolumn{1}{c}{\textcolor{black}{brown}} &
\multicolumn{1}{c}{\textcolor{black}{black, red, white}}\\
\bottomrule
\end{tabular}
\end{center}
\caption{Questions about color on DAQUAR.}
\label{fig:colors}
\end{table*}
\begin{table*}[t]

\begin{center}
\begin{tabular}{l@{\ }c@{\ }c@{\ }c}

 \toprule
\multicolumn{2}{c}{\includegraphics[width=0.3\linewidth]{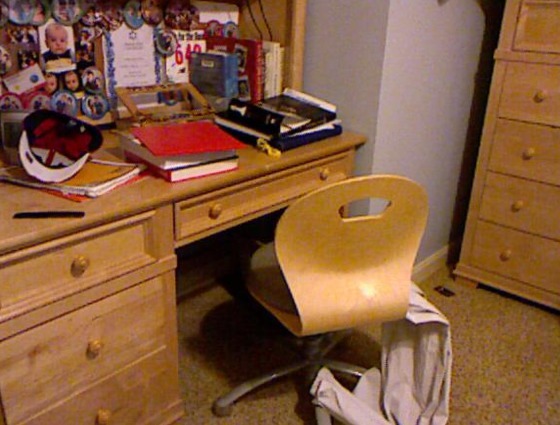}} &
\includegraphics[width=0.3\linewidth]{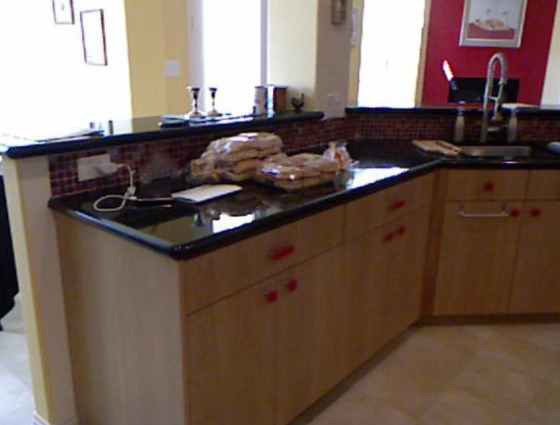} &
\includegraphics[width=0.3\linewidth]{image1021}
\\
\multicolumn{2}{c}{What is hanged on the chair?}  &
\multicolumn{1}{c}{What is the object close to the sink?} &
\multicolumn{1}{c}{What is the object on the table in the corner?}
\\\midrule
\textit{\AproachName:}&\multicolumn{1}{c}{\textcolor{green}{clothes}} &
\multicolumn{1}{c}{\textcolor{green}{faucet}} &
\multicolumn{1}{c}{\textcolor{green}{lamp}}
\\\midrule
\textit{Question-only:}&\multicolumn{1}{c}{\textcolor{red}{jacket}} &
\multicolumn{1}{c}{\textcolor{green}{faucet}} &
\multicolumn{1}{c}{\textcolor{red}{plant}}
\\\midrule
\textit{Ground truth answers:}&\multicolumn{1}{c}{\textcolor{black}{clothes}} &
\multicolumn{1}{c}{\textcolor{black}{faucet}} &
\multicolumn{1}{c}{\textcolor{black}{lamp}}\\
\bottomrule
\end{tabular}
\end{center}
\caption{Correct answers by our ``\AproachName'' architecture on DAQUAR.}
\label{fig:objects1}
\end{table*}
\begin{table*}[t]

\begin{center}
\begin{tabular}{l@{\ }c@{\ }c@{\ }c}

 \toprule
\multicolumn{2}{c}{\includegraphics[width=0.3\linewidth]{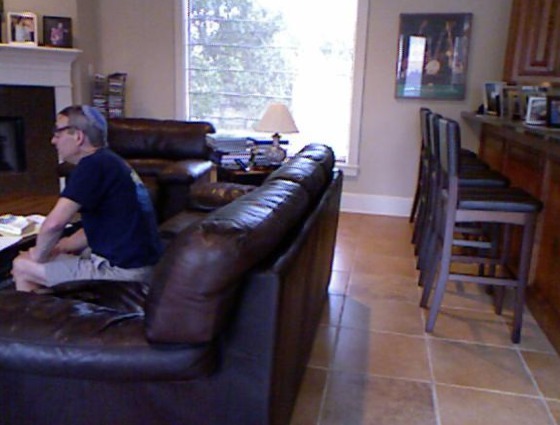}} &
\includegraphics[width=0.3\linewidth]{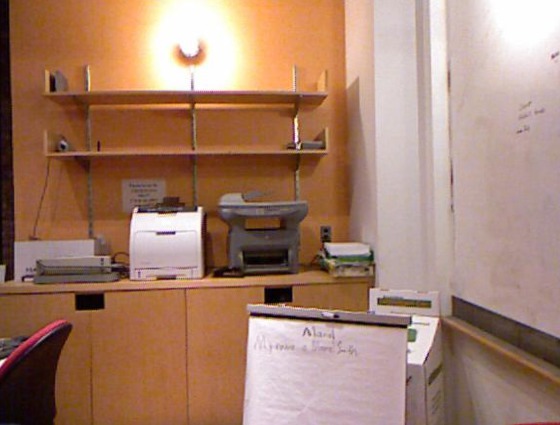} &
\includegraphics[width=0.3\linewidth]{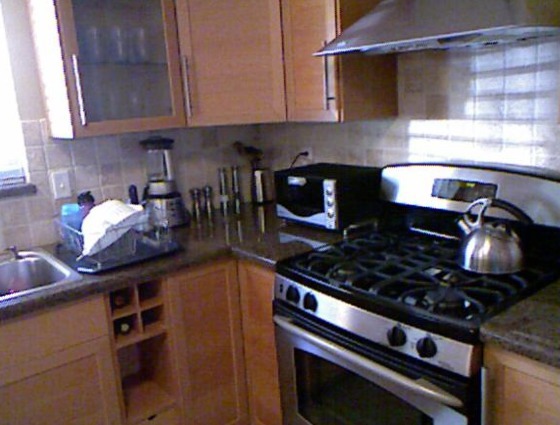}
\\
\multicolumn{2}{c}{What are the things on the cabinet?}  &
\multicolumn{1}{c}{What is in front of the shelf?} &
\multicolumn{1}{c}{How many burner knobs are there?}
\\\midrule
\textit{\AproachName:}&\multicolumn{1}{c}{\textcolor{green}{photo}} &
\multicolumn{1}{c}{\textcolor{green}{chair}} &
\multicolumn{1}{c}{\textcolor{green}{4}}
\\\midrule
\textit{Question-only:}&\multicolumn{1}{c}{\textcolor{green}{photo}} &
\multicolumn{1}{c}{\textcolor{red}{basket}} &
\multicolumn{1}{c}{\textcolor{red}{6}}
\\\midrule
\textit{Ground truth answers:}&\multicolumn{1}{c}{\textcolor{black}{photo}} &
\multicolumn{1}{c}{\textcolor{black}{chair}} &
\multicolumn{1}{c}{\textcolor{black}{4}}\\
\bottomrule
\end{tabular}
\end{center}
\caption{Correct answers by our ``\AproachName'' architecture on DAQUAR.}
\label{fig:objects2}
\end{table*}
\begin{table*}[t]

\begin{center}
\begin{tabular}{l@{\ }c@{\ }c@{\ }c}

\toprule
\multicolumn{2}{c}{\includegraphics[width=0.3\linewidth]{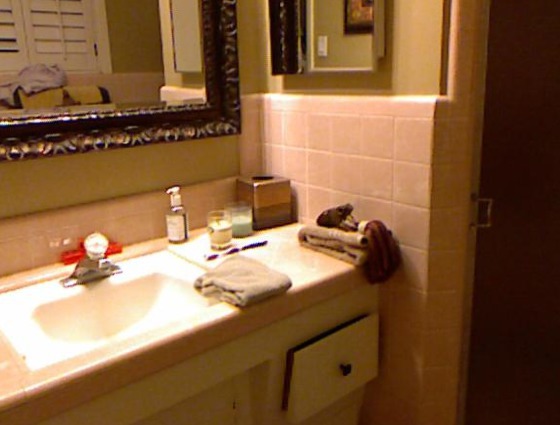}} &
\includegraphics[width=0.3\linewidth]{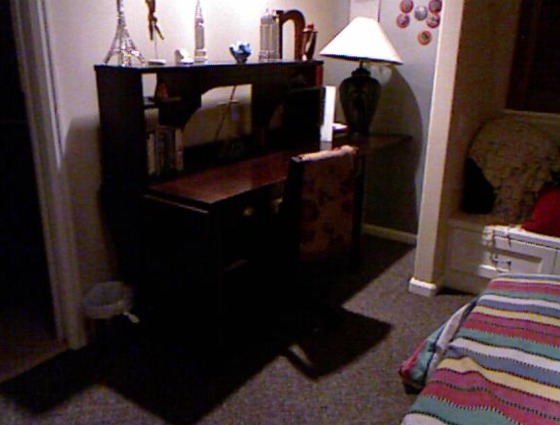} &
\includegraphics[width=0.3\linewidth]{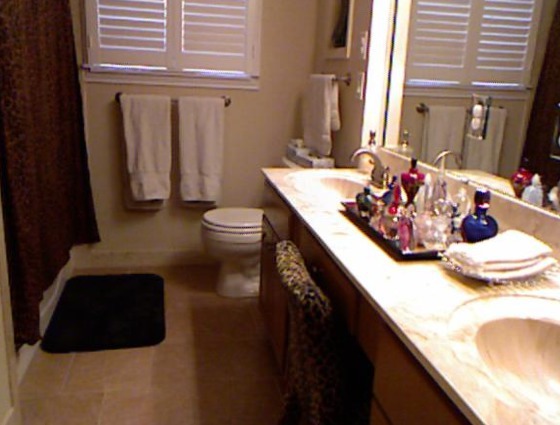}
\\
\multicolumn{2}{c}{What is the object close to the counter?}  &
\multicolumn{1}{c}{What is the colour of the table and chair?} &
\multicolumn{1}{c}{How many towels are hanged?}
\\\midrule
\textit{\AproachName:}&\multicolumn{1}{c}{\textcolor{green}{sink}} &
\multicolumn{1}{c}{\textcolor{green}{brown}} &
\multicolumn{1}{c}{\textcolor{green}{3}}
\\\midrule
\textit{Question-only:}&\multicolumn{1}{c}{\textcolor{red}{stove}} &
\multicolumn{1}{c}{\textcolor{green}{brown}} &
\multicolumn{1}{c}{\textcolor{red}{4}}
\\\midrule
\textit{Ground truth answers:}&\multicolumn{1}{c}{\textcolor{black}{sink}} &
\multicolumn{1}{c}{\textcolor{black}{brown}} &
\multicolumn{1}{c}{\textcolor{black}{3}}\\
\bottomrule
\end{tabular}
\end{center}
\caption{Correct answers by our ``\AproachName'' architecture on DAQUAR.}
\label{fig:objects3}
\end{table*}
\begin{table*}[t]

\begin{center}
\begin{tabular}{l@{\ }c@{\ }c@{\ }c}

\toprule
\multicolumn{2}{c}{\includegraphics[width=0.3\linewidth]{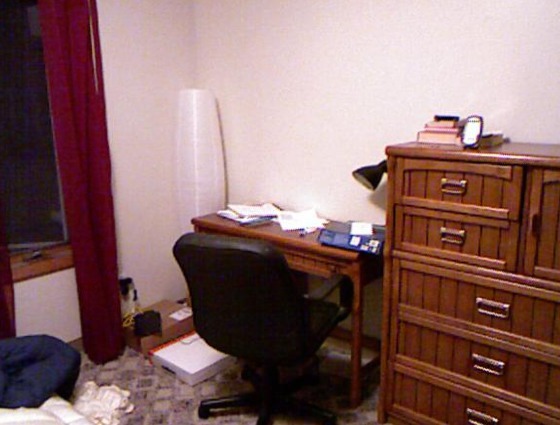}} &
\includegraphics[width=0.3\linewidth]{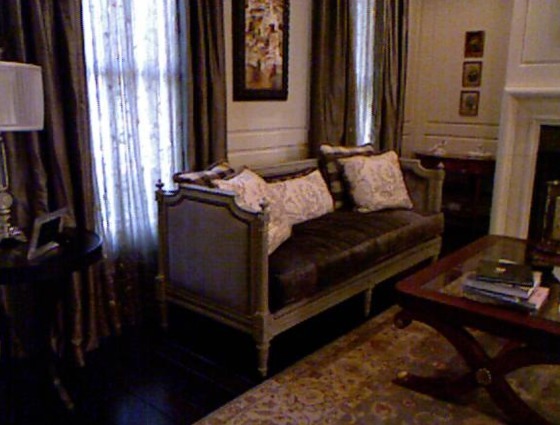} &
\includegraphics[width=0.3\linewidth]{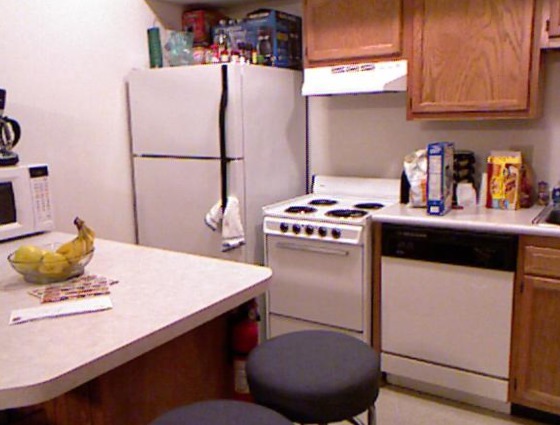}
\\
\multicolumn{2}{c}{What is on the right most side on the table?}  &
\multicolumn{1}{c}{What are the things on the coffee table?} &
\multicolumn{1}{c}{What is in front of the table?}
\\\midrule
\textit{\AproachName:}&\multicolumn{1}{c}{\textcolor{green}{lamp}} &
\multicolumn{1}{c}{\textcolor{green}{books}} &
\multicolumn{1}{c}{\textcolor{green}{chair}}
\\\midrule
\textit{Question-only:}&\multicolumn{1}{c}{\textcolor{red}{machine}} &
\multicolumn{1}{c}{\textcolor{red}{jacket}} &
\multicolumn{1}{c}{\textcolor{green}{chair}}
\\\midrule
\textit{Ground truth answers:}&\multicolumn{1}{c}{\textcolor{black}{lamp}} &
\multicolumn{1}{c}{\textcolor{black}{books}} &
\multicolumn{1}{c}{\textcolor{black}{chair}}\\
\bottomrule
\end{tabular}
\end{center}
\caption{Correct answers by our ``\AproachName'' architecture on DAQUAR.}
\label{fig:objects4}
\end{table*}
\begin{table*}[t]

\begin{center}
\begin{tabular}{l@{\ }c@{\ }c@{\ }c}

 \toprule
\multicolumn{2}{c}{\includegraphics[width=0.3\linewidth]{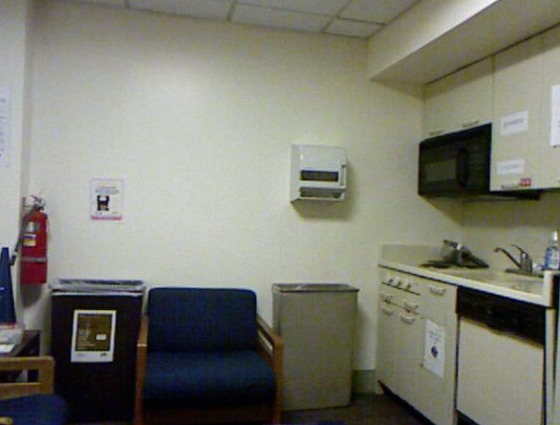}} &
\includegraphics[width=0.3\linewidth]{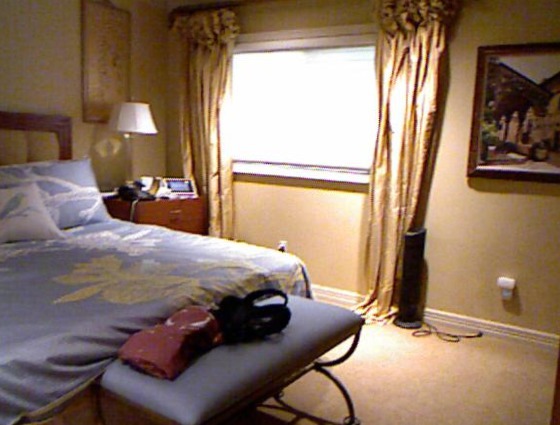} &
\includegraphics[width=0.3\linewidth]{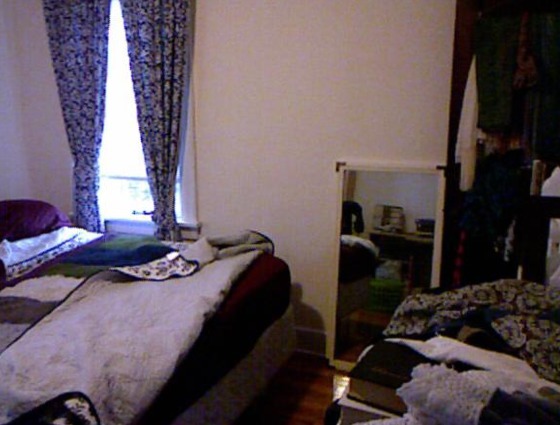}
\\
\multicolumn{2}{c}{What is on the left side of}  &
\multicolumn{1}{c}{What are the things on the cabinet?} &
\multicolumn{1}{c}{What color is the frame}
\\
\multicolumn{2}{c}{the white oven on the floor and} &
\multicolumn{1}{c}{} &
\multicolumn{1}{c}{of the mirror close to the wardrobe?}
\\
\multicolumn{2}{c}{on right side of the blue armchair?} & &
\\\midrule
\textit{\AproachName:}&\multicolumn{1}{c}{\textcolor{red}{oven}} &
\multicolumn{1}{c}{\textcolor{red}{chair}, \textcolor{green}{lamp}, \textcolor{green}{photo}} &
\multicolumn{1}{c}{\textcolor{red}{pink}}
\\\midrule
\textit{Question-only:}&\multicolumn{1}{c}{\textcolor{red}{exercise equipment}} &
\multicolumn{1}{c}{\textcolor{red}{candelabra}} &
\multicolumn{1}{c}{\textcolor{red}{curtain}}
\\\midrule
\textit{Ground truth answers:}&\multicolumn{1}{c}{\textcolor{black}{garbage bin}} &
\multicolumn{1}{c}{\textcolor{black}{lamp, photo, telephone}} &
\multicolumn{1}{c}{\textcolor{black}{white}}\\
\bottomrule
\end{tabular}
\end{center}
\caption{Failure cases on DAQUAR.}

\label{fig:failures}
\end{table*}

\begin{table*}[p!]
\begin{center}
\begin{tabular}{l@{\ }c@{\ }c@{\ }c}
\toprule
\multicolumn{2}{c}{\includegraphics[width=0.3\linewidth, height=0.25\linewidth]{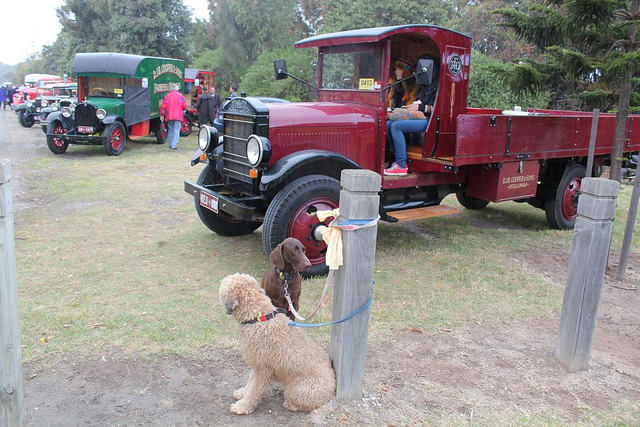}} &
\includegraphics[width=0.3\linewidth,height=0.25\linewidth]{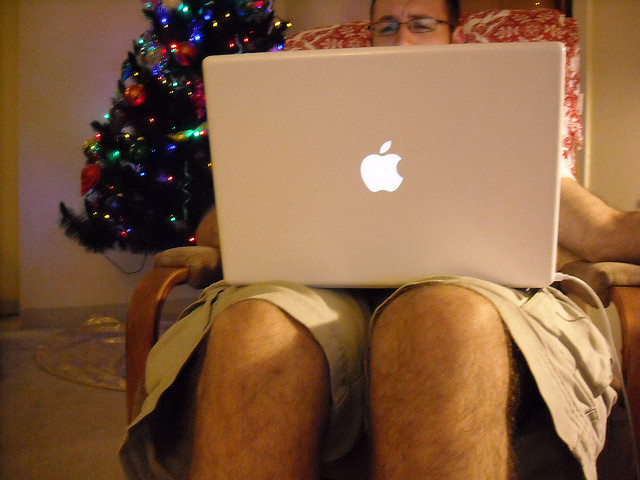} &
\includegraphics[width=0.3\linewidth,height=0.25\linewidth]{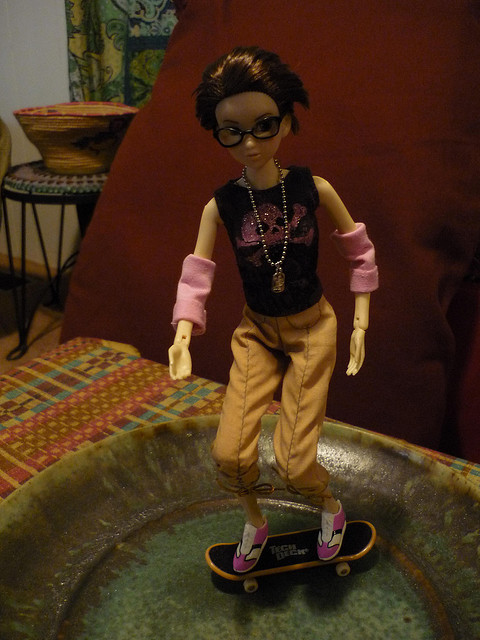} 
\\
\multicolumn{2}{c}{Are the dogs tied?} 
& \multicolumn{1}{c}{Is it summer time?} 
& \multicolumn{1}{c}{Is this a real person?}
\\\midrule
\textit{\AproachName:}&\multicolumn{1}{c}{\textcolor{green}{yes}} & \multicolumn{1}{c}{\textcolor{green}{no}} & \multicolumn{1}{c}{\textcolor{green}{no}}
\\
\bottomrule
\end{tabular}
\end{center}
\caption{Examples of 'yes/no' questions and answers produced by our the best model on test VQA.}

\label{fig:vqa-image_qa-yes_no}
\end{table*}

\begin{table*}[p!]
\begin{center}
\begin{tabular}{l@{\ }c@{\ }c@{\ }c}
\toprule
\multicolumn{2}{c}{\includegraphics[width=0.3\linewidth, height=0.25\linewidth]{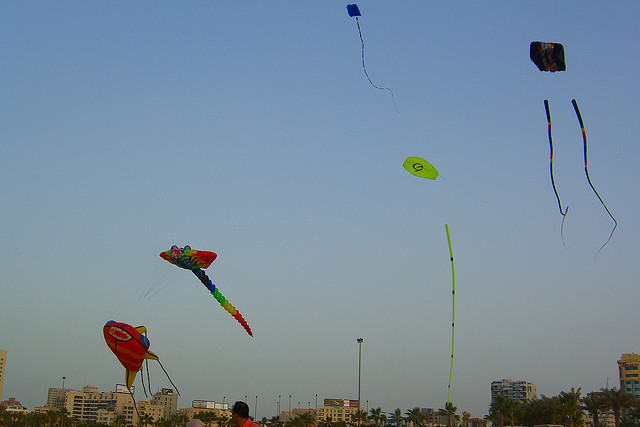}} &
\includegraphics[width=0.3\linewidth,height=0.25\linewidth]{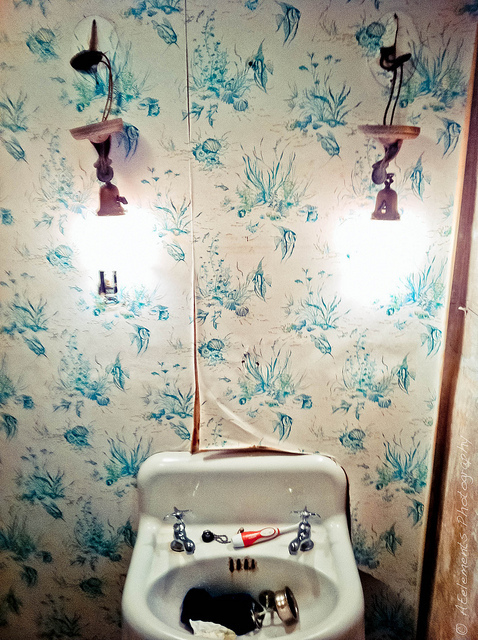} &
\includegraphics[width=0.3\linewidth,height=0.25\linewidth]{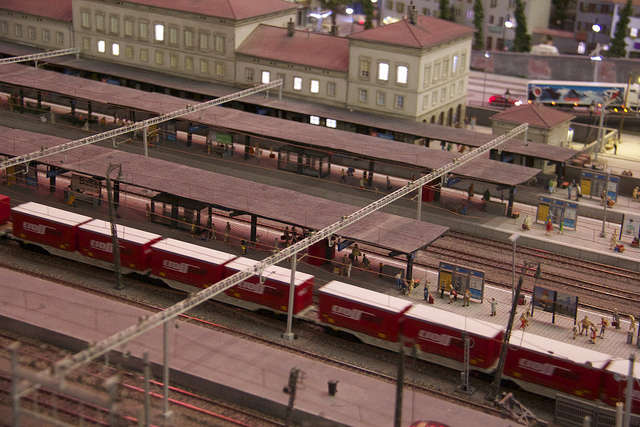} 
\\
\multicolumn{2}{c}{How many kites are only yellow?} 
& \multicolumn{1}{c}{How many taps are on the sink?} 
& \multicolumn{1}{c}{How many windows are lit?}
\\\midrule
\textit{\AproachName:}&\multicolumn{1}{c}{\textcolor{green}{1}} & \multicolumn{1}{c}{\textcolor{green}{2}} & \multicolumn{1}{c}{\textcolor{green}{12}}
\\
\bottomrule
\end{tabular}
\end{center}
\caption{Examples of 'counting' questions and answers produced by our the best model on test VQA.}

\label{fig:vqa-image_qa-counting}
\end{table*}

\begin{table*}[p!]
\begin{center}
\begin{tabular}{l@{\ }c@{\ }c@{\ }c}
\toprule
\multicolumn{2}{c}{\includegraphics[width=0.3\linewidth, height=0.25\linewidth]{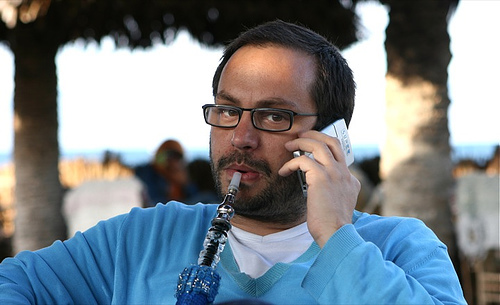}} &
\includegraphics[width=0.3\linewidth,height=0.25\linewidth]{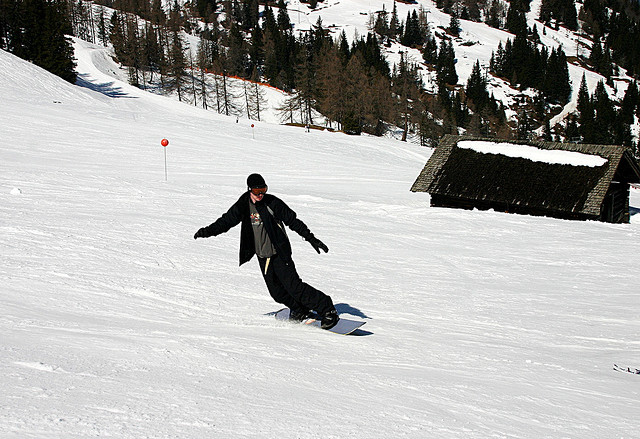} &
\includegraphics[width=0.3\linewidth,height=0.25\linewidth]{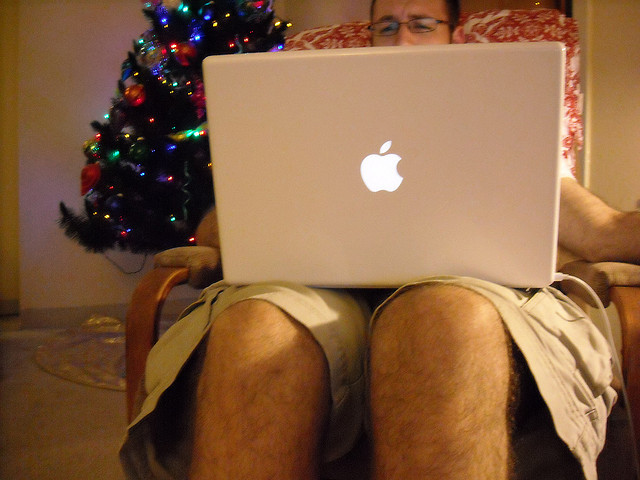} 
\\
\multicolumn{2}{c}{What is the man holding to his ear?} 
& \multicolumn{1}{c}{What sport is this man enjoying?} 
& \multicolumn{1}{c}{What brand is the laptop?}
\\\midrule
\textit{\AproachName:}&\multicolumn{1}{c}{\textcolor{green}{phone}} & \multicolumn{1}{c}{\textcolor{green}{snowboarding}} & \multicolumn{1}{c}{\textcolor{green}{apple}}
\\
\bottomrule
\end{tabular}
\end{center}
\caption{Examples of 'what' questions and answers produced by our the best model on test VQA.}

\label{fig:vqa-image_qa-what}
\end{table*}

\begin{table*}[p!]
\begin{center}
\begin{tabular}{l@{\ }c@{\ }c@{\ }c}
\toprule
\multicolumn{2}{c}{\includegraphics[width=0.3\linewidth, height=0.25\linewidth]{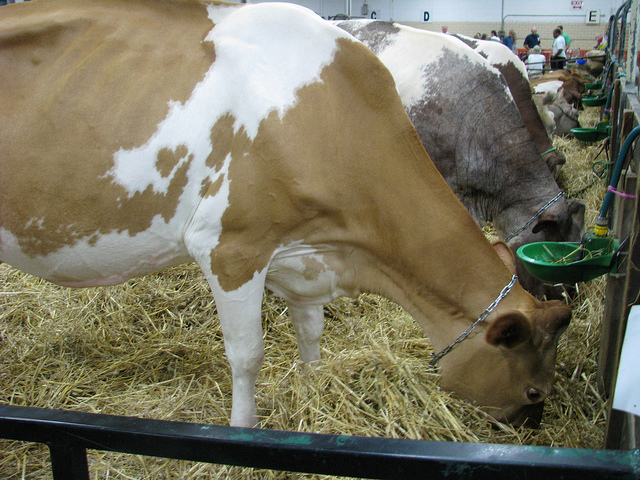}} &
\includegraphics[width=0.3\linewidth,height=0.25\linewidth]{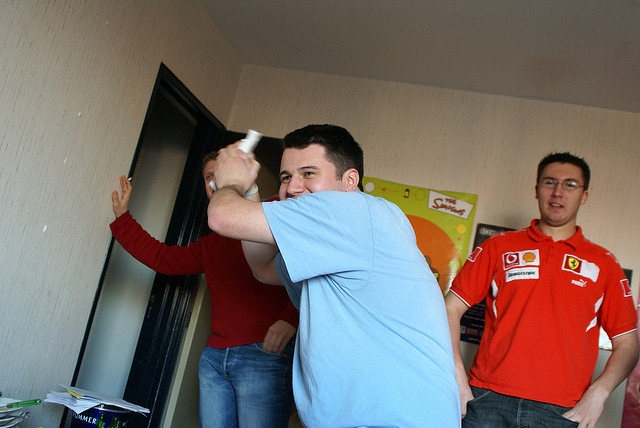} &
\includegraphics[width=0.3\linewidth,height=0.25\linewidth]{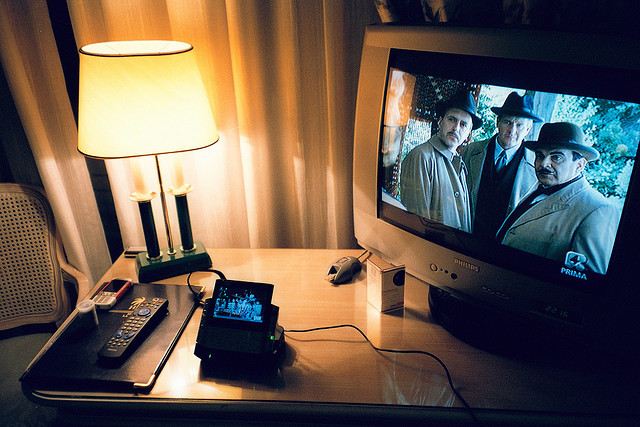} 
\\
\multicolumn{2}{c}{Color of cow?} 
& \multicolumn{1}{c}{What is the man doing?} 
& \multicolumn{1}{c}{Where is the TV control?}
\\\midrule
\textit{\AproachName:}&\multicolumn{1}{c}{\textcolor{green}{brown and white}} & \multicolumn{1}{c}{\textcolor{green}{playing wii}} & \multicolumn{1}{c}{\textcolor{green}{on table}}
\\
\bottomrule
\end{tabular}
\end{center}
\caption{Examples of 'compound answers' questions and answers produced by our the best model on test VQA.}

\label{fig:vqa-image_qa-compound}
\end{table*}

\section{Conclusions}\label{sec:conclusions}
We have presented a neural architecture for answering natural language questions about images that contrasts with prior efforts based on semantic parsing and outperforms a prior symbolic based approach by doubling performance on this challenging task. A variant of our model, that does not use the image to answer the question, already explains a substantial part of the overall performance and helps to understand the contribution of the visual features in this task.
From a comparison with the human baseline where humans are not shown the image to answer the question, we conclude that this language-only model has learnt biases and patterns that can be seen as forms of common sense and prior knowledge that  are also used by humans to accomplish this task. %
 We have extended our existing DAQUAR dataset to the new \daquarNew, which now provides multiple reference answers which allows to study inter-human agreement and consensus on the question answering task. We propose two new metrics: ``Average Consensus'', which takes into account human disagreement, and ``Min Consensus''
that captures disagreement in human question answering. Finally, we extend our analysis to the large-scale VQA dataset showing competitive performance, yet still using global visual model, and training the model solely on the provided question answer image triples. A broader analysis of the different Deep Learning components and design choices in our model has led to improved results and highlights the importance of a strong visual model. %

\begin{acknowledgements}
Marcus Rohrbach was supported by a fellowship within the FITweltweit-Program of the German Academic Exchange Service (DAAD). The project was in part supported by the Collaborative Research Center (CRC) 1223 from the German Research Foundation (DFG).
\end{acknowledgements}

\onecolumn \twocolumn
\bibliographystyle{plainnat}
\bibliography{biblioLong,egbib,rohrbach}   %

\begin{thebibliography}{89}
\providecommand{\natexlab}[1]{#1}
\providecommand{\url}[1]{\texttt{#1}}
\expandafter\ifx\csname urlstyle\endcsname\relax
  \providecommand{\doi}[1]{doi: #1}\else
  \providecommand{\doi}{doi: \begingroup \urlstyle{rm}\Url}\fi

\bibitem[Akata et~al.(2016)Akata, Malinowski, Fritz, and
  Schiele]{akata2016multi}
Zeynep Akata, Mateusz Malinowski, Mario Fritz, and Bernt Schiele.
\newblock Multi-cue zero-shot learning with strong supervision.
\newblock In \emph{Proceedings of the IEEE Conference on Computer Vision and
  Pattern Recognition (CVPR)}, 2016.

\bibitem[Andreas et~al.(2016{\natexlab{a}})Andreas, Rohrbach, Darrell, and
  Klein]{andreas16cvpr}
Jacob Andreas, Marcus Rohrbach, Trevor Darrell, and Dan Klein.
\newblock Neural module networks.
\newblock In \emph{Proceedings of the IEEE Conference on Computer Vision and
  Pattern Recognition (CVPR)}, 2016{\natexlab{a}}.

\bibitem[Andreas et~al.(2016{\natexlab{b}})Andreas, Rohrbach, Darrell, and
  Klein]{andreas16naacl}
Jacob Andreas, Marcus Rohrbach, Trevor Darrell, and Dan Klein.
\newblock Learning to compose neural networks for question answering.
\newblock In \emph{Proceedings of the Conference of the North American Chapter
  of the Association for Computational Linguistics (NAACL)},
  2016{\natexlab{b}}.

\bibitem[Antol et~al.(2015)Antol, Agrawal, Lu, Mitchell, Batra, Zitnick, and
  Parikh]{antol2015vqa}
Stanislaw Antol, Aishwarya Agrawal, Jiasen Lu, Margaret Mitchell, Dhruv Batra,
  C~Lawrence Zitnick, and Devi Parikh.
\newblock Vqa: Visual question answering.
\newblock In \emph{Proceedings of the IEEE International Conference on Computer
  Vision (ICCV)}, 2015.

\bibitem[Bastien et~al.(2012)Bastien, Lamblin, Pascanu, Bergstra, Goodfellow,
  Bergeron, Bouchard, and Bengio]{Bastien-Theano-2012}
Fr{\'{e}}d{\'{e}}ric Bastien, Pascal Lamblin, Razvan Pascanu, James Bergstra,
  Ian~J. Goodfellow, Arnaud Bergeron, Nicolas Bouchard, and Yoshua Bengio.
\newblock Theano: new features and speed improvements.
\newblock Deep Learning and Unsupervised Feature Learning NIPS 2012 Workshop,
  2012.

\bibitem[Berant and Liang(2014)]{berant2014semantic}
Jonathan Berant and Percy Liang.
\newblock Semantic parsing via paraphrasing.
\newblock In \emph{Proceedings of the Annual Meeting of the Association for
  Computational Linguistics (ACL)}, 2014.

\bibitem[Chen et~al.(2015)Chen, Wang, Chen, Gao, Xu, and Nevatia]{chen2015abc}
Kan Chen, Jiang Wang, Liang-Chieh Chen, Haoyuan Gao, Wei Xu, and Ram Nevatia.
\newblock Abc-cnn: An attention based convolutional neural network for visual
  question answering.
\newblock \emph{arXiv:1511.05960}, 2015.

\bibitem[Cho et~al.(2014)Cho, van Merrienboer, Gulcehre, Bougares, Schwenk,
  Bahdanau, and Bengio]{cho2014learning}
Kyunghyun Cho, Bart van Merrienboer, Caglar Gulcehre, Fethi Bougares, Holger
  Schwenk, Dzmitry Bahdanau, and Yoshua Bengio.
\newblock Learning phrase representations using rnn encoder-decoder for
  statistical machine translation.
\newblock In \emph{Proceedings of the Conference on Empirical Methods in
  Natural Language Processing (EMNLP)}, 2014.

\bibitem[Chollet(2015)]{chollet2015}
François Chollet.
\newblock keras.
\newblock \url{https://github.com/fchollet/keras}, 2015.

\bibitem[Chowdhury et~al.(2016)Chowdhury, Malinowski, Bulling, and
  Fritz]{sreyasi16icmr}
Sreyasi~Nag Chowdhury, Mateusz Malinowski, Andreas Bulling, and Mario Fritz.
\newblock Xplore-m-ego: Contextual media retrieval using natural language
  queries.
\newblock In \emph{ACM International Conference on Multimedia Retrieval
  (ICMR)}, 2016.

\bibitem[Cohen et~al.(1960)]{cohen1960coefficient}
Jacob Cohen et~al.
\newblock A coefficient of agreement for nominal scales.
\newblock \emph{Educational and psychological measurement}, 1960.

\bibitem[Donahue et~al.(2015)Donahue, Hendricks, Guadarrama, Rohrbach,
  Venugopalan, Saenko, and Darrell]{donahue15cvpr}
Jeff Donahue, Lisa~Anne Hendricks, Sergio Guadarrama, Marcus Rohrbach,
  Subhashini Venugopalan, Kate Saenko, and Trevor Darrell.
\newblock Long-term recurrent convolutional networks for visual recognition and
  description.
\newblock In \emph{Proceedings of the IEEE Conference on Computer Vision and
  Pattern Recognition (CVPR)}, 2015.

\bibitem[Fleiss and Cohen(1973)]{fleiss1973equivalence}
Joseph~L Fleiss and Jacob Cohen.
\newblock The equivalence of weighted kappa and the intraclass correlation
  coefficient as measures of reliability.
\newblock \emph{Educational and psychological measurement}, 1973.

\bibitem[Fukui et~al.(2016)Fukui, Park, Yang, Rohrbach, Darrell, and
  Rohrbach]{fukui16emnlp}
Akira Fukui, Dong~Huk Park, Daylen Yang, Anna Rohrbach, Trevor Darrell, and
  Marcus Rohrbach.
\newblock Multimodal compact bilinear pooling for visual question answering and
  visual grounding.
\newblock In \emph{Proceedings of the Conference on Empirical Methods in
  Natural Language Processing (EMNLP)}, 2016.

\bibitem[Gao et~al.(2015)Gao, Mao, Zhou, Huang, Wang, and Xu]{gao2015you}
Haoyuan Gao, Junhua Mao, Jie Zhou, Zhiheng Huang, Lei Wang, and Wei Xu.
\newblock Are you talking to a machine? dataset and methods for multilingual
  image question answering.
\newblock In \emph{Advances in Neural Information Processing Systems (NIPS)},
  2015.

\bibitem[Geman et~al.(2015)Geman, Geman, Hallonquist, and
  Younes]{geman2015visual}
Donald Geman, Stuart Geman, Neil Hallonquist, and Laurent Younes.
\newblock Visual turing test for computer vision systems.
\newblock In \emph{Proceedings of the National Academy of Sciences}. National
  Academy of Sciences, 2015.

\bibitem[He et~al.(2015)He, Zhang, Ren, and Sun]{he2015deep}
Kaiming He, Xiangyu Zhang, Shaoqing Ren, and Jian Sun.
\newblock Deep residual learning for image recognition.
\newblock \emph{arXiv:1512.03385}, 2015.

\bibitem[Hochreiter and Schmidhuber(1997)]{hochreiter97nc}
Sepp Hochreiter and J{\"u}rgen Schmidhuber.
\newblock Long short-term memory.
\newblock \emph{Neural Computation}, 1997.

\bibitem[Hu et~al.(2016{\natexlab{a}})Hu, Rohrbach, and Darrell]{hu16eccv}
Ronghang Hu, Marcus Rohrbach, and Trevor Darrell.
\newblock Segmentation from natural language expressions.
\newblock In \emph{Proceedings of the European Conference on Computer Vision
  (ECCV)}, 2016{\natexlab{a}}.

\bibitem[Hu et~al.(2016{\natexlab{b}})Hu, Xu, Rohrbach, Feng, Saenko, and
  Darrell]{hu16cvpr}
Ronghang Hu, Huazhe Xu, Marcus Rohrbach, Jiashi Feng, Kate Saenko, and Trevor
  Darrell.
\newblock Natural language object retrieval.
\newblock In \emph{Proceedings of the IEEE Conference on Computer Vision and
  Pattern Recognition (CVPR)}, 2016{\natexlab{b}}.

\bibitem[Ilievski et~al.(2016)Ilievski, Yan, and Feng]{ilievski2016fda}
Ilija Ilievski, Shuicheng Yan, and Jiashi Feng.
\newblock A focused dynamic attention model for visual question answering.
\newblock \emph{arXiv:1604.01485}, 2016.

\bibitem[Iyyer et~al.(2014)Iyyer, Boyd-Graber, Claudino, Socher, and
  III]{iyyer2014neural}
Mohit Iyyer, Jordan Boyd-Graber, Leonardo Claudino, Richard Socher, and
  Hal~Daum{\'e} III.
\newblock A neural network for factoid question answering over paragraphs.
\newblock In \emph{Proceedings of the Conference on Empirical Methods in
  Natural Language Processing (EMNLP)}, 2014.

\bibitem[Jia et~al.(2014)Jia, Shelhamer, Donahue, Karayev, Long, Girshick,
  Guadarrama, and Darrell]{jia2014caffe}
Yangqing Jia, Evan Shelhamer, Jeff Donahue, Sergey Karayev, Jonathan Long, Ross
  Girshick, Sergio Guadarrama, and Trevor Darrell.
\newblock Caffe: Convolutional architecture for fast feature embedding.
\newblock \emph{arXiv:1408.5093}, 2014.

\bibitem[Jiang et~al.(2015)Jiang, Wang, Porikli, and
  Li]{jiang2015compositional}
Aiwen Jiang, Fang Wang, Fatih Porikli, and Yi~Li.
\newblock Compositional memory for visual question answering.
\newblock \emph{arXiv:1511.05676}, 2015.

\bibitem[Kafle and Kanan(2016)]{kafle2016answer}
Kushal Kafle and Christopher Kanan.
\newblock Answer-type prediction for visual question answering.
\newblock In \emph{Proceedings of the IEEE Conference on Computer Vision and
  Pattern Recognition (CVPR)}, 2016.

\bibitem[Kalchbrenner et~al.(2014)Kalchbrenner, Grefenstette, and
  Blunsom]{kalchbrenner2014convolutional}
Nal Kalchbrenner, Edward Grefenstette, and Phil Blunsom.
\newblock A convolutional neural network for modelling sentences.
\newblock In \emph{Proceedings of the Annual Meeting of the Association for
  Computational Linguistics (ACL)}, 2014.

\bibitem[Karpathy and Fei-Fei(2015)]{karpathy15cvpr}
Andrej Karpathy and Li~Fei-Fei.
\newblock Deep visual-semantic alignments for generating image descriptions.
\newblock In \emph{Proceedings of the IEEE Conference on Computer Vision and
  Pattern Recognition (CVPR)}, 2015.

\bibitem[Karpathy et~al.(2014)Karpathy, Joulin, and Fei-Fei]{karpathy2014deep}
Andrej Karpathy, Armand Joulin, and Li~Fei-Fei.
\newblock Deep fragment embeddings for bidirectional image sentence mapping.
\newblock In \emph{Advances in Neural Information Processing Systems (NIPS)},
  2014.

\bibitem[Kazemzadeh et~al.(2014)Kazemzadeh, Ordonez, Matten, and
  Berg]{kazemzadeh14emnlp}
Sahar Kazemzadeh, Vicente Ordonez, Mark Matten, and Tamara~L. Berg.
\newblock Referit game: Referring to objects in photographs of natural scenes.
\newblock In \emph{Proceedings of the Conference on Empirical Methods in
  Natural Language Processing (EMNLP)}, 2014.

\bibitem[Kim et~al.(2016)Kim, On, Kim, Ha, and Zhang]{kim2016hadamard}
Jin-Hwa Kim, Kyoung~Woon On, Jeonghee Kim, Jung-Woo Ha, and Byoung-Tak Zhang.
\newblock Hadamard product for low-rank bilinear pooling.
\newblock \emph{arXiv preprint arXiv:1610.04325}, 2016.

\bibitem[Kim(2014)]{kim2014convolutional}
Yoon Kim.
\newblock Convolutional neural networks for sentence classification.
\newblock In \emph{Proceedings of the Conference on Empirical Methods in
  Natural Language Processing (EMNLP)}, 2014.

\bibitem[Kingma and Ba(2014)]{kingma2014adam}
Diederik Kingma and Jimmy Ba.
\newblock Adam: A method for stochastic optimization.
\newblock \emph{arXiv:1412.6980}, 2014.

\bibitem[Klein and Manning(2003)]{klein2003accurate}
Dan Klein and Christopher~D Manning.
\newblock Accurate unlexicalized parsing.
\newblock In \emph{Proceedings of the 41st Annual Meeting on Association for
  Computational Linguistics-Volume 1}, pages 423--430. Association for
  Computational Linguistics, 2003.

\bibitem[Kong et~al.(2014)Kong, Lin, Bansal, Urtasun, and Fidler]{kong2014you}
Chen Kong, Dahua Lin, Mohit Bansal, Raquel Urtasun, and Sanja Fidler.
\newblock What are you talking about? text-to-image coreference.
\newblock In \emph{Proceedings of the IEEE Conference on Computer Vision and
  Pattern Recognition (CVPR)}, 2014.

\bibitem[Krishna et~al.(2016)Krishna, Zhu, Groth, Johnson, Hata, Kravitz, Chen,
  Kalantidis, Li, Shamma, Bernstein, and Fei-Fei]{krishna16arxiv}
Ranjay Krishna, Yuke Zhu, Oliver Groth, Justin Johnson, Kenji Hata, Joshua
  Kravitz, Stephanie Chen, Yannis Kalantidis, Li-Jia Li, David~A Shamma,
  Michael Bernstein, and Li~Fei-Fei.
\newblock Visual genome: Connecting language and vision using crowdsourced
  dense image annotations.
\newblock \emph{arXiv:1602.07332}, 2016.

\bibitem[Krishnamurthy and Kollar(2013)]{krishnamurthy2013jointly}
Jayant Krishnamurthy and Thomas Kollar.
\newblock Jointly learning to parse and perceive: Connecting natural language
  to the physical world.
\newblock In \emph{Transactions of the Association for Computational
  Linguistics (TACL)}, 2013.

\bibitem[Krizhevsky et~al.(2012)Krizhevsky, Sutskever, and
  Hinton]{krizhevsky2012imagenet}
Alex Krizhevsky, Ilya Sutskever, and Geoffrey~E Hinton.
\newblock Imagenet classification with deep convolutional neural networks.
\newblock In \emph{Advances in Neural Information Processing Systems (NIPS)},
  2012.

\bibitem[Kumar et~al.(2015)Kumar, Irsoy, Su, Bradbury, English, Pierce,
  Ondruska, Gulrajani, and Socher]{kumar15arxiv}
Ankit Kumar, Ozan Irsoy, Jonathan Su, James Bradbury, Robert English, Brian
  Pierce, Peter Ondruska, Ishaan Gulrajani, and Richard Socher.
\newblock Ask me anything: Dynamic memory networks for natural language
  processing.
\newblock \emph{arXiv preprint arXiv:1506.07285}, 2015.

\bibitem[LeCun et~al.(1998)LeCun, Bottou, Bengio, and
  Haffner]{lecun1998gradient}
Yann LeCun, L{\'e}on Bottou, Yoshua Bengio, and Patrick Haffner.
\newblock Gradient-based learning applied to document recognition.
\newblock \emph{Proceedings of the IEEE}, 1998.

\bibitem[Liang et~al.(2013)Liang, Jordan, and Klein]{liang2013learning}
Percy Liang, Michael~I Jordan, and Dan Klein.
\newblock Learning dependency-based compositional semantics.
\newblock \emph{Computational Linguistics}, 2013.

\bibitem[Lin et~al.(2014)Lin, Maire, Belongie, Hays, Perona, Ramanan,
  Doll{\'a}r, and Zitnick]{lin2014microsoft}
Tsung-Yi Lin, Michael Maire, Serge Belongie, James Hays, Pietro Perona, Deva
  Ramanan, Piotr Doll{\'a}r, and C~Lawrence Zitnick.
\newblock Microsoft coco: Common objects in context.
\newblock In \emph{Proceedings of the European Conference on Computer Vision
  (ECCV)}, 2014.

\bibitem[{Lu} et~al.(2016){Lu}, {Yang}, {Batra}, and {Parikh}]{lu2016hiecoatt}
Jiasen {Lu}, Jianwei {Yang}, Dhruv {Batra}, and Devi {Parikh}.
\newblock {Hierarchical Co-Attention for Visual Question Answering}.
\newblock In \emph{Advances in Neural Information Processing Systems (NIPS)},
  2016.

\bibitem[Ma et~al.(2016)Ma, Lu, and Li]{learning_to_answer_questions}
Lin Ma, Zhengdong Lu, and Hang Li.
\newblock Learning to answer questions from image using convolutional neural
  network.
\newblock In \emph{Proceedings of the Conference on Artificial Intelligence
  (AAAI)}, 2016.

\bibitem[Malinowski and Fritz(2014{\natexlab{a}})]{malinowski14nips}
Mateusz Malinowski and Mario Fritz.
\newblock A multi-world approach to question answering about real-world scenes
  based on uncertain input.
\newblock In \emph{Advances in Neural Information Processing Systems (NIPS)},
  2014{\natexlab{a}}.

\bibitem[Malinowski and Fritz(2014{\natexlab{b}})]{malinowski14visualturing}
Mateusz Malinowski and Mario Fritz.
\newblock Towards a visual turing challenge.
\newblock In \emph{Learning Semantics (NIPS workshop)}, 2014{\natexlab{b}}.

\bibitem[Malinowski and Fritz(2014{\natexlab{c}})]{malinowski2014pooling}
Mateusz Malinowski and Mario Fritz.
\newblock A pooling approach to modelling spatial relations for image retrieval
  and annotation.
\newblock \emph{arXiv:1411.5190}, 2014{\natexlab{c}}.

\bibitem[Malinowski and Fritz(2015)]{malinowski2015hard}
Mateusz Malinowski and Mario Fritz.
\newblock Hard to cheat: A turing test based on answering questions about
  images.
\newblock \emph{AAAI Workshop: Beyond the Turing Test}, 2015.

\bibitem[Malinowski and Fritz(2016)]{malinowski2016tutorial}
Mateusz Malinowski and Mario Fritz.
\newblock Tutorial on answering questions about images with deep learning.
\newblock \emph{arXiv preprint arXiv:1610.01076}, 2016.

\bibitem[Malinowski et~al.(2015)Malinowski, Rohrbach, and
  Fritz]{malinowski2015ask}
Mateusz Malinowski, Marcus Rohrbach, and Mario Fritz.
\newblock Ask your neurons: A neural-based approach to answering questions
  about images.
\newblock In \emph{Proceedings of the IEEE International Conference on Computer
  Vision (ICCV)}, pages 1--9, 2015.

\bibitem[Manning and Sch{\"u}tze(1999)]{manning1999foundations}
Christopher~D Manning and Hinrich Sch{\"u}tze.
\newblock \emph{Foundations of statistical natural language processing}, volume
  999.
\newblock MIT Press, 1999.

\bibitem[Mao et~al.(2016)Mao, Huang, Toshev, Camburu, Yuille, and
  Murphy]{mao16cvpr}
Junhua Mao, Jonathan Huang, Alexander Toshev, Oana Camburu, Alan Yuille, and
  Kevin Murphy.
\newblock Generation and comprehension of unambiguous object descriptions.
\newblock In \emph{Proceedings of the IEEE Conference on Computer Vision and
  Pattern Recognition (CVPR)}, 2016.

\bibitem[Matuszek et~al.(2012)Matuszek, Fitzgerald, Zettlemoyer, Bo, and
  Fox]{matuszek2012joint}
Cynthia Matuszek, Nicholas Fitzgerald, Luke Zettlemoyer, Liefeng Bo, and Dieter
  Fox.
\newblock A joint model of language and perception for grounded attribute
  learning.
\newblock In \emph{Proceedings of the International Conference on Machine
  Learning (ICML)}, 2012.

\bibitem[Nakashole et~al.(2013)Nakashole, Tylenda, and
  Weikum]{nakashole2013fine}
Ndapandula Nakashole, Tomasz Tylenda, and Gerhard Weikum.
\newblock Fine-grained semantic typing of emerging entities.
\newblock In \emph{Proceedings of the Annual Meeting of the Association for
  Computational Linguistics (ACL)}, 2013.

\bibitem[Noh et~al.(2015)Noh, Seo, and Han]{noh2015images}
Hyeonwoo Noh, Paul~Hongsuck Seo, and Bohyung Han.
\newblock Image question answering using convolutional neural network with
  dynamic parameter prediction.
\newblock \emph{arXiv:1511.05756}, 2015.

\bibitem[Pennington et~al.(2014)Pennington, Socher, and
  Manning]{pennington2014glove}
Jeffrey Pennington, Richard Socher, and Christopher~D. Manning.
\newblock Glove: Global vectors for word representation.
\newblock In \emph{Proceedings of the Conference on Empirical Methods in
  Natural Language Processing (EMNLP)}, 2014.

\bibitem[Plummer et~al.(2015)Plummer, Wang, Cervantes, Caicedo, Hockenmaier,
  and Lazebnik]{plummer15iccv}
Bryan Plummer, Liwei Wang, Chris Cervantes, Juan Caicedo, Julia Hockenmaier,
  and Svetlana Lazebnik.
\newblock Flickr30k entities: Collecting region-to-phrase correspondences for
  richer image-to-sentence models.
\newblock In \emph{Proceedings of the IEEE International Conference on Computer
  Vision (ICCV)}, 2015.

\bibitem[Plummer et~al.(2016)Plummer, Wang, Cervantes, Caicedo, Hockenmaier,
  and Lazebnik]{plummer16arxiv}
Bryan Plummer, Liwei Wang, Chris Cervantes, Juan Caicedo, Julia Hockenmaier,
  and Svetlana Lazebnik.
\newblock Flickr30k entities: Collecting region-to-phrase correspondences for
  richer image-to-sentence models.
\newblock \emph{arXiv:1505.04870}, 2016.

\bibitem[Prakash and Storer(2016)]{prakashhighway}
Aaditya Prakash and James Storer.
\newblock Highway networks for visual question answering.
\newblock 2016.

\bibitem[Regneri et~al.(2013)Regneri, Rohrbach, Wetzel, Thater, Schiele, and
  Pinkal]{regneri13tacl}
Michaela Regneri, Marcus Rohrbach, Dominikus Wetzel, Stefan Thater, Bernt
  Schiele, and Manfred Pinkal.
\newblock {Grounding Action Descriptions in Videos}.
\newblock \emph{Transactions of the Association for Computational Linguistics
  (TACL)}, 1, 2013.

\bibitem[Ren et~al.(2015)Ren, Kiros, and Zemel]{ren2015image}
Mengye Ren, Ryan Kiros, and Richard Zemel.
\newblock Image question answering: A visual semantic embedding model and a new
  dataset.
\newblock In \emph{Advances in Neural Information Processing Systems (NIPS)},
  2015.

\bibitem[Rohrbach et~al.(2015{\natexlab{a}})Rohrbach, Rohrbach, Hu, Darrell,
  and Schiele]{rohrbach16eccv}
Anna Rohrbach, Marcus Rohrbach, Ronghang Hu, Trevor Darrell, and Bernt Schiele.
\newblock Grounding of textual phrases in images by reconstruction.
\newblock In \emph{Proceedings of the European Conference on Computer Vision
  (ECCV)}, 2015{\natexlab{a}}.

\bibitem[Rohrbach et~al.(2015{\natexlab{b}})Rohrbach, Rohrbach, Tandon, and
  Schiele]{rohrbach15cvpr}
Anna Rohrbach, Marcus Rohrbach, Niket Tandon, and Bernt Schiele.
\newblock A dataset for movie description.
\newblock In \emph{Proceedings of the IEEE Conference on Computer Vision and
  Pattern Recognition (CVPR)}, 2015{\natexlab{b}}.

\bibitem[Russakovsky et~al.(2014)Russakovsky, Deng, Su, Krause, Satheesh, Ma,
  Huang, Karpathy, Khosla, Bernstein, Berg, and Fei-Fei]{ILSVRCarxiv14}
Olga Russakovsky, Jia Deng, Hao Su, Jonathan Krause, Sanjeev Satheesh, Sean Ma,
  Zhiheng Huang, Andrej Karpathy, Aditya Khosla, Michael Bernstein,
  Alexander~C. Berg, and Li~Fei-Fei.
\newblock Imagenet large scale visual recognition challenge.
\newblock \emph{arXiv:1409.0575}, 2014.

\bibitem[Saito et~al.(2016)Saito, Shin, Ushiku, and Harada]{saito2016dualnet}
Kuniaki Saito, Andrew Shin, Yoshitaka Ushiku, and Tatsuya Harada.
\newblock Dualnet: Domain-invariant network for visual question answering.
\newblock \emph{arXiv preprint arXiv:1606.06108}, 2016.

\bibitem[Shih et~al.(2016)Shih, Singh, and Hoiem]{shih2015look}
Kevin~J Shih, Saurabh Singh, and Derek Hoiem.
\newblock Where to look: Focus regions for visual question answering.
\newblock In \emph{Proceedings of the IEEE Conference on Computer Vision and
  Pattern Recognition (CVPR)}, 2016.

\bibitem[Silberman et~al.(2012)Silberman, Hoiem, Kohli, and
  Fergus]{silbermanECCV12}
Nathan Silberman, Derek Hoiem, Pushmeet Kohli, and Rob Fergus.
\newblock Indoor segmentation and support inference from rgbd images.
\newblock In \emph{Proceedings of the European Conference on Computer Vision
  (ECCV)}, 2012.

\bibitem[Simonyan and Zisserman(2014)]{simonyan2014very}
Karen Simonyan and Andrew Zisserman.
\newblock Very deep convolutional networks for large-scale image recognition.
\newblock \emph{arXiv:1409.1556}, 2014.

\bibitem[Sutskever et~al.(2014)Sutskever, Vinyals, and Le]{sutskever14nips}
Ilya Sutskever, Oriol Vinyals, and Quoc V.~V Le.
\newblock Sequence to sequence learning with neural networks.
\newblock In \emph{Advances in Neural Information Processing Systems (NIPS)},
  2014.

\bibitem[Szegedy et~al.(2014)Szegedy, Liu, Jia, Sermanet, Reed, Anguelov,
  Erhan, Vanhoucke, and Rabinovich]{szegedy2014going}
Christian Szegedy, Wei Liu, Yangqing Jia, Pierre Sermanet, Scott Reed, Dragomir
  Anguelov, Dumitru Erhan, Vincent Vanhoucke, and Andrew Rabinovich.
\newblock Going deeper with convolutions.
\newblock \emph{arXiv:1409.4842}, 2014.

\bibitem[Tapaswi et~al.(2016)Tapaswi, Zhu, Stiefelhagen, Torralba, Urtasun, and
  Fidler]{tapaswi16cvpr}
Makarand Tapaswi, Yukun Zhu, Rainer Stiefelhagen, Antonio Torralba, Raquel
  Urtasun, and Sanja Fidler.
\newblock Movieqa: Understanding stories in movies through question-answering.
\newblock In \emph{Proceedings of the IEEE Conference on Computer Vision and
  Pattern Recognition (CVPR)}, 2016.

\bibitem[Trecvid(2014)]{trecvid_med14}
Trecvid.
\newblock Trecvid med 14.
\newblock \url{http://nist.gov/itl/iad/mig/med14.cfm}, 2014.

\bibitem[Venugopalan et~al.(2015{\natexlab{a}})Venugopalan, Rohrbach, Donahue,
  Mooney, Darrell, and Saenko]{venugopalan15iccv}
Subhashini Venugopalan, Marcus Rohrbach, Jeff Donahue, Raymond Mooney, Trevor
  Darrell, and Kate Saenko.
\newblock Sequence to sequence -- video to text.
\newblock In \emph{Proceedings of the IEEE International Conference on Computer
  Vision (ICCV)}, 2015{\natexlab{a}}.

\bibitem[Venugopalan et~al.(2015{\natexlab{b}})Venugopalan, Xu, Donahue,
  Rohrbach, Mooney, and Saenko]{venugopalan15naacl}
Subhashini Venugopalan, Huijuan Xu, Jeff Donahue, Marcus Rohrbach, Raymond
  Mooney, and Kate Saenko.
\newblock Translating videos to natural language using deep recurrent neural
  networks.
\newblock In \emph{Proceedings of the Conference of the North American Chapter
  of the Association for Computational Linguistics (NAACL)},
  2015{\natexlab{b}}.

\bibitem[Vinyals et~al.(2014)Vinyals, Toshev, Bengio, and
  Erhan]{vinyals2014show}
Oriol Vinyals, Alexander Toshev, Samy Bengio, and Dumitru Erhan.
\newblock Show and tell: A neural image caption generator.
\newblock \emph{arXiv:1411.4555}, 2014.

\bibitem[Wang et~al.(2016)Wang, Li, and Lazebnik]{wang2016cvpr}
Liwei Wang, Yin Li, and Svetlana Lazebnik.
\newblock Learning deep structure-preserving image-text embeddings.
\newblock In \emph{Proceedings of the IEEE Conference on Computer Vision and
  Pattern Recognition (CVPR)}, 2016.

\bibitem[Weston et~al.(2014)Weston, Chopra, and Bordes]{weston2014memory}
Jason Weston, Sumit Chopra, and Antoine Bordes.
\newblock Memory networks.
\newblock \emph{arXiv:1410.3916}, 2014.

\bibitem[Wu et~al.(2016)Wu, Wang, Shen, Hengel, and Dick]{wu16cvpr}
Qi~Wu, Peng Wang, Chunhua Shen, Anton van~den Hengel, and Anthony Dick.
\newblock {Ask Me Anything: Free-form Visual Question Answering Based on
  Knowledge from External Sources}.
\newblock In \emph{Proceedings of the IEEE Conference on Computer Vision and
  Pattern Recognition (CVPR)}, 2016.

\bibitem[Wu and Palmer(1994)]{wu1994verbs}
Zhibiao Wu and Martha Palmer.
\newblock Verbs semantics and lexical selection.
\newblock In \emph{Proceedings of the Annual Meeting of the Association for
  Computational Linguistics (ACL)}, 1994.

\bibitem[Xiong et~al.(2016)Xiong, Merity, and Socher]{xiong16dynamic}
Caiming Xiong, Stephen Merity, and Richard Socher.
\newblock Dynamic memory networks for visual and textual question answering.
\newblock \emph{arXiv preprint arXiv:1603.01417}, 2016.

\bibitem[Xu and Saenko(2015)]{xu2015ask}
Huijuan Xu and Kate Saenko.
\newblock Ask, attend and answer: Exploring question-guided spatial attention
  for visual question answering.
\newblock \emph{arXiv:1511.05234}, 2015.

\bibitem[Xu et~al.(2015)Xu, Ba, Kiros, Courville, Salakhutdinov, Zemel, and
  Bengio]{xu15icml}
Kelvin Xu, Jimmy Ba, Ryan Kiros, Aaron Courville, Ruslan Salakhutdinov, Richard
  Zemel, and Yoshua Bengio.
\newblock Show, attend and tell: Neural image caption generation with visual
  attention.
\newblock \emph{Proceedings of the International Conference on Machine Learning
  (ICML)}, 2015.

\bibitem[Yang et~al.(2015)Yang, He, Gao, Deng, and Smola]{yang2015stacked}
Zichao Yang, Xiaodong He, Jianfeng Gao, Li~Deng, and Alex Smola.
\newblock Stacked attention networks for image question answering.
\newblock \emph{arXiv:1511.02274}, 2015.

\bibitem[Yu et~al.(2015)Yu, Park, Berg, and Berg]{yu2015visual}
Licheng Yu, Eunbyung Park, Alexander~C Berg, and Tamara~L Berg.
\newblock Visual madlibs: Fill in the blank description generation and question
  answering.
\newblock In \emph{Proceedings of the IEEE International Conference on Computer
  Vision (ICCV)}, pages 2461--2469, 2015.

\bibitem[Yu et~al.(2016)Yu, Poirson, Yang, Berg, and Berg]{yu16eccv}
Licheng Yu, Patrick Poirson, Shan Yang, Alexander~C Berg, and Tamara~L Berg.
\newblock Modeling context in referring expressions.
\newblock In \emph{European Conference on Computer Vision}, pages 69--85.
  Springer, 2016.

\bibitem[Zaremba and Sutskever(2014)]{zaremba14arxiv}
Wojciech Zaremba and Ilya Sutskever.
\newblock Learning to execute.
\newblock \emph{arXiv preprint arXiv:1410.4615}, 2014.

\bibitem[Zhou et~al.(2015)Zhou, Tian, Sukhbaatar, Szlam, and
  Fergus]{zhou2015simple}
Bolei Zhou, Yuandong Tian, Sainbayar Sukhbaatar, Arthur Szlam, and Rob Fergus.
\newblock Simple baseline for visual question answering.
\newblock \emph{arXiv:1512.02167}, 2015.

\bibitem[Zhu et~al.(2015)Zhu, Xu, Yang, and Hauptmann]{zhu2015uncovering}
Linchao Zhu, Zhongwen Xu, Yi~Yang, and Alexander~G Hauptmann.
\newblock Uncovering temporal context for video question and answering.
\newblock \emph{arXiv:1511.04670}, 2015.

\bibitem[Zhu et~al.(2016)Zhu, Groth, Bernstein, and Fei-Fei]{zhu16cvpr}
Yuke Zhu, Oliver Groth, Michael Bernstein, and Li~Fei-Fei.
\newblock {Visual7W: Grounded Question Answering in Images}.
\newblock In \emph{Proceedings of the IEEE Conference on Computer Vision and
  Pattern Recognition (CVPR)}, 2016.

\bibitem[Zitnick et~al.(2013)Zitnick, Parikh, and
  Vanderwende]{zitnick2013learning}
C~Lawrence Zitnick, Devi Parikh, and Lucy Vanderwende.
\newblock Learning the visual interpretation of sentences.
\newblock In \emph{Proceedings of the IEEE International Conference on Computer
  Vision (ICCV)}, 2013.

\end{thebibliography}

\end{document}